\def\ARXIVVERSION{1}
\ifdefined\ARXIVVERSION
  \documentclass[11pt]{article}
  \usepackage[a4paper,margin=0.92in]{geometry}
  \usepackage{lmodern}
  \usepackage{natbib}
\else
  \documentclass{article}
  \usepackage{iclr2027_conference,times}
\fi

\usepackage[T1]{fontenc}
\usepackage[utf8]{inputenc}
\usepackage{microtype}
\usepackage{graphicx}
\usepackage{booktabs}
\usepackage{longtable}
\usepackage{array}
\usepackage{tabularx}
\usepackage{amsmath,amssymb}
\usepackage{xcolor}
\usepackage{seqsplit}
\usepackage{caption}
\usepackage{hyperref}
\usepackage[nameinlink,noabbrev,capitalise]{cleveref}
\usepackage{placeins}

\definecolor{directblue}{HTML}{0072B2}
\hypersetup{colorlinks=true,allcolors=directblue}
\newcommand{\revision}[1]{\texttt{\seqsplit{#1}}}
\newcommand{\glmwrapper}[1]{\texttt{Answer:\ <|begin\_of\_box|>\allowbreak#1\allowbreak<|end\_of\_box|>}}
\newcolumntype{Y}{>{\raggedright\arraybackslash}X}
\newcolumntype{L}[1]{>{\raggedright\arraybackslash}p{#1}}
\crefname{appendix}{Appendix}{Appendices}
\Crefname{appendix}{Appendix}{Appendices}

\title{When Video Verifier Scores Mislead:\\
Capability Assessment and Verifier Selection}
\ifdefined\ARXIVVERSION
  \author{Charles Weng \qquad Zixi Huang\\[0.3em]Johns Hopkins University}
  \hypersetup{
    pdftitle={When Video Verifier Scores Mislead: Capability Assessment and Verifier Selection},
    pdfauthor={Charles Weng and Zixi Huang}
  }
\else
  \author{Anonymous authors}
\fi
\date{}

\definecolor{revisionblue}{RGB}{0,62,151}
\newif\ifshowrev
\showrevfalse 
\newcommand{\revcolor}{\ifshowrev\color{revisionblue}\fi}
\newenvironment{revisedtext}{\par\begingroup\revcolor}{\par\endgroup}
\DeclareRobustCommand{\rev}[1]{\ifshowrev\textcolor{revisionblue}{#1}\else#1\fi}
\newcommand{\RevisionPositionTables}{%
\begin{table}[ht]
\centering\footnotesize\revcolor\setlength{\tabcolsep}{5pt}
\caption{\rev{Truth-switch label mass and same-runtime accuracy (\%). Dash means no matched generation cell.}}
\label{tab:position-truth}
\begin{tabular}{@{}llrrrrr@{}}
\toprule
Model & Input or negative & Original mass & Validated mass & Original direct & Validated direct & CoT \\
\midrule
Qwen & Event text & 99.84 & 98.78 & 10.00 & 10.00 & 100.00 \\
Qwen & Frames & 99.67 & 99.74 & 5.00 & 9.17 & 100.00 \\
Qwen & State text & 100.00 & 99.39 & 100.00 & 100.00 & 96.67 \\
GLM & Event text & 0.59 & 95.22 & 8.33 & 10.83 & 96.67 \\
GLM & Frames & 0.00 & 91.12 & 0.00 & 9.17 & 90.00 \\
GLM & State text & 0.91 & 98.78 & 100.00 & 100.00 & 100.00 \\
InternVL & Event text & 4.93 & 88.96 & 12.50 & 12.50 & 90.83 \\
InternVL & Frames & 11.55 & 99.00 & 4.17 & 5.00 & 70.00 \\
InternVL & State text & 32.67 & 99.53 & 100.00 & 100.00 & 100.00 \\
Molmo & Event text & 98.92 & 91.51 & 10.83 & 10.83 & 37.50 \\
Molmo & Frames & 99.96 & 87.72 & 3.33 & 1.67 & 25.00 \\
Molmo & State text & 99.75 & 93.76 & 100.00 & 100.00 & 90.00 \\
\bottomrule
\end{tabular}
\end{table}

\FloatBarrier

\begin{table}[ht]
\centering\footnotesize\revcolor\setlength{\tabcolsep}{5pt}
\caption{\rev{Ordered natural video label mass and same-runtime accuracy (\%). Dash means no matched generation cell.}}
\label{tab:position-ordered}
\begin{tabular}{@{}llrrrrr@{}}
\toprule
Model & Input or negative & Original mass & Validated mass & Original direct & Validated direct & CoT \\
\midrule
Qwen & Denial & 100.00 & 99.99 & 62.38 & 66.62 & 43.12 \\
Qwen & Content & 99.99 & 99.99 & 64.00 & 65.50 & 64.12 \\
GLM & Denial & 0.80 & 94.89 & 61.12 & 63.25 & 57.12 \\
GLM & Content & 0.47 & 94.40 & 56.12 & 59.00 & 63.75 \\
InternVL & Denial & 100.00 & 99.97 & 81.62 & 86.50 & 63.62 \\
InternVL & Content & 100.00 & 100.00 & 53.87 & 54.50 & 60.12 \\
Molmo & Denial & 99.99 & 91.61 & 30.50 & 37.50 & 32.38 \\
Molmo & Content & 99.99 & 91.14 & 57.88 & 59.38 & 53.25 \\
\bottomrule
\end{tabular}
\end{table}

\begin{table}[ht]
\centering\footnotesize\revcolor\setlength{\tabcolsep}{5pt}
\caption{\rev{Shuffled natural video label mass and same-runtime accuracy (\%). Dash means no matched generation cell.}}
\label{tab:position-shuffled}
\begin{tabular}{@{}llrrrrr@{}}
\toprule
Model & Input or negative & Original mass & Validated mass & Original direct & Validated direct & CoT \\
\midrule
Qwen & Denial & 100.00 & 99.99 & 56.62 & 61.00 & -- \\
Qwen & Content & 99.99 & 99.99 & 63.75 & 65.88 & -- \\
GLM & Denial & 0.71 & 94.45 & 62.00 & 64.50 & -- \\
GLM & Content & 0.32 & 93.86 & 54.75 & 57.50 & -- \\
InternVL & Denial & 100.00 & 99.96 & 84.75 & 88.38 & -- \\
InternVL & Content & 100.00 & 100.00 & 56.38 & 55.88 & -- \\
Molmo & Denial & 99.99 & 91.65 & 28.88 & 36.12 & -- \\
Molmo & Content & 99.99 & 90.90 & 56.50 & 58.75 & -- \\
\bottomrule
\end{tabular}
\end{table}

\begin{table}[ht]
\centering\footnotesize\revcolor\setlength{\tabcolsep}{5pt}
\caption{\rev{No video natural video label mass and same-runtime accuracy (\%). Dash means no matched generation cell.}}
\label{tab:position-none}
\begin{tabular}{@{}llrrrrr@{}}
\toprule
Model & Input or negative & Original mass & Validated mass & Original direct & Validated direct & CoT \\
\midrule
Qwen & Denial & 100.00 & 99.97 & 2.88 & 8.25 & -- \\
Qwen & Content & 99.99 & 99.93 & 58.75 & 55.88 & -- \\
GLM & Denial & 0.96 & 89.91 & 42.00 & 51.00 & -- \\
GLM & Content & 0.71 & 88.89 & 50.25 & 57.75 & -- \\
InternVL & Denial & 100.00 & 71.38 & 66.75 & 74.62 & -- \\
InternVL & Content & 100.00 & 58.88 & 51.50 & 53.62 & -- \\
Molmo & Denial & 99.96 & 89.94 & 40.62 & 37.38 & -- \\
Molmo & Content & 99.95 & 88.68 & 53.87 & 54.00 & -- \\
\bottomrule
\end{tabular}
\end{table}

\FloatBarrier
}
\newcommand{\RevisionNaturalTables}{%
\begin{table}[ht]
\centering\small\revcolor
\caption{\rev{CoT minus direct (expressed choice). Simultaneous 95\% intervals, points.}}
\label{tab:interface-expressed}
\begin{tabular}{@{}lrrr@{}}
\toprule
Model & Event denial & Content-matched & Denial minus content \\
\midrule
Qwen & $-23.50 [-30.50, -16.50]$ & $-1.38 [-8.38, +5.63]$ & $-22.12 [-29.12, -15.12]$ \\
GLM & $-6.12 [-13.12, +0.88]$ & $+4.75 [-2.25, +11.75]$ & $-10.88 [-17.88, -3.87]$ \\
InternVL & $-22.88 [-29.88, -15.88]$ & $+5.62 [-1.38, +12.62]$ & $-28.50 [-35.50, -21.50]$ \\
Molmo & $-5.12 [-12.12, +1.88]$ & $-6.12 [-13.12, +0.88]$ & $+1.00 [-6.00, +8.00]$ \\
\bottomrule
\end{tabular}
\end{table}

\begin{table}[ht]
\centering\small\revcolor
\caption{\rev{CoT minus direct (caps counted wrong). Simultaneous 95\% intervals, points.}}
\label{tab:interface-cap-fail}
\begin{tabular}{@{}lrrr@{}}
\toprule
Model & Event denial & Content-matched & Denial minus content \\
\midrule
Qwen & $-23.50 [-30.50, -16.50]$ & $-1.38 [-8.38, +5.63]$ & $-22.12 [-29.12, -15.12]$ \\
GLM & $-6.25 [-13.25, +0.75]$ & $+4.75 [-2.25, +11.75]$ & $-11.00 [-18.00, -4.00]$ \\
InternVL & $-22.88 [-29.88, -15.88]$ & $+5.62 [-1.38, +12.62]$ & $-28.50 [-35.50, -21.50]$ \\
Molmo & $-5.12 [-12.12, +1.88]$ & $-6.12 [-13.12, +0.88]$ & $+1.00 [-6.00, +8.00]$ \\
\bottomrule
\end{tabular}
\end{table}

\FloatBarrier

\begin{table}[ht]
\centering\footnotesize\revcolor
\caption{\rev{Qwen natural-video generation settings. Accuracy and consistency are percentages.}}
\label{tab:generation-qwen}
\setlength{\tabcolsep}{4.75pt}
\begin{tabular}{@{}lllrrrrrrr@{}}
\toprule
Negative & Prompt & Mode & Videos & Effective cap & Acc. & Tokens & Cap & Unparsed & Consist. \\
\midrule
Denial & G0 & off & 400 & 8,192 & 61.75 & 4 & 0 & 0 & 83.50 \\
Denial & G1 & off & 400 & 8,192 & 43.12 & 314 & 0 & 0 & 74.75 \\
Denial & P0 & off & 400 & 8,192 & 57.00 & 4 & 0 & 0 & 87.00 \\
Denial & P1 & off & 400 & 8,192 & 43.88 & 459 & 2 & 2 & 80.75 \\
Denial & Original concise & off & 400 & 8,192 & 62.75 & 2 & 0 & 0 & 83.50 \\
Content & G0 & off & 400 & 8,192 & 66.12 & 4 & 0 & 0 & 82.75 \\
Content & G1 & off & 400 & 8,192 & 64.12 & 365 & 4 & 4 & 78.00 \\
Content & P0 & off & 400 & 8,192 & 66.38 & 4 & 0 & 0 & 83.75 \\
Content & P1 & off & 400 & 8,192 & 65.00 & 500 & 10 & 9 & 78.25 \\
Content & Original concise & off & 400 & 8,192 & 65.25 & 2 & 0 & 0 & 75.50 \\
\bottomrule
\end{tabular}
\end{table}

\FloatBarrier

\begin{table}[ht]
\centering\footnotesize\revcolor
\caption{\rev{GLM natural-video generation settings. Accuracy and consistency are percentages.}}
\label{tab:generation-glm}
\setlength{\tabcolsep}{4.75pt}
\begin{tabular}{@{}lllrrrrrrr@{}}
\toprule
Negative & Prompt & Mode & Videos & Effective cap & Acc. & Tokens & Cap & Unparsed & Consist. \\
\midrule
Denial & G0 & off & 400 & 5,927--8,192 & 53.37 & 41 & 0 & 0 & 79.25 \\
Denial & G1 & off & 400 & 5,923--8,192 & 57.12 & 266 & 10 & 8 & 72.00 \\
Denial & P0 & off & 80 & 6,120--8,192 & 64.38 & 14 & 0 & 0 & 81.25 \\
Denial & P1 & off & 400 & 5,864--8,192 & 55.38 & 400 & 10 & 8 & 70.50 \\
Denial & Original concise & off & 400 & 6,066--8,192 & 53.00 & 79 & 0 & 0 & 73.50 \\
Denial & G0 & on & 400 & 5,931--8,192 & 43.12 & 603 & 1 & 1 & 79.75 \\
Denial & G1 & on & 400 & 5,927--8,192 & 44.12 & 854 & 2 & 2 & 81.75 \\
Denial & P0 & on & 80 & 6,124--8,192 & 48.12 & 875 & 2 & 2 & 81.25 \\
Denial & P1 & on & 80 & 6,120--8,192 & 50.62 & 1192 & 1 & 1 & 73.75 \\
Content & G0 & off & 400 & 5,918--8,192 & 63.00 & 32 & 0 & 0 & 79.50 \\
Content & G1 & off & 400 & 5,914--8,192 & 63.75 & 141 & 0 & 0 & 79.00 \\
Content & P0 & off & 80 & 6,114--8,192 & 60.00 & 10 & 0 & 0 & 70.00 \\
Content & P1 & off & 400 & 5,855--8,192 & 62.50 & 183 & 0 & 0 & 77.00 \\
Content & Original concise & off & 400 & 6,057--8,192 & 59.75 & 106 & 0 & 0 & 80.00 \\
Content & G0 & on & 400 & 5,922--8,192 & 63.25 & 704 & 0 & 0 & 76.50 \\
Content & G1 & on & 400 & 5,918--8,192 & 64.50 & 998 & 1 & 1 & 79.50 \\
Content & P0 & on & 80 & 6,118--8,192 & 58.75 & 1164 & 4 & 4 & 78.75 \\
Content & P1 & on & 80 & 6,114--8,192 & 56.88 & 1322 & 1 & 1 & 81.25 \\
\bottomrule
\end{tabular}
\end{table}

\FloatBarrier

\begin{table}[ht]
\centering\footnotesize\revcolor
\caption{\rev{InternVL natural-video generation settings. Accuracy and consistency are percentages.}}
\label{tab:generation-internvl}
\setlength{\tabcolsep}{4.75pt}
\begin{tabular}{@{}lllrrrrrrr@{}}
\toprule
Negative & Prompt & Mode & Videos & Effective cap & Acc. & Tokens & Cap & Unparsed & Consist. \\
\midrule
Denial & G0 & off & 400 & 3,417--8,192 & 86.62 & 13 & 0 & 0 & 91.25 \\
Denial & G1 & off & 400 & 3,413--8,192 & 63.62 & 75 & 0 & 0 & 74.25 \\
Denial & P0 & off & 400 & 3,358--8,192 & 85.12 & 4 & 0 & 0 & 89.75 \\
Denial & P1 & off & 400 & 3,354--8,192 & 60.50 & 221 & 0 & 0 & 68.00 \\
Denial & Original concise & off & 400 & 3,556--8,192 & 82.50 & 2 & 0 & 0 & 89.00 \\
Denial & P0 & on & 80 & 3,239--8,192 & 70.62 & 293 & 0 & 0 & 76.25 \\
Denial & P1 & on & 80 & 3,235--8,192 & 68.12 & 331 & 0 & 0 & 68.75 \\
Content & G0 & off & 400 & 3,389--8,192 & 57.00 & 26 & 0 & 0 & 65.50 \\
Content & G1 & off & 400 & 3,385--8,192 & 60.12 & 69 & 0 & 0 & 72.25 \\
Content & P0 & off & 400 & 3,330--8,192 & 57.25 & 5 & 0 & 0 & 77.50 \\
Content & P1 & off & 400 & 3,326--8,192 & 59.13 & 212 & 0 & 0 & 74.25 \\
Content & Original concise & off & 400 & 3,528--8,192 & 56.88 & 2 & 0 & 0 & 80.25 \\
Content & P0 & on & 80 & 3,211--8,192 & 56.25 & 281 & 0 & 0 & 77.50 \\
Content & P1 & on & 80 & 3,207--8,192 & 58.75 & 364 & 0 & 1 & 75.00 \\
\bottomrule
\end{tabular}
\end{table}

\FloatBarrier

\begin{table}[ht]
\centering\footnotesize\revcolor
\caption{\rev{Molmo natural-video generation settings. Accuracy and consistency are percentages.}}
\label{tab:generation-molmo}
\setlength{\tabcolsep}{4.75pt}
\begin{tabular}{@{}lllrrrrrrr@{}}
\toprule
Negative & Prompt & Mode & Videos & Effective cap & Acc. & Tokens & Cap & Unparsed & Consist. \\
\midrule
Denial & G0 & off & 400 & 3,057--8,192 & 27.38 & 22 & 0 & 0 & 71.25 \\
Denial & G1 & off & 400 & 3,053--8,192 & 32.38 & 76 & 0 & 1 & 79.50 \\
Denial & P0 & off & 80 & 2,999--8,192 & 28.75 & 43 & 0 & 0 & 80.00 \\
Denial & Original concise & off & 400 & 3,196--8,192 & 30.50 & 2 & 0 & 0 & 78.00 \\
Content & G0 & off & 400 & 3,029--8,192 & 53.37 & 27 & 0 & 4 & 45.50 \\
Content & G1 & off & 400 & 3,025--8,192 & 53.25 & 99 & 4 & 75 & 65.50 \\
Content & P0 & off & 80 & 2,971--8,192 & 58.75 & 56 & 0 & 17 & 55.00 \\
Content & Original concise & off & 400 & 3,168--8,192 & 57.50 & 2 & 0 & 0 & 64.50 \\
\bottomrule
\end{tabular}
\end{table}

\FloatBarrier

\begin{table}[ht]
\centering\small\revcolor
\caption{\rev{CoT minus direct by question type under the common protocol. Descriptive changes in points, event-denial / content-matched.}}
\label{tab:revision-natural-by-type}
\begin{tabular}{@{}lrrrr@{}}
\toprule
Question type & Qwen & GLM & InternVL & Molmo \\
\midrule
Event order & $-24.17$ / $-10.83$ & $+1.25$ / $+0.83$ & $-14.17$ / $+11.25$ & $-14.17$ / $-8.75$ \\
Event-relative state & $-36.50$ / $+2.00$ & $-18.00$ / $+2.00$ & $-31.00$ / $+3.00$ & $+13.00$ / $+12.00$ \\
Temporal location & $-12.00$ / $+10.00$ & $+10.50$ / $+15.00$ & $-28.00$ / $+5.00$ & $-0.50$ / $-31.00$ \\
Visibility continuity & $-20.62$ / $-5.62$ & $-23.12$ / $+1.25$ & $-19.38$ / $+1.25$ & $-20.00$ / $+6.25$ \\
\bottomrule
\end{tabular}
\end{table}
}
\newcommand{\RevisionTransferTables}{%
\begin{table}[ht]
\centering\small\revcolor
\caption{\rev{Cross-panel verifier-selection cost with pointwise 95\% intervals, in points.}}
\label{tab:selection-cross-panel-sets}
\begin{tabular}{@{}lllr@{}}
\toprule
Eligible set & Select on & Evaluate on & Cost \\
\midrule
Eight, direct as submitted & TOC-Bench & Vinoground & $+3.25 [-0.41, +5.39]$ \\
Eight, direct as submitted & Vinoground & TOC-Bench & $+9.25 [+4.75, +14.25]$ \\
Eight, validated direct & TOC-Bench & Vinoground & $+3.63 [+0.16, +5.66]$ \\
Eight, validated direct & Vinoground & TOC-Bench & $+8.88 [+4.75, +14.75]$ \\
Twelve, with Kimi-VL and Aria & TOC-Bench & Vinoground & $+3.63 [+0.01, +5.66]$ \\
Twelve, with Kimi-VL and Aria & Vinoground & TOC-Bench & $+8.88 [+4.75, +14.75]$ \\
\bottomrule
\end{tabular}
\end{table}

\begin{table}[ht]
\centering\small\revcolor
\caption{\rev{Interaction of interface and negative type with validated direct scores: the CoT-minus-direct
change against event-denial negatives minus the change against content-matched negatives.
Pointwise 95\% intervals, points.}}
\label{tab:transfer-interaction}
\begin{tabular}{@{}lrr@{}}
\toprule
Model & TOC-Bench & Vinoground \\
\midrule
Qwen & $-20.50 [-26.25, -14.75]$ & $-5.44 [-8.44, -2.52]$ \\
GLM & $-25.00 [-30.25, -19.75]$ & $-12.62 [-15.45, -9.98]$ \\
InternVL & $-17.75 [-23.50, -11.88]$ & $-6.25 [-9.20, -3.27]$ \\
Molmo & $+6.75 [+1.25, +12.38]$ & $-2.69 [-6.40, +1.09]$ \\
Kimi & $-4.38 [-7.62, -1.25]$ & $+1.75 [-0.10, +3.65]$ \\
Aria & $+0.25 [-4.88, +5.50]$ & $+1.88 [-1.41, +5.21]$ \\
\bottomrule
\end{tabular}
\end{table}

\begin{table}[ht]
\centering\footnotesize\revcolor
\caption{\rev{Fixed-caption Vinoground paired success (\%) on 400 pairs. Pointwise 95\% intervals.}}
\label{tab:transfer-paired}
\begin{tabular}{@{}llrr@{}}
\toprule
Model & Interface & Paired success & 95\% interval \\
\midrule
Qwen & Direct, original position & 63.25 & [59.16, 67.37] \\
Qwen & CoT judgment & 61.75 & [57.56, 65.98] \\
GLM & Direct, original position & 51.50 & [47.44, 55.48] \\
GLM & CoT judgment & 64.62 & [60.47, 68.73] \\
InternVL & Direct, original position & 57.25 & [53.13, 61.36] \\
InternVL & CoT judgment & 54.87 & [50.79, 58.98] \\
Molmo & Direct, original position & 33.00 & [29.41, 36.71] \\
Molmo & CoT judgment & 35.00 & [31.21, 38.89] \\
Qwen & Validated direct & 63.50 & [59.36, 67.62] \\
GLM & Validated direct & 58.75 & [54.71, 62.79] \\
InternVL & Validated direct & 56.38 & [52.18, 60.54] \\
Molmo & Validated direct & 31.75 & [28.13, 35.49] \\
Kimi & Validated direct & 37.62 & [33.71, 41.66] \\
Kimi & CoT judgment & 37.50 & [33.62, 41.34] \\
Aria & Validated direct & 35.38 & [31.71, 39.10] \\
Aria & CoT judgment & 34.88 & [31.25, 38.47] \\
\bottomrule
\end{tabular}
\end{table}
}

\newcommand{\RevisionSubmittedPositionTables}{%
\begin{table}[ht]
\centering\footnotesize\revcolor
\caption{\rev{Qwen submitted natural-video cells. Accuracy is the submitted score. Mass, variance, and top-1 label rate are measured before answer generation on the original rendered input.}}
\label{tab:original-qwen}
\begin{tabular}{@{}lllrrrr@{}}
\toprule
Evidence & Negative & Interface & Accuracy (\%) & A/B mass (\%) & Margin variance & A/B top 1 (\%) \\
\midrule
Ordered & Denial & Direct & 61.38 & 100.000 & 19.6 & 100.0 \\
Ordered & Denial & CoT & 43.75 & 0.006 & 1.9 & 0.0 \\
Ordered & Content & Direct & 64.38 & 99.992 & 14.9 & 100.0 \\
Ordered & Content & CoT & 63.88 & 0.002 & 1.17 & 0.0 \\
Shuffled & Denial & Direct & 56.38 & 100.000 & 19.6 & 100.0 \\
Shuffled & Denial & CoT & 39.88 & 0.011 & 2.5 & 0.0 \\
Shuffled & Content & Direct & 63.50 & 99.995 & 14.3 & 100.0 \\
Shuffled & Content & CoT & 61.38 & 0.003 & 1.69 & 0.0 \\
No video & Denial & Direct & 2.88 & 100.000 & 126 & 100.0 \\
No video & Denial & CoT & 22.25 & 0.000 & 6.74 & 0.0 \\
No video & Content & Direct & 58.63 & 99.993 & 35.8 & 100.0 \\
No video & Content & CoT & 52.50 & 0.000 & 0.204 & 0.0 \\
\bottomrule
\end{tabular}
\end{table}
\FloatBarrier
\begin{table}[ht]
\centering\footnotesize\revcolor
\caption{\rev{GLM submitted natural-video cells. Accuracy is the submitted score. Mass, variance, and top-1 label rate are measured before answer generation on the original rendered input.}}
\label{tab:original-glm}
\begin{tabular}{@{}lllrrrr@{}}
\toprule
Evidence & Negative & Interface & Accuracy (\%) & A/B mass (\%) & Margin variance & A/B top 1 (\%) \\
\midrule
Ordered & Denial & Direct & 61.12 & 0.800 & 6.68 & 0.0 \\
Ordered & Denial & CoT & 29.00 & 0.000 & 0.138 & 0.0 \\
Ordered & Content & Direct & 55.62 & 0.471 & 5.99 & 0.0 \\
Ordered & Content & CoT & 48.25 & 0.000 & 0.163 & 0.0 \\
Shuffled & Denial & Direct & 61.12 & 0.712 & 6.37 & 0.0 \\
Shuffled & Denial & CoT & 31.37 & 0.000 & 0.142 & 0.0 \\
Shuffled & Content & Direct & 54.37 & 0.322 & 5.19 & 0.0 \\
Shuffled & Content & CoT & 47.12 & 0.000 & 0.152 & 0.0 \\
No video & Denial & Direct & 43.12 & 0.955 & 1.74 & 0.0 \\
No video & Denial & CoT & 23.50 & 0.000 & 0.281 & 0.0 \\
No video & Content & Direct & 50.12 & 0.709 & 2.12 & 0.0 \\
No video & Content & CoT & 47.75 & 0.000 & 0.168 & 0.0 \\
\bottomrule
\end{tabular}
\end{table}
\FloatBarrier
\begin{table}[ht]
\centering\footnotesize\revcolor
\caption{\rev{InternVL submitted natural-video cells. Accuracy is the submitted score. Mass, variance, and top-1 label rate are measured before answer generation on the original rendered input.}}
\label{tab:original-internvl}
\begin{tabular}{@{}lllrrrr@{}}
\toprule
Evidence & Negative & Interface & Accuracy (\%) & A/B mass (\%) & Margin variance & A/B top 1 (\%) \\
\midrule
Ordered & Denial & Direct & 81.88 & 99.999 & 7.63 & 100.0 \\
Ordered & Denial & CoT & 73.88 & 0.233 & 13.4 & 0.2 \\
Ordered & Content & Direct & 54.12 & 99.998 & 2.34 & 100.0 \\
Ordered & Content & CoT & 59.38 & 0.134 & 10.4 & 0.1 \\
Shuffled & Denial & Direct & 84.62 & 99.999 & 6.82 & 100.0 \\
Shuffled & Denial & CoT & 72.50 & 0.166 & 13.4 & 0.0 \\
Shuffled & Content & Direct & 55.50 & 99.998 & 1.58 & 100.0 \\
Shuffled & Content & CoT & 57.75 & 0.150 & 10.5 & 0.1 \\
No video & Denial & Direct & 66.88 & 99.998 & 1.81 & 100.0 \\
No video & Denial & CoT & 31.75 & 0.000 & 0.26 & 0.0 \\
No video & Content & Direct & 51.75 & 99.997 & 1.47 & 100.0 \\
No video & Content & CoT & 52.50 & 0.000 & 0.214 & 0.0 \\
\bottomrule
\end{tabular}
\end{table}
\FloatBarrier
\begin{table}[ht]
\centering\footnotesize\revcolor
\caption{\rev{Molmo submitted natural-video cells. Accuracy is the submitted score. Mass, variance, and top-1 label rate are measured before answer generation on the original rendered input.}}
\label{tab:original-molmo}
\begin{tabular}{@{}lllrrrr@{}}
\toprule
Evidence & Negative & Interface & Accuracy (\%) & A/B mass (\%) & Margin variance & A/B top 1 (\%) \\
\midrule
Ordered & Denial & Direct & 30.63 & 99.993 & 4.39 & 100.0 \\
Ordered & Denial & CoT & 37.00 & 0.352 & 1.25 & 0.0 \\
Ordered & Content & Direct & 57.50 & 99.991 & 2.72 & 100.0 \\
Ordered & Content & CoT & 52.12 & 0.244 & 0.789 & 0.0 \\
Shuffled & Denial & Direct & 28.88 & 99.993 & 4.45 & 100.0 \\
Shuffled & Denial & CoT & 30.50 & 0.370 & 1.19 & 0.0 \\
Shuffled & Content & Direct & 56.75 & 99.990 & 2.45 & 100.0 \\
Shuffled & Content & CoT & 49.88 & 0.254 & 0.765 & 0.0 \\
No video & Denial & Direct & 40.62 & 99.962 & 5.21 & 100.0 \\
No video & Denial & CoT & 27.00 & 3.831 & 2.25 & 0.0 \\
No video & Content & Direct & 53.75 & 99.951 & 1.76 & 100.0 \\
No video & Content & CoT & 44.38 & 2.952 & 0.602 & 0.0 \\
\bottomrule
\end{tabular}
\end{table}
\FloatBarrier
}

\begin{document}
\maketitle

\begin{abstract}
\revcolor
\ifdefined\ARXIVVERSION\else
\looseness=-1\relax
\spaceskip=\fontdimen2\font plus \fontdimen3\font minus 1.1pt\relax
\fi
Video verifiers judge whether a candidate answer or trajectory agrees with a video, and are
themselves chosen by a score. We show that this score depends on two choices in the test, the
type of negative and the judgment interface, strongly enough to reverse conclusions about what a
model can do and which verifier to deploy. We separate the two with controlled constructions
that have exact truth conditions: endpoint-matched negatives, wrong only in their internal
states, and a truth-switch test, in which the same two candidates swap correctness between two
videos. Under direct judgment, where the verdict is read from answer scores, models that reject
a visibly wrong start fall to near chance against endpoint-matched negatives. When the frames
disclose each event, writing out reasoning before the answer lifts three of four models from
under 10\% to 70--100\% on the truth-switch test, and staged inputs trace the direct failure to
one step, composing events into states. But when the event must first be found, on
human-validated natural video, the same change brings no significant gain under a protocol
common to all models and can lower accuracy if one candidate denies that the event occurs. The
negative type also changes which verifier is selected: one chosen on event-denial negatives is
8.9 points less accurate against negatives that keep the event and change a detail, the errors a
generator mostly makes, and recovers about half as much accuracy when it reranks that
generator's answers. A verifier should be evaluated on the errors it will face, and neither a
low direct score nor a gain from reasoning should be read as a property of the model alone.
\end{abstract}

\section{Introduction}
\label{sec:introduction}

\begin{figure*}[t]
  \centering
  \includegraphics[width=\linewidth]{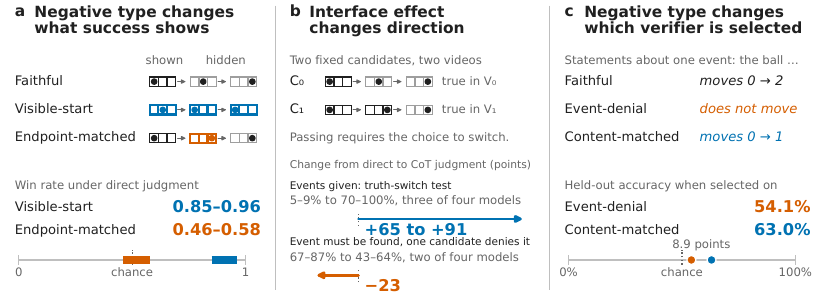}
  \caption{\rev{\textbf{What can a verifier score establish?} The same frozen verifiers support
  different conclusions under different evaluation designs. (a) Colored states differ from
  the faithful trajectory, and only the first state is shown in the video (\cref{sec:construction}).
  (b) Direct to CoT judgment on the truth-switch test and on natural video
(\cref{sec:truth-switch,sec:natural-elicitation}). (c) Accuracy on held-out content-matched
  negatives (\cref{sec:selection}). The statements are schematic.}}
  \label{fig:overview}
\end{figure*}

A video verifier judges whether a candidate answer or trajectory agrees with a
video. Using such a model to evaluate or select outputs
\citep{cobbe2021verifiers,lightman2023verify,gao2023scaling} raises two questions: what does a
verifier's score establish about its abilities (Q1), and does the evaluation identify a useful
verifier for the intended task (Q2)? A strong video
model is not automatically a reliable evaluator \citep{liu2025videojudge}. We
study how the negative type and the judgment interface lead to different
conclusions about the same frozen models (\cref{fig:overview}).

Consider a verifier checking the trajectory of an object hidden under shuffled
cups. A negative (incorrect) trajectory may begin under the wrong cup but
otherwise follow the task's rules; the verifier can reject this
\emph{visible-start} negative from the initial reveal alone. A negative can
instead begin and end under the correct cups and differ only in between. This
\emph{endpoint-matched} negative is also legal, since another sequence of
allowed events would produce it, and rejecting it requires the internal
evidence rather than an incorrect endpoint. A high score against the first type
of negative says little about the second.

Poor performance on endpoint-matched negatives is open to two readings: the
model cannot make the judgment, or the interface fails to elicit an ability the
model has. \rev{We compare \emph{direct judgment}, which reads the verdict from the scores of the two answer tokens
without generating text, with \emph{CoT judgment}, in which the model writes a rationale before its
final answer (\cref{sec:outcomes}). CoT judgment is our probe of the interface, not a recommended verifier.} Higher accuracy with
CoT judgment alone does not show that the new decisions follow the visual
evidence. We therefore pair videos that share their endpoints and both
candidate trajectories but have opposite correct answers. Within a candidate
order the text input is identical, so a verifier answers both videos correctly
only if its choice changes with the internal evidence.

\rev{On natural video the two questions take a concrete form: does the interface effect, the change in a score from direct to CoT judgment, keep its direction when the event must first be found (Q1), and does a verifier chosen on one negative
type hold for another (Q2)? We test four models on 400 TOC-Bench videos against two negative
types, event-denial and content-matched (illustrated in \cref{fig:overview}c), under both
interfaces. Two independent annotators validate the panel, 400 Vinoground pairs
\citep{zhang2024vinoground} form a second panel, and two further models join both.}

We show what a low score means, which way the interface effect points, and what a mismatched
test costs. Three findings answer Q1 and a fourth answers Q2:
{\setlength{\leftmargini}{1.6em}
\begin{itemize}
\item \textbf{Negative type changes what success shows.} \rev{We construct legal endpoint-matched negatives
that separate success on visible-start negatives from verification of the internal
trajectory: under direct judgment, win rates fall from 0.85--0.96 against visible-start
negatives to 0.46--0.58, near chance (0.50), against endpoint-matched ones. The drop
persists when the events are supplied as text (\cref{sec:construction}).}
\item \textbf{The interface changes the capability diagnosis.} \rev{We introduce a truth-switch test over frames that disclose each event: CoT judgment raises
paired success from 5--9\% to 70--100\% for Qwen, GLM, and InternVL, while Molmo reaches only
the 25\% that guessing gives. Staged inputs trace the direct failure to state updating, and
controls attribute the gain to written reasoning (\cref{sec:truth-switch}).}
\item \textbf{The interface effect depends on what the judgment demands.} \rev{With the events given, CoT judgment raises accuracy even against event-denial negatives. On a
human-validated natural panel, where the event must first be found, it yields no significant
accuracy gain for any model under a common protocol (\cref{sec:outcomes}) and lowers event-denial accuracy for two of four. A second panel shows the same pattern (\cref{sec:natural}).}
\item \textbf{Negative type changes which verifier is selected.} \rev{Model comparisons depend on the negative type. When content-matched negatives are the target, choosing among four models under both
interfaces by event-denial accuracy costs 8.9 points of held-out accuracy, and the
verifier so chosen recovers about half as much accuracy when it reranks a generator's
answers (\cref{sec:selection}).}
\end{itemize}}

\section{Audit design}
\label{sec:design}

\subsection{Negatives, interfaces, and evidence controls}
\label{sec:candidates}

\rev{We vary two design choices, the negative type and the judgment interface, across four
studies (\cref{tab:studies}). The remaining
inputs are held fixed within each study.} Evidence controls
(no video, and on natural video also shuffled frames) check whether a model's
answers depend on the frames.

In the controlled tasks an executable program defines the faithful state
trajectory. A trajectory is \emph{legal} if some allowed event sequence
produces it and \emph{faithful} if the events in the presented video produce
it. Shell tracks a hidden object among three cups. Toggle tracks three binary
switches. A bounded-counter task serves as an extension.
\rev{The construction study contains 324 program units, each a faithful program with two
endpoint-matched alternatives (\cref{app:rendering}), and the truth-switch study uses 60 program
pairs.}

The natural-video and selection analyses use 400 TOC-Bench videos
\citep{chen2026tocbench}: 120 event-order, 100 event-relative-state, 100
temporal-location, and 80 visibility-continuity cases. For each video, a
reference proposition was drafted with an LLM from the video and question
and verified by one of the authors against the full video
(\cref{app:provenance}). Two non-author annotators then independently
validated every reference and negative (\cref{sec:natural-design}).
Each question refers to an \emph{anchor event}. The two negative types pair
the faithful proposition with either an \emph{event-denial} negative, which
states that the anchor event never occurs, or a \emph{content-matched} negative, which keeps the anchor event and replaces the queried detail
with a neighboring value under a fixed rule (\cref{app:provenance}). Confirming that the anchor event occurs is
enough to reject the first. Rejecting the second requires checking the
queried detail.

The truth-switch and natural-video studies evaluate Qwen3-VL-32B
\citep{bai2025qwen3vl}, GLM-4.5V \citep{hong2025glm45v}, InternVL3.5-38B
\citep{wang2025internvl35}, and Molmo2-8B \citep{clark2026molmo2}. The
construction study adds Kimi-VL-A3B \citep{moonshot2025kimivl} and Aria
\citep{li2024aria}. A second panel of 400 Vinoground pairs
\citep{zhang2024vinoground} repeats the natural-video analyses, and Kimi-VL-A3B
and Aria join both natural panels as an extension
(\cref{app:revision-transfer}).

\begin{table}[t]
\caption{\rev{\textbf{The four studies.} Units are the bootstrap resampling units (\cref{sec:outcomes}).}}
\label{tab:studies}
\centering\fontsize{8.25}{9.5}\selectfont\revcolor
\renewcommand{\arraystretch}{1.12}
\setlength{\tabcolsep}{1.9pt}
\begin{tabular*}{\linewidth}{@{\extracolsep{\fill}}lclr@{\extracolsep{0pt}\hspace{0.4em}}l@{\extracolsep{\fill}}cll@{}}
\toprule
Study & Q. & Sec. & \multicolumn{2}{l}{Units} & Models & Comparison & Outcome \\
\midrule
Construction       & Q1 & \ref{sec:construction}       & 324 & program units    & 6 & Negative types (direct only)       & Win rate \\
Truth-switch       & Q1 & \ref{sec:truth-switch}       &  60 & program pairs    & 4 & Direct vs.\ CoT judgment           & Paired success \\
\hspace{.45em}event-denial control & Q1 & \ref{sec:natural-elicitation} & 60 & programs & 4 & Negative types $\times$ interfaces & Accuracy \\
Natural video      & Q1 & \ref{sec:natural}            & 400 & TOC-Bench videos & 4 & Negative types $\times$ interfaces & Accuracy \\
\hspace{.45em}second panel & Q1, Q2 & \ref{sec:natural-elicitation}& 400 & Vinoground pairs & 4 & Negative types $\times$ interfaces & Accuracy \\
Verifier selection & Q2 & \ref{sec:selection}         & 400 & TOC-Bench videos & 4 & Selection on each negative type    & Held-out accuracy \\
\hspace{.45em}reranking    & Q2 & \ref{sec:selection}         & 400 & TOC-Bench videos & 4 & Verifier selected on each type     & Top-1 accuracy \\
\bottomrule
\end{tabular*}
\end{table}

\subsection{Outcomes and inference}
\label{sec:outcomes}

The construction study uses direct judgment only and scores each candidate on its own:
a candidate's score is the log-odds that it matches the evidence, and the win rate is the
share of faithful--negative comparisons in which the faithful trajectory scores higher,
with ties counted as one half (chance 0.50). The other studies present both candidates
together as options A and B. Paired success in the truth-switch study requires correct
answers on both videos of a pair in the same candidate order (\cref{sec:truth-switch}).
A constant answer scores 0\% and independent guessing scores 25\%. Natural-video accuracy averages the two candidate orders
within each video (chance 50\%).

The main comparisons use one common protocol for every model. Direct judgment generates no
text: it compares the probabilities of A and B at the validated answer position, where the
model itself writes its answer letter. We locate this position from each model's own concise
answers; with video, the two labels carry at least 87\% of the probability there for each of
the four models of \cref{sec:truth-switch,sec:natural} (\cref{app:revision-position}). CoT
judgment asks for a brief rationale with an 8,192-token budget and adds a verification
procedure on the controlled tasks only. A fixed rule extracts the last expressed choice.
When it flags an output, an LLM assistant checks for a stated choice. Both steps withhold the
correct answer. The review covers controlled and natural-video runs
(\cref{app:revision-interface,app:revision-natural}). The later review of 40 natural-video
outputs moves no cell by more than 0.50 points. Native thinking, the
built-in reasoning mode of GLM and InternVL, is off. Ties and outputs without a choice count as
incorrect; counting every truncated output as incorrect changes no conclusion
(\cref{app:revision-interface,app:revision-natural}). The rationale analysis and verifier
selection instead keep the per-model protocol of our submission, and the second panel its CoT
settings (\cref{app:checkpoints}).

Uncertainty is computed by bootstrap at the program, program-pair, or video
level, preserving each study's stratification. Simultaneous 95\% intervals
are reported within each declared family of estimates, and we call a contrast
significant when its simultaneous interval excludes zero. Differences between
percentages are in percentage points, written ``points''. \Cref{app:statistics}
gives the families and procedures.

\section{Visible-start success does not show internal-trajectory verification}
\label{sec:construction}

Let a program produce the faithful trajectory $c^+=(s_0,\ldots,s_H)$ through
transitions $s_{t+1}=\tau(s_t,a_t)$. We construct a legal negative
$c^-=(\tilde s_0,\ldots,\tilde s_H)$ satisfying
\begin{equation}
\tilde s_0=s_0,\qquad \tilde s_H=s_H,\qquad
\exists\,t\in\{1,\ldots,H-1\}:\ \tilde s_t\neq s_t.
\label{eq:endpoint-matched}
\end{equation}
Legality requires an allowed event sequence realizing every transition of
$c^-$; the trajectory is nevertheless incorrect for the displayed program. We
compare these \emph{endpoint-matched} negatives with \emph{visible-start}
negatives, legal trajectories whose initial state is visibly incorrect. Two
further negative types give context: an \emph{isolated-start} negative changes
only the initial state, and a \emph{single-deviation} negative is
endpoint-matched with one internal deviation (\cref{app:rendering}).

The rendered input reveals the initial state, depicts each event without
revealing the subsequent hidden state, and covers the final state. All
candidate trajectories share one text schema. For a fixed video, endpoint
matching removes the candidate endpoints as a basis for choosing between
trajectories.

\begin{figure}[t]
  \centering
  \includegraphics[width=\linewidth]{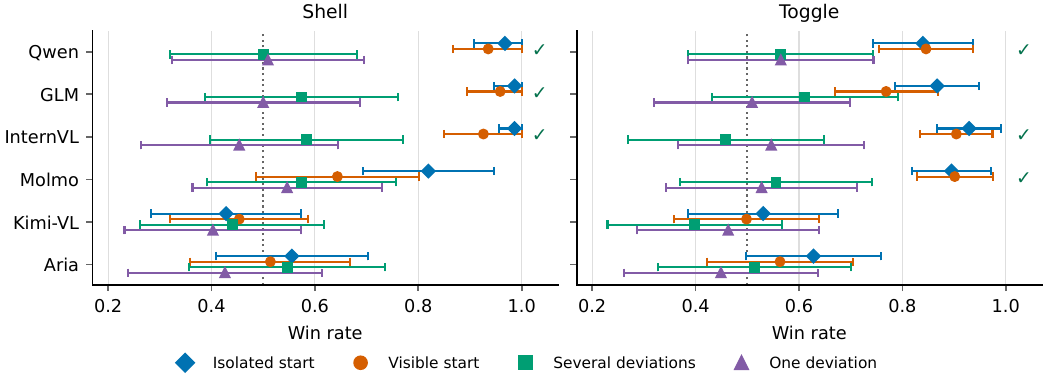}
  \caption{\rev{\textbf{Visible-start versus endpoint-matched win rates.} The two deviation types are endpoint-matched. Bars are simultaneous 95\% intervals, and the dotted line marks chance. Check marks indicate the six qualifying cells.}}
  \label{fig:construction}
\end{figure}

Our main outcome is the \emph{endpoint-matching drop}: the win rate against
visible-start negatives minus the win rate against endpoint-matched negatives.
A cell (one model on one task) \emph{qualifies} when it meets three criteria
fixed before the drop was inspected (\cref{app:scores}): a text-record test, in
which the model receives the true state sequence as text; a visible-start
criterion, which requires a high win rate that depends on the video; and a
drop criterion, which requires a drop that is larger with video than without.

Six cells qualify: Qwen and InternVL on both shell and toggle, GLM on shell, and Molmo on
toggle. In these cells, win rates range from 0.85 to 0.96 against visible-start negatives
and from 0.46 to 0.58 against endpoint-matched ones. Qwen's shell win rate, for example,
falls from 0.935 to 0.500. GLM on toggle falls from 0.77 to 0.61 and misses the drop
criterion. \cref{fig:construction} displays all twelve shell/toggle cells.

Every endpoint-matched interval on shell and toggle includes chance (including GLM toggle, $0.61\;[0.43,0.79]$), though the intervals are too wide to establish equivalence to
chance. Kimi fails the text-record test, Aria fails the visible-start criterion, and no
counter cell qualifies.

\begin{revisedtext}
The drop persists when we supply the events as text. With the frames replaced by the initial
state and the event list as text, endpoint-matched win rates stay at 0.53--0.66 on both
tasks for the four models of \cref{sec:truth-switch}, and the same models verify these trajectories
when given the state sequence (\cref{app:revision-controlled}).
\end{revisedtext}

Success against a visibly wrong start therefore says little about deviations confined to
the internal trajectory. The drop marks a limit of direct judgment against one negative type. \Cref{sec:truth-switch} changes the interface to ask what this limit means.

\section{The interface changes the capability diagnosis}
\label{sec:truth-switch}

\subsection{A fixed-candidate truth-switch test}
\label{sec:truth-switch-design}

A stronger interface could raise scores without making the decisions
\emph{evidence-responsive}, that is, without making the choice follow the video
when only the internal evidence changes. We test this with two executable
witness videos for each endpoint-matched pair of candidates. Video $V_0$
realizes trajectory $C_0$ and video $V_1$ realizes trajectory $C_1$; the
programs have equal initial and final states but different internal
trajectories. Both videos use the construction study's renderer, nuisance
seed, and covered final state, and their decoded initial and final frames
match exactly (\cref{fig:truth-switch} in \cref{app:truth-switch-construction}).

For each order $o\in\{1,2\}$ of the two candidates, each video is evaluated in
a fresh conversation. The text prompt and input token IDs are identical across
the two videos within an order and interface. Let $\hat C(V_j;o)$ be the
candidate selected for video $V_j$. We define
\begin{equation}
B_{p,o}=\mathbf 1\{\hat C(V_0;o)=C_0\ \wedge\ \hat C(V_1;o)=C_1\},\qquad
\widehat B=\frac{1}{60}\sum_{p=1}^{60}\frac{B_{p,1}+B_{p,2}}{2}.
\label{eq:paired-success}
\end{equation}
The \emph{paired success} $\widehat B$ is the rate of answering both videos
correctly, averaged over the two orders of each program pair $p$. The rule follows Vinoground's text score
\citep{zhang2024vinoground}.

\begin{table}[t]
\centering\scriptsize\revcolor
\caption{\rev{\textbf{Truth-switch paired success under the common protocol.} Percentages for 60 pairs, with simultaneous 95\% intervals. ``8 tokens'' is a generated answer under an eight-token limit. ``Score-level'' compares A$-$B margins (chance 50\%). Independent guessing gives 25\% paired success.}}
\label{tab:revision-truth}
\begin{tabular}{@{}lccccc@{}}
\toprule
Model & Direct & 8 tokens & Score-level & CoT & CoT $-$ direct \\
\midrule
Qwen3-VL-32B & $9.17\;[0.00,19.17]$ & 15.83 & 72.50 & $100.00\;[90.00,100.00]$ & $+90.83\;[80.83,100.00]$ \\
GLM-4.5V & $9.17\;[0.00,19.17]$ & 7.50 & 62.50 & $90.00\;[80.00,100.00]$ & $+80.83\;[70.83,90.83]$ \\
InternVL3.5-38B & $5.00\;[0.00,15.00]$ & 5.00 & 58.75 & $70.00\;[60.00,80.00]$ & $+65.00\;[55.00,75.00]$ \\
Molmo2-8B & $1.67\;[0.00,11.67]$ & 1.67 & 54.17 & $25.00\;[15.00,35.00]$ & $+23.33\;[13.33,33.33]$ \\
\bottomrule
\end{tabular}
\end{table}

The 60 pairs were fixed in a pilot, 15 in each shell/toggle by four/six-event cell,
without selection on model correctness. All four models run under the common protocol, giving 240 responses per model and interface
(two videos and two orders per pair). The twelve estimates of \cref{tab:revision-truth} form
one simultaneous family. \cref{app:truth-switch} reports the same test under the per-model
protocol, and \cref{app:revision-reconciliation} reconciles the two sets of numbers.

\subsection{CoT judgments follow the changed evidence}
\label{sec:truth-switch-results}

Under CoT judgment, three of the four models are evidence-responsive: the same candidate
pair receives opposite, correct decisions when the internal evidence changes. Paired
success rises from 9.17\% to 100.00\% for Qwen, from 9.17\% to 90.00\% for GLM, and from
5.00\% to 70.00\% for InternVL (\cref{tab:revision-truth}). Their near-chance direct win rates in \cref{sec:construction}
therefore understate what they can do. Molmo reaches 25.00\%, the rate that independent
guessing gives, and 14 of its 240 CoT responses make no choice.

Single-video accuracy hides the direct failure. Under direct judgment it is 49--53\% for
all four models while paired success is 2--9\%. The models usually give the same answer to both
videos of a pair.

The direct scores themselves carry little of the signal. A score-level test compares the
A-minus-B margins of the two videos within an order, so that a fixed preference scores 50\%.
The margins move in the correct direction in 72.50\% of pair-orders for Qwen and 62.50\% for
GLM, and the pointwise intervals for InternVL and Molmo include chance
(\cref{tab:revision-truth}; \cref{app:truth-switch-margin}).

\subsection{\texorpdfstring{\rev{Where direct judgment fails, and what produces the gain}}{Where direct judgment fails, and what produces the gain}}
\label{sec:truth-switch-controls}

\begin{figure}[t]
\centering
\includegraphics[width=\linewidth]{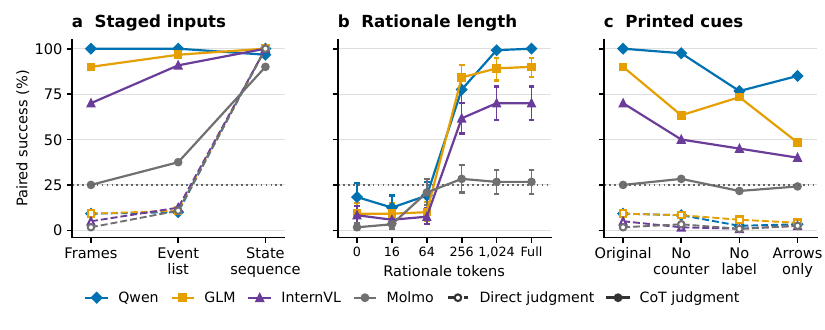}
\caption{\rev{\textbf{Where direct judgment fails, and what produces the gain.} Truth-switch paired success. (a) Frames are replaced by the event list, then by the state sequence. (b) The answer is read after a prefix of the model's own rationale; bars are pointwise 95\% intervals. (c) Printed annotations are removed. The dotted line marks guessing (25\%).}}
\label{fig:revision-mechanism}
\end{figure}

\begin{revisedtext}
\noindent\textbf{Direct judgment fails at one step (\cref{fig:revision-mechanism}a).} We replace the frames with the
verified initial state and event list, and then with the verified state sequence. Under
direct judgment the event list leaves paired success at 10.00--12.50\%, and the state
sequence raises it to 100.00\% for all four models. The failing step is state updating, the composing of events into states, and the drop of
\cref{sec:construction} reflects the same step. Under CoT judgment the
stages separate by model: reading the frames costs InternVL 20.83 points, and state
updating costs Molmo 52.50 (\cref{app:revision-controlled}).

\noindent\textbf{Written reasoning, not generation or wording, produces the gain (\cref{fig:revision-mechanism}b).} The two interfaces also
differ in prompt, generation, and output budget, so we vary each alone (\cref{app:revision-interface}). An
answer generated under an eight-token limit stays at the direct floor (\cref{tab:revision-truth}), dropping the verification procedure from the prompt moves results by at most 7.5 points, and at most one of 240 rationales reaches
the token cap. Reading the answer after a prefix of the model's own rationale shows where
the gain arises: three models stay below 20\% through 64 rationale tokens and reach
62--84\% by 256. InternVL writes about 3,100 tokens even when asked for a concise answer,
and then reaches 95.00\%.

\noindent\textbf{Much of the gain survives the removal of printed labels (\cref{fig:revision-mechanism}c).} Each
event frame prints a text label and the event's position. Without the position counter the two
witness videos differ only in the order of their frames, and Qwen, GLM, and InternVL reach
97.50\%, 63.33\%, and 50.00\%. With arrow position as the only cue they reach 85.00\%, 48.33\%, and
40.00\%, against at most 4.17\% under direct judgment (\cref{app:revision-render}).
These findings establish responsiveness to disclosed event evidence in stylized keyframes,
with tracking of unannotated motion outside their scope.
\end{revisedtext}

\section{Natural video: the interface effect depends on the negative type}
\label{sec:natural}

\subsection{Negative type and interface on natural video}
\label{sec:natural-design}

In \cref{sec:truth-switch} the frames disclosed every event. Natural video does not: the event a
statement refers to must first be found. We ask whether the interface effect keeps its direction
there (Q1). The natural-video study crosses four models, the two negative types, the two interfaces,
and both candidate orders; each comparison uses the same candidate text, media (64, 48,
80, and 48 frames for Qwen, GLM, Molmo, and InternVL), and order. Direct judgment covers
three evidence conditions (ordered, shuffled, and no video); CoT judgment and a generated-answer control (a concise answer without a rationale) cover
ordered video. The design contains 32,000 responses.

For the question ``At the moment the cell
phone first shows up, what is the state of the jars?'' the faithful statement
is ``At the event `the cell phone first shows up', the entity `the jars' is
partly hidden.'', the event-denial negative is ``The anchor event `the cell
phone first shows up' does not occur anywhere in the video.'', and the
content-matched negative is ``At the event `the cell phone first shows up', the
entity `the jars' is fully visible.'' Which type is harder varies by model
(\cref{fig:natural}).

\begin{revisedtext}
\noindent\textbf{Validation.} Two non-author annotators, with no AI pre-labeling, each judged all 1,200 propositions
independently against the full videos, with the three propositions of a video shuffled under
neutral identifiers and no model output shown.
They agree on 98.2\% of truth values ($\kappa = 0.96$), and both confirm the faithful
proposition as true and the negative as false on 383 of 400 videos for each negative
type. They then judged whether the evidence that contradicts each negative is visible in
the frames each model received. The evidence is visible for 90--92\% of propositions,
with agreement of at least 99.5\%. The interface contrasts and the selection cost below hold when each model is restricted to the
videos that both annotators validate and whose evidence both see in that model's frames
(343--351 videos; \cref{app:revision-validation}).
\end{revisedtext}

\subsection{Interface effect by negative type}
\label{sec:natural-elicitation}

\begin{figure}[t]
\centering
\includegraphics[width=\linewidth]{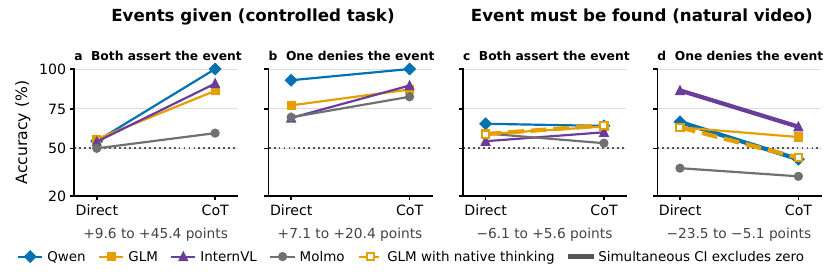}
\caption{\rev{\textbf{The interface effect depends on what the judgment demands.} Accuracy under direct and CoT
judgment with the events given (a, b; controlled task, \cref{app:revision-denial}) and with the
event to be found (c, d; TOC-Bench). Both candidates assert the event in (a, c); one denies it
in (b, d). In (c, d), thick lines mark changes whose simultaneous 95\% interval excludes zero
(\cref{tab:revision-natural}). \cref{tab:revision-controlled-denial} gives pointwise intervals
for (a, b). Dashed lines show GLM with native thinking (a separate family); the dotted line
marks chance.}}
\label{fig:natural}
\end{figure}

Under the common protocol, CoT judgment brings no significant accuracy gain on TOC-Bench, and
its effect depends on the negative type (\cref{fig:natural}c,d; intervals in
\cref{tab:revision-natural}). Against event-denial negatives it lowers accuracy by 23.50 points
for Qwen [$-30.50$, $-16.50$] and by 22.88 points for InternVL [$-29.88$, $-15.88$]. GLM and
Molmo show smaller declines of 6.12 and 5.12 points, with intervals that include zero. For
Qwen, InternVL, and GLM, the change against event-denial negatives is significantly more
negative than the change against content-matched negatives.

\begin{revisedtext}
Three controls locate the decline (\cref{app:revision-natural}). First,
when both answers are generated and only the request for a rationale differs, event-denial
accuracy falls by 18.62 points for Qwen [$-23.12$, $-14.13$] and by 23.00 points for InternVL
[$-26.88$, $-19.12$] (pointwise intervals), while content-matched accuracy changes by $-2.00$ and +3.12 points; for
GLM and Molmo, event-denial accuracy rises slightly (+3.75 and +5.00 points). Second, under one
fixed prompt, InternVL's answer scores 79.88\% when read before its rationale and 61.25\% after
it, whereas Qwen is already at 48.00\% before any rationale token, against 61.75\% for a
concise answer. Third, with GLM's native thinking on, event-denial accuracy is 44.12\% against
63.25\% under direct judgment ($-19.13$ points [$-24.38$, $-13.88$]). Content-matched accuracy
rises by 5.50 points [0.25, 10.75], a significant gain outside the common protocol.
Both intervals are simultaneous within their own three-estimate family.
\end{revisedtext}

\begin{figure}[!ht]
\centering
\includegraphics[width=\linewidth]{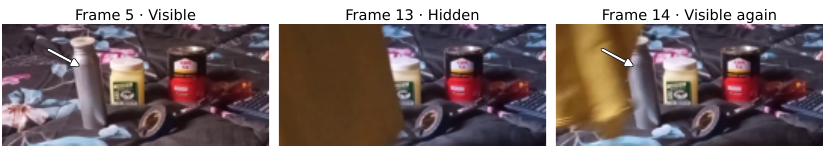}
\par\smallskip
{\footnotesize
\begin{tabularx}{\linewidth}{@{}YY@{}}
\textbf{Qwen, against the event-denial negative} &
\textbf{Qwen, against the content-matched negative} \\
``At no point in the video is the bottle completely hidden from view.'' &
``the bottle is fully hidden behind the yellow cloth. [\ldots] the bottle
becomes visible again starting from Frame 14.'' \\
$\rightarrow$ Chooses the denial $\times$ &
$\rightarrow$ Chooses the faithful statement $\checkmark$ \\
\end{tabularx}\par}
\caption{\rev{\textbf{Same frames, different conclusions.} Qwen's CoT
rationales use the same 64 frames with the faithful candidate second.
The anchor event is the bottle becoming fully hidden. Arrows mark the bottle.
Frames from Perception Test \citep{patraucean2023perception}.}}
\label{fig:bottle-occlusion}
\end{figure}

\noindent\textbf{Why the effect reverses.} The decline arises where an event must be found and one candidate
denies it. On the controlled task, where the events are given, CoT judgment raises accuracy
against an event-denial negative for all four models, by 7.08 to 20.42 points
(\cref{fig:natural}b; \cref{app:revision-denial}). On natural video the event must first be found,
and reasoning does not make this reliable. Asked without candidates, Qwen affirms a
reappearance that is visible in its own frames on only 55.32\% of clips even with a rationale
(31.91\% with a concise answer). A missed event then lands on the
event-denial negative, which acts as a ``none of the above'' option. In the rationale analysis of Qwen's per-model-protocol responses (\cref{app:trace-audit}), 396
of the 440 wrong CoT answers conclude that the anchor event never happens. On 36 of the 54
videos that CoT judgment loses, the same model answers the content-matched pair correctly in
both orders; \cref{fig:bottle-occlusion} shows one such clip. The decline appears in all four
question types for Qwen and InternVL (\cref{tab:revision-natural-by-type}).

\begin{revisedtext}
\noindent\textbf{A second panel.} The pattern repeats on 400 Vinoground pairs
(800 clips), where the content-matched negative is the pair's other caption and the
event-denial negative denies the captioned event. With per-model CoT settings, CoT judgment lowers event-denial accuracy for all four models, by
3.81 to 10.31 points, and changes content-matched accuracy by at most 2.31 points. The
event-denial change is the more negative one for all four models, significantly for three
(pointwise intervals; \cref{tab:transfer-interaction}). With Kimi-VL-A3B and Aria added, it is significantly more negative for four of six models on
TOC-Bench and three of six on Vinoground, again by pointwise intervals. Molmo's difference is positive on TOC-Bench,
and Aria shows none (\cref{app:revision-transfer}). The paired
captions also give a natural counterpart of the truth-switch test. There, direct paired
success is 56--64\% for three models, far above the 25\% that guessing gives, and CoT
judgment changes it by $-1.75$ to $+5.87$ points.
\end{revisedtext}

\subsection{Large model gaps need not survive a change of negative type}
\label{sec:natural-comparisons}

Model comparisons under CoT judgment depend on the negative type. For each of the six model
pairs we compare the accuracy gap against event-denial negatives with the gap against
content-matched negatives; five of the six changes are significant
(\cref{tab:natural-primary}). Against event-denial negatives, InternVL scores 63.62\% and Qwen
scores 43.12\%, a Qwen-minus-InternVL gap of $-20.50$ points [$-29.00$, $-12.00$]. Against
content-matched negatives they score 60.12\% and 64.12\% (\cref{fig:natural}), a gap of +4.00
points [$-4.50$, 12.50]. The change is $-24.50$ points [$-33.00$, $-16.00$]. Under CoT judgment
no model pair has significant gaps of opposite sign. Under direct judgment the order of Qwen
and InternVL reverses (\cref{tab:revision-evidence}), which drives the selection result of
\cref{sec:selection}.

\subsection{Evidence controls}
\label{sec:natural-evidence}

Frame order matters little under direct judgment: shuffling the frames changes accuracy
by at most 5.62 points in any cell. The effect of removing video varies by model. Qwen's
accuracy against event-denial negatives falls from 66.62\% to 8.25\%, while InternVL's falls only from 86.50\% to 74.62\% (81.62\% to 66.75\% at the readout position
of our submission; \cref{app:revision-position}). Most of InternVL's accuracy against
event-denial negatives is therefore available without the video (Q1), which bears on the
selection result below. \cref{app:natural} reports all cells and output diagnostics, with the
CoT evidence contrasts under the per-model protocol.

\subsection{Negative type changes which verifier is selected}
\label{sec:selection}

Changed model gaps matter in practice only relative to a target (Q2). We therefore
evaluate verifier selection with a declared target, distinguishing faithful
propositions from content-matched negatives on the same 400 videos. The analysis was specified after the cell accuracies in
\cref{app:natural} were known; its rules, eligible sets, and partitions were fixed before any
selection outcome was computed. The analysis keeps the eight per-model configurations with their direct
scores as submitted, so accuracies here differ slightly from those above
(\cref{app:revision-reconciliation}).

A \emph{configuration} is one model under one interface. Within an eligible
set of configurations, the \emph{denial-selected} policy picks the
configuration with the highest event-denial accuracy on the selection videos,
and the \emph{content-matched-selected} policy picks the one with the highest
content-matched accuracy (\cref{eq:selection} in \cref{app:selection}); both
are evaluated on content-matched negatives from held-out videos. The
\emph{selection cost} is the held-out accuracy of the content-matched-selected
configuration minus that of the denial-selected one. We use twenty stratified
five-fold partitions, with 320 selection and 80 evaluation videos per fold,
and three eligible sets: all eight configurations, the four direct ones, and
the four CoT ones. Whole-video bootstrap draws repeat the selection in each
draw, and the three selection costs form one simultaneous family
(\cref{tab:selection}).

\begin{table}[!htb]
  \caption{\rev{\textbf{Selecting on the wrong negative type reduces accuracy on the declared target.} Held-out content-matched accuracy of the configuration each policy selects. Selection costs carry simultaneous 95\% intervals.}}
  \label{tab:selection}
  \centering
  \small
  \begin{tabular*}{\linewidth}{@{\extracolsep{\fill}}lccc@{}}
\toprule
Eligible set & Denial-selected (\%) & Content-matched-selected (\%) & Selection cost (pp) \\
\midrule
All eight & $54.13$ & $63.01$ & $+8.88\;[2.46,\,15.30]$ \\
Four direct & $54.13$ & $64.38$ & $+10.25\;[3.83,\,16.67]$ \\
Four CoT & $59.38$ & $63.88$ & $+4.50\;[-1.92,\,10.92]$ \\
\bottomrule
\end{tabular*}

\end{table}

Across all eight configurations, the denial-selected policy always chooses direct InternVL,
which keeps most of its event-denial accuracy without the video (\cref{sec:natural-evidence}).
The content-matched-selected policy chooses direct Qwen in 70\% of folds and CoT Qwen in the rest. The within-panel selection cost is 8.88 points [2.46, 15.30]; among the four CoT configurations
alone it is not significant (\cref{tab:selection}; \cref{app:selection}).
On event-denial comparisons the denial-selected configuration scores 81.88\% versus
55.82\% for the content-matched-selected configuration.

\begin{revisedtext}
\noindent\textbf{Robustness.} The cost is positive across splits, configurations, validated videos, and panels, though not
always significant (\cref{app:revision-selection}). It is positive in all 21 splits of each
eligible set; on the separately fixed 200/200 split the all-eight cost of 4.75 points is not
significant, inside an interval about 13 points wide. With 26 eligible configurations (the
eight above, validated direct judgment, additional generation settings, and two added models)
the cost is 8.39 points [1.98, 13.58]. With eight common-protocol configurations it is
9.36 points [4.17, 14.50] (pointwise; \cref{app:revision-selection}). On the videos that both annotators validate, it is
9.21--10.70 points. With validated direct scores, selecting on the second panel and evaluating
on TOC-Bench costs 8.88 points [4.75, 14.75]; the reverse direction costs 3.63 points [0.16, 5.66] (pointwise intervals). On both panels the
event-denial criterion selects direct InternVL.

\noindent\textbf{A realistic target.} Content-matched negatives resemble the errors a generator
makes. Qwen2.5-VL-7B, a model outside the verifier panel, answered the panel's questions.
Of 100 of its wrong answers labeled by hand, 73 are content errors, 3 deny the event, and
24 are errors of other kinds. We weight the two negative types using the LLM grader's labels
for the full generator pool. Selection on this mixture scores 7.45 points higher on the
weighted accuracy than event-denial selection [2.47, 12.24] (\cref{app:revision-selection}).

\noindent\textbf{Reranking.} Selecting on event denials also reduces the reranking gain. The same generator wrote five
candidate answers per video, and each verifier, under direct judgment, kept its
highest-scored candidate (\cref{tab:revision-rerank}). Direct Qwen, the choice of content-matched selection
among the direct configurations, reaches 53.25--53.50\% top-1 accuracy. Direct InternVL,
the choice of event-denial selection, reaches 47.79--48.54\%. Depending on how the few candidates of unknown correctness are counted, the difference is 4.96
points [1.29, 8.71] or 5.46 points [1.79, 9.21], and it is 5.65 points [2.06, 9.29] with
human-audited labels (paired whole-video bootstrap; \cref{app:revision-rerank}). Over the
44.00\% of the unreranked first sample, direct Qwen gains 9--10 points and direct InternVL
4--5, about half as much.
\end{revisedtext}

\begin{table}[t]
\centering\footnotesize\revcolor
\caption{\rev{\textbf{Best-of-five reranking on the 400 TOC-Bench videos.} Top-1 accuracy of each reranker under direct judgment. Ranges are bounds over candidates whose correctness is unknown.}}
\label{tab:revision-rerank}
\begin{tabular}{@{}lc@{}}
\toprule
Reranker & Top-1 accuracy (\%) \\
\midrule
First sample, no reranking & 44.00 \\
Random choice among the five & 43.15--43.45 \\
Molmo2-8B & 46.50 \\
InternVL3.5-38B (selected on event-denial negatives) & 47.79--48.54 \\
GLM-4.5V & 50.50--50.75 \\
Qwen3-VL-32B (selected on content-matched negatives) & 53.25--53.50 \\
Oracle: any of the five correct & 72.25--72.50 \\
\bottomrule
\end{tabular}
\end{table}
\section{Related work}
\label{sec:related}

Hard negatives, paired counterfactuals, and judge biases are known concerns; we measure them
where the truth conditions are exact. Prior paired benchmarks perturb captions or match natural
clips
\citep{gardner2020contrast,thrush2022winoground,yuksekgonul2023aro,hsieh2023sugarcrepe,bansal2024videocon,saravanan2025velociti,zhang2024vinoground}
and show that a model fails, not why. We change the evidence under fixed text, with executable
truth conditions, so the inputs can be staged, which places the direct failure in state
updating. Lower accuracy under chain-of-thought prompting has been reported
\citep{saravanan2025velociti}; crossing the interface with the negative type shows when it
occurs. Judge-reliability studies motivate checking readout position and generated choices
\citep{zheng2023judging,wang2024answerc,chen2024judgement}; we add the cost of a mismatched
negative type in selection and reranking. VET-Bench and VSTAT evaluate solvers of
state-tracking problems \citep{liu2026shell,yu2026vstat}; our verifiers check supplied
trajectories against disclosed events. \cref{tab:revision-related} details the closest
comparisons.

\section{Discussion and conclusion}
\label{sec:discussion}

\noindent\textbf{Answers.} Q1: a verifier's score establishes success against the tested negative type under the tested
interface, and no more. It does not show the ability another negative type requires. A low
direct score does not show missing ability: three of four models verify an internal trajectory
over disclosed events once they write out their reasoning, and the missing step is state
updating. Nor is a gain from reasoning a property of the model: reasoning helps where the
events are given, and where the event must first be found it brings no significant gain under
the common protocol and lowers event-denial accuracy for two of four models. Q2: whether the
evaluation identifies a useful verifier depends on whether its negatives match the errors the
verifier will face. Selecting on event-denial negatives costs 8.9 points against
content-matched negatives, a generator's most common error, and about half of the reranking
gain.

\noindent\textbf{Limitations.} The controlled tasks disclose each event and fix object identity and slot positions, so they do not
cover events that must be recovered from unannotated motion.
Native thinking is off in the common protocol; with it on, GLM and InternVL reach 100\% paired
success in a 16-pair probe and GLM's event-denial decline on natural video deepens
(\cref{app:revision-interface,app:revision-natural}). The natural-video findings cover two
panels and six mid-sized open models. Selection uses a declared content-matched target, which
the generator's errors support although about a quarter of them fall outside both negative
types. Reranking uses one generator, one panel, and a scoring prompt that differs from the
verification prompt. The study measures offline judgments of frozen models.

\noindent\textbf{Practices.} Evaluate a verifier on the negative type it will face, re-test a low direct score with
reasoning, re-measure an interface effect on the target, and report direct scores with their
label probability, since a low one can produce an artificial floor (\cref{app:revision-position}).

\label{sec:discussion-end}

\clearpage
\section*{Reproducibility statement}

Appendices A--H give the cohorts, checkpoint revisions, negative constructions, prompts,
answer-extraction rules, annotation protocol, and statistical procedures. An anonymous
artifact with the task generators, prompts, per-response outputs, annotation labels, and
analysis scripts accompanies the submission; the full release, including the
truth-switch, selection, and reranking packages, will be published with the camera-ready
version. The artifact identifies the model checkpoints and the TOC-Bench and
Vinoground source videos.

\section*{AI use statement}

An LLM drafted the reference propositions for the natural-video panel. One author verified
every proposition against its full video (\cref{app:provenance}). An LLM assistant coded the
rationales and read the validation sample under a fixed codebook. An LLM assistant also
proposed the frame-level judgments, which one author reviewed item by item
(\cref{app:trace-audit}). A reference-conditioned LLM grader labeled the reranking candidates
for correctness and error type (\cref{app:revision-rerank}). An LLM assistant and a separate Qwen3-32B classifier checked
verifier outputs flagged by the extraction rule, with the correct answers withheld
(\cref{app:revision-interface,app:revision-natural,app:revision-transfer}).

LLM assistants helped write analysis and experiment code, run and monitor experiments, and
draft and edit the manuscript, including the abstract, introduction, and discussion. They
provided feedback on the design of the additional controls and the interpretation of the
results. The authors checked the reported numbers, made the final decisions, and are
responsible for the manuscript.

\section*{Ethics statement}

The natural-video studies use TOC-Bench and Vinoground videos under their documented
licenses. \Cref{fig:bottle-occlusion} reproduces three cropped Perception Test
frames under \href{https://creativecommons.org/licenses/by/4.0/}{CC BY 4.0}.
Human involvement comprised the authors'
verification of reference propositions and review of the frame-level check, and
annotation by people who are not authors: the validation study of \cref{app:revision-validation}, the checks
of the second panel's event-denial statements, and the audit of reranking labels.
No personal data were collected.

\bibliographystyle{iclr2027_conference}
\bibliography{references}

\begin{thebibliography}{56}
\providecommand{\natexlab}[1]{#1}
\providecommand{\url}[1]{\texttt{#1}}
\expandafter\ifx\csname urlstyle\endcsname\relax
  \providecommand{\doi}[1]{doi: #1}\else
  \providecommand{\doi}{doi: \begingroup \urlstyle{rm}\Url}\fi

\bibitem[Alzahrani et~al.(2024)Alzahrani, Alyahya, Alnumay, AlRashed, Alsubaie,
  Almushayqih, Mirza, Alotaibi, Al-Twairesh, Alowisheq, Bari, and
  Khan]{alzahrani2024benchmarks}
Norah Alzahrani, Hisham Alyahya, Yazeed Alnumay, Sultan AlRashed, Shaykhah
  Alsubaie, Yousef Almushayqih, Faisal Mirza, Nouf Alotaibi, Nora Al-Twairesh,
  Areeb Alowisheq, M.~Saiful Bari, and Haidar Khan.
\newblock When benchmarks are targets: Revealing the sensitivity of large
  language model leaderboards.
\newblock In \emph{Proceedings of the 62nd Annual Meeting of the Association
  for Computational Linguistics (Volume 1: Long Papers)}, pp.\  13787--13805,
  Bangkok, Thailand, 2024. Association for Computational Linguistics.
\newblock \doi{10.18653/v1/2024.acl-long.744}.
\newblock URL \url{https://aclanthology.org/2024.acl-long.744/}.

\bibitem[Bai et~al.(2025)Bai, Cai, Chen, et~al.]{bai2025qwen3vl}
Shuai Bai, Yuxuan Cai, Ruizhe Chen, et~al.
\newblock {Qwen3-VL} technical report.
\newblock \emph{arXiv preprint arXiv:2511.21631}, 2025.
\newblock \doi{10.48550/arXiv.2511.21631}.
\newblock URL \url{https://arxiv.org/abs/2511.21631}.

\bibitem[Bansal et~al.(2024)Bansal, Bitton, Szpektor, Chang, and
  Grover]{bansal2024videocon}
Hritik Bansal, Yonatan Bitton, Idan Szpektor, Kai-Wei Chang, and Aditya Grover.
\newblock {VideoCon}: Robust video-language alignment via contrast captions.
\newblock In \emph{Proceedings of the IEEE/CVF Conference on Computer Vision
  and Pattern Recognition}, pp.\  13927--13937, 2024.
\newblock URL
  \url{https://openaccess.thecvf.com/content/CVPR2024/html/Bansal_VideoCon_Robust_Video-Language_Alignment_via_Contrast_Captions_CVPR_2024_paper.html}.

\bibitem[Bates et~al.(2024)Bates, Hastie, and
  Tibshirani]{bates2023crossvalidation}
Stephen Bates, Trevor Hastie, and Robert Tibshirani.
\newblock Cross-validation: What does it estimate and how well does it do it?
\newblock \emph{Journal of the American Statistical Association}, 119\penalty0
  (546):\penalty0 1434--1445, 2024.
\newblock \doi{10.1080/01621459.2023.2197686}.
\newblock URL \url{https://arxiv.org/abs/2104.00673}.

\bibitem[Buch et~al.(2022)Buch, Eyzaguirre, Gaidon, Wu, Fei-Fei, and
  Niebles]{buch2022atp}
Shyamal Buch, Crist{\'o}bal Eyzaguirre, Adrien Gaidon, Jiajun Wu, Li~Fei-Fei,
  and Juan~Carlos Niebles.
\newblock Revisiting the ``video'' in video-language understanding.
\newblock In \emph{Proceedings of the IEEE/CVF Conference on Computer Vision
  and Pattern Recognition}, pp.\  2917--2927, 2022.

\bibitem[Chen et~al.(2024{\natexlab{a}})Chen, Chen, Zhang, Wang, Liu, Zhou,
  Zhang, Wan, Zhou, and Sun]{chen2024mllmjudge}
Dongping Chen, Ruoxi Chen, Shilin Zhang, Yaochen Wang, Yinuo Liu, Huichi Zhou,
  Qihui Zhang, Yao Wan, Pan Zhou, and Lichao Sun.
\newblock {MLLM-as-a-Judge}: Assessing multimodal {LLM}-as-a-judge with
  vision-language benchmark.
\newblock In \emph{Proceedings of the 41st International Conference on Machine
  Learning}, volume 235 of \emph{Proceedings of Machine Learning Research},
  pp.\  6562--6595. PMLR, 2024{\natexlab{a}}.
\newblock URL \url{https://proceedings.mlr.press/v235/chen24h.html}.

\bibitem[Chen et~al.(2024{\natexlab{b}})Chen, Chen, Liu, Jiang, and
  Wang]{chen2024judgement}
Guiming~Hardy Chen, Shunian Chen, Ziche Liu, Feng Jiang, and Benyou Wang.
\newblock Humans or {LLM}s as the judge? {A} study on judgement bias.
\newblock In \emph{Proceedings of the 2024 Conference on Empirical Methods in
  Natural Language Processing}, pp.\  8301--8327, Miami, Florida, USA,
  2024{\natexlab{b}}. Association for Computational Linguistics.
\newblock \doi{10.18653/v1/2024.emnlp-main.474}.
\newblock URL \url{https://aclanthology.org/2024.emnlp-main.474/}.

\bibitem[Chen et~al.(2026)Chen, Meng, Chen, Zhao, Gui, and
  Guo]{chen2026tocbench}
Junzhe Chen, Siyuan Meng, Yuxi Chen, Man Zhao, Wenyao Gui, and Xiaojie Guo.
\newblock {TOC-Bench}: {A} temporal object consistency benchmark for video
  large language models.
\newblock \emph{arXiv preprint arXiv:2605.09904}, 2026.
\newblock \doi{10.48550/arXiv.2605.09904}.
\newblock URL \url{https://arxiv.org/abs/2605.09904}.

\bibitem[Clark et~al.(2026)Clark, Zhang, Ma, Park, et~al.]{clark2026molmo2}
Christopher Clark, Jieyu Zhang, Zixian Ma, Jae~Sung Park, et~al.
\newblock {Molmo2}: Open weights and data for vision-language models with video
  understanding and grounding.
\newblock \emph{arXiv preprint arXiv:2601.10611}, 2026.
\newblock \doi{10.48550/arXiv.2601.10611}.
\newblock URL \url{https://arxiv.org/abs/2601.10611}.

\bibitem[Cobbe et~al.(2021)Cobbe, Kosaraju, Bavarian, Chen, Jun, Kaiser,
  Plappert, Tworek, Hilton, Nakano, Hesse, and Schulman]{cobbe2021verifiers}
Karl Cobbe, Vineet Kosaraju, Mohammad Bavarian, Mark Chen, Heewoo Jun, Lukasz
  Kaiser, Matthias Plappert, Jerry Tworek, Jacob Hilton, Reiichiro Nakano,
  Christopher Hesse, and John Schulman.
\newblock Training verifiers to solve math word problems.
\newblock \emph{arXiv preprint arXiv:2110.14168}, 2021.
\newblock \doi{10.48550/arXiv.2110.14168}.
\newblock URL \url{https://arxiv.org/abs/2110.14168}.

\bibitem[Cronbach \& Meehl(1955)Cronbach and Meehl]{cronbach1955construct}
Lee~J. Cronbach and Paul~E. Meehl.
\newblock Construct validity in psychological tests.
\newblock \emph{Psychological Bulletin}, 52\penalty0 (4):\penalty0 281--302,
  1955.
\newblock \doi{10.1037/h0040957}.

\bibitem[Ding et~al.(2023)Ding, Liu, He, Jiang, Torr, and Bai]{ding2023mose}
Henghui Ding, Chang Liu, Shuting He, Xudong Jiang, Philip H.~S. Torr, and Song
  Bai.
\newblock {MOSE}: A new dataset for video object segmentation in complex
  scenes.
\newblock In \emph{Proceedings of the IEEE/CVF International Conference on
  Computer Vision}, pp.\  20224--20234, 2023.
\newblock URL
  \url{https://openaccess.thecvf.com/content/ICCV2023/html/Ding_MOSE_A_New_Dataset_for_Video_Object_Segmentation_in_Complex_ICCV_2023_paper.html}.

\bibitem[Gao et~al.(2023)Gao, Schulman, and Hilton]{gao2023scaling}
Leo Gao, John Schulman, and Jacob Hilton.
\newblock Scaling laws for reward model overoptimization.
\newblock In \emph{Proceedings of the 40th International Conference on Machine
  Learning}, volume 202 of \emph{Proceedings of Machine Learning Research},
  pp.\  10835--10866. PMLR, 2023.
\newblock URL \url{https://proceedings.mlr.press/v202/gao23h.html}.

\bibitem[Gardner et~al.(2020)Gardner, Artzi, Basmov, Berant, Bogin, Chen,
  Dasigi, Dua, Elazar, Gottumukkala, Gupta, Hajishirzi, Ilharco, Khashabi, Lin,
  Liu, Liu, Mulcaire, Ning, Singh, Smith, Subramanian, Tsarfaty, Wallace,
  Zhang, and Zhou]{gardner2020contrast}
Matt Gardner, Yoav Artzi, Victoria Basmov, Jonathan Berant, Ben Bogin, Sihao
  Chen, Pradeep Dasigi, Dheeru Dua, Yanai Elazar, Ananth Gottumukkala, Nitish
  Gupta, Hannaneh Hajishirzi, Gabriel Ilharco, Daniel Khashabi, Kevin Lin,
  Jiangming Liu, Nelson~F. Liu, Phoebe Mulcaire, Qiang Ning, Sameer Singh,
  Noah~A. Smith, Sanjay Subramanian, Reut Tsarfaty, Eric Wallace, Ally Zhang,
  and Ben Zhou.
\newblock Evaluating models' local decision boundaries via contrast sets.
\newblock In \emph{Findings of the Association for Computational Linguistics:
  EMNLP 2020}, pp.\  1307--1323, 2020.
\newblock \doi{10.18653/v1/2020.findings-emnlp.117}.

\bibitem[Geirhos et~al.(2020)Geirhos, Jacobsen, Michaelis, Zemel, Brendel,
  Bethge, and Wichmann]{geirhos2020shortcut}
Robert Geirhos, J{ö}rn-Henrik Jacobsen, Claudio Michaelis, Richard Zemel,
  Wieland Brendel, Matthias Bethge, and Felix~A. Wichmann.
\newblock Shortcut learning in deep neural networks.
\newblock \emph{Nature Machine Intelligence}, 2:\penalty0 665--673, 2020.
\newblock \doi{10.1038/s42256-020-00257-z}.
\newblock URL \url{https://arxiv.org/abs/2004.07780}.

\bibitem[{GLM-V Team}(2025)]{hong2025glm45v}
{GLM-V Team}.
\newblock {GLM-4.1V-Thinking} and {GLM-4.5V}: {Towards} versatile multimodal
  reasoning with scalable reinforcement learning.
\newblock \emph{arXiv preprint arXiv:2507.01006}, 2025.
\newblock \doi{10.48550/arXiv.2507.01006}.
\newblock URL \url{https://arxiv.org/abs/2507.01006}.

\bibitem[He et~al.(2024)He, Jiang, Zhang, Ku, Soni, Siu, Chen, Chandra, Jiang,
  Arulraj, Wang, Do, Ni, Lyu, Narsupalli, Fan, Lyu, Lin, and
  Chen]{he2024videoscore}
Xuan He, Dongfu Jiang, Ge~Zhang, Max Ku, Achint Soni, Sherman Siu, Haonan Chen,
  Abhranil Chandra, Ziyan Jiang, Aaran Arulraj, Kai Wang, Quy~Duc Do, Yuansheng
  Ni, Bohan Lyu, Yaswanth Narsupalli, Rongqi Fan, Zhiheng Lyu, Bill~Yuchen Lin,
  and Wenhu Chen.
\newblock {VideoScore}: Building automatic metrics to simulate fine-grained
  human feedback for video generation.
\newblock In \emph{Proceedings of the 2024 Conference on Empirical Methods in
  Natural Language Processing}, pp.\  2105--2123, Miami, Florida, USA, 2024.
  Association for Computational Linguistics.
\newblock \doi{10.18653/v1/2024.emnlp-main.127}.
\newblock URL \url{https://aclanthology.org/2024.emnlp-main.127/}.

\bibitem[Hsieh et~al.(2023)Hsieh, Zhang, Ma, Kembhavi, and
  Krishna]{hsieh2023sugarcrepe}
Cheng-Yu Hsieh, Jieyu Zhang, Zixian Ma, Aniruddha Kembhavi, and Ranjay Krishna.
\newblock {SugarCrepe}: Fixing hackable benchmarks for vision-language
  compositionality.
\newblock In \emph{Advances in Neural Information Processing Systems 36
  (Datasets and Benchmarks Track)}, 2023.
\newblock \doi{10.52202/075280-1355}.
\newblock URL
  \url{https://papers.neurips.cc/paper_files/paper/2023/hash/63461de0b4cb760fc498e85b18a7fe81-Abstract-Datasets_and_Benchmarks.html}.

\bibitem[Huang et~al.(2026)Huang, Zhou, Wang, Zhang, et~al.]{huang2026deft}
Zixuan Huang, Yang Zhou, Kaixuan Wang, Guli Zhang, et~al.
\newblock Deferred exposure of future trajectories for verifiable reasoning in
  autonomous driving {VLM}s.
\newblock \emph{arXiv preprint arXiv:2608.01755}, 2026.
\newblock \doi{10.48550/arXiv.2608.01755}.
\newblock URL \url{https://arxiv.org/abs/2608.01755}.

\bibitem[Jacobs \& Wallach(2021)Jacobs and Wallach]{jacobs2021measurement}
Abigail~Z. Jacobs and Hanna Wallach.
\newblock Measurement and fairness.
\newblock In \emph{Proceedings of the 2021 ACM Conference on Fairness,
  Accountability, and Transparency}, pp.\  375--385. Association for Computing
  Machinery, 2021.
\newblock \doi{10.1145/3442188.3445901}.
\newblock URL \url{https://doi.org/10.1145/3442188.3445901}.

\bibitem[Kim et~al.(2026)Kim, Kim, Chung, Oh, et~al.]{kim2026sgpvr}
Hyomin Kim, Junghye Kim, Joanie~Hayoun Chung, Yoonjin Oh, et~al.
\newblock Plan-and-verify video reward reasoning with spatio-temporal scene
  graph grounding.
\newblock \emph{arXiv preprint arXiv:2606.11838}, 2026.
\newblock \doi{10.48550/arXiv.2606.11838}.
\newblock URL \url{https://arxiv.org/abs/2606.11838}.

\bibitem[{Kimi Team} et~al.(2025){Kimi Team}, Du, Yin, Xing,
  et~al.]{moonshot2025kimivl}
{Kimi Team}, Angang Du, Bohong Yin, Bowei Xing, et~al.
\newblock {Kimi-VL} technical report.
\newblock \emph{arXiv preprint arXiv:2504.07491}, 2025.
\newblock \doi{10.48550/arXiv.2504.07491}.
\newblock URL \url{https://arxiv.org/abs/2504.07491}.

\bibitem[Lee(2026)]{lee2026accuracy}
Jae~Joong Lee.
\newblock Accuracy without grounding: {Diagnosing} visual dependency
  dissociation in video {LLM} benchmarks.
\newblock \emph{arXiv preprint arXiv:2607.13305}, 2026.
\newblock URL \url{https://arxiv.org/abs/2607.13305}.

\bibitem[Lee et~al.(2026)Lee, Park, and Im]{lee2026mmjudgebias}
Sua Lee, Sanghee Park, and Jinbae Im.
\newblock {MM-JudgeBias}: A benchmark for evaluating compositional biases in
  {MLLM}-as-a-judge.
\newblock In \emph{Proceedings of the 64th Annual Meeting of the Association
  for Computational Linguistics (Volume 1: Long Papers)}, pp.\  25336--25373,
  San Diego, California, United States, 2026. Association for Computational
  Linguistics.
\newblock \doi{10.18653/v1/2026.acl-long.1162}.
\newblock URL \url{https://aclanthology.org/2026.acl-long.1162/}.

\bibitem[Lei et~al.(2023)Lei, Berg, and Bansal]{lei2023singleframe}
Jie Lei, Tamara Berg, and Mohit Bansal.
\newblock Revealing single frame bias for video-and-language learning.
\newblock In \emph{Proceedings of the 61st Annual Meeting of the Association
  for Computational Linguistics (Volume 1: Long Papers)}, pp.\  487--507, 2023.
\newblock \doi{10.18653/v1/2023.acl-long.29}.

\bibitem[Li et~al.(2024)Li, Liu, Wu, Wang, et~al.]{li2024aria}
Dongxu Li, Yudong Liu, Haoning Wu, Yue Wang, et~al.
\newblock {Aria}: {An} open multimodal native mixture-of-experts model.
\newblock \emph{arXiv preprint arXiv:2410.05993}, 2024.
\newblock \doi{10.48550/arXiv.2410.05993}.
\newblock URL \url{https://arxiv.org/abs/2410.05993}.

\bibitem[Li et~al.(2025)Li, Wei, Xie, Yang, Song, Wang, An, Liu, Li, Lin, Kong,
  and Liu]{li2024vlrewardbench}
Lei Li, Yuancheng Wei, Zhihui Xie, Xuqing Yang, Yifan Song, Peiyi Wang, Chenxin
  An, Tianyu Liu, Sujian Li, Bill~Yuchen Lin, Lingpeng Kong, and Qi~Liu.
\newblock {VL-RewardBench}: {A} challenging benchmark for vision-language
  generative reward models.
\newblock In \emph{Proceedings of the IEEE/CVF Conference on Computer Vision
  and Pattern Recognition}, pp.\  24657--24668, 2025.
\newblock URL
  \url{https://openaccess.thecvf.com/content/CVPR2025/html/Li_VL-RewardBench_A_Challenging_Benchmark_for_Vision-Language_Generative_Reward_Models_CVPR_2025_paper.html}.

\bibitem[Liang et~al.(2026)Liang, Shah, Zhou, Sharan, Goel, Sanyal, Chinchali,
  and Datta]{liang2026rtneus}
Shawn Liang, Sahil Shah, Chengwei Zhou, S~P Sharan, Harsh Goel, Arnab Sanyal,
  Sandeep Chinchali, and Gourav Datta.
\newblock {RT-NeuS}: Towards real-time neuro-symbolic video understanding via
  adaptive temporal verification.
\newblock \emph{arXiv preprint arXiv:2602.23553}, 2026.
\newblock \doi{10.48550/arXiv.2602.23553}.
\newblock URL \url{https://arxiv.org/abs/2602.23553v2}.

\bibitem[Lightman et~al.(2024)Lightman, Kosaraju, Burda, Edwards, Baker, Lee,
  Leike, Schulman, Sutskever, and Cobbe]{lightman2023verify}
Hunter Lightman, Vineet Kosaraju, Yuri Burda, Harrison Edwards, Bowen Baker,
  Teddy Lee, Jan Leike, John Schulman, Ilya Sutskever, and Karl Cobbe.
\newblock Let's verify step by step.
\newblock In \emph{International Conference on Learning Representations}, 2024.
\newblock URL
  \url{https://proceedings.iclr.cc/paper_files/paper/2024/hash/aca97732e30bcf1303bc22ac3924fd16-Abstract-Conference.html}.

\bibitem[Liu \& Zhang(2025)Liu and Zhang]{liu2025videojudge}
Ming Liu and Wensheng Zhang.
\newblock Is your video language model a reliable judge?
\newblock In \emph{International Conference on Learning Representations}, 2025.
\newblock URL \url{https://openreview.net/forum?id=m8yby1JfbU}.

\bibitem[Liu \& Lee(2026)Liu and Lee]{liu2026shell}
Tiedong Liu and Wee~Sun Lee.
\newblock Can vision-language models solve the shell game?
\newblock \emph{arXiv preprint arXiv:2603.08436}, 2026.
\newblock URL \url{https://arxiv.org/abs/2603.08436}.

\bibitem[Liu et~al.(2023)Liu, Iter, Xu, Wang, Xu, and Zhu]{liu2023geval}
Yang Liu, Dan Iter, Yichong Xu, Shuohang Wang, Ruochen Xu, and Chenguang Zhu.
\newblock {G-Eval}: {NLG} evaluation using {GPT-4} with better human alignment.
\newblock In \emph{Proceedings of the 2023 Conference on Empirical Methods in
  Natural Language Processing}, pp.\  2511--2522, Singapore, 2023. Association
  for Computational Linguistics.
\newblock \doi{10.18653/v1/2023.emnlp-main.153}.
\newblock URL \url{https://aclanthology.org/2023.emnlp-main.153/}.

\bibitem[Liu et~al.(2024)Liu, Li, Liu, Wang, Ren, Li, Chen, Sun, and
  Hou]{liu2024tempcompass}
Yuanxin Liu, Shicheng Li, Yi~Liu, Yuxiang Wang, Shuhuai Ren, Lei Li, Sishuo
  Chen, Xu~Sun, and Lu~Hou.
\newblock {TempCompass}: Do video {LLM}s really understand videos?
\newblock In \emph{Findings of the Association for Computational Linguistics:
  ACL 2024}, pp.\  8731--8772, Bangkok, Thailand, 2024. Association for
  Computational Linguistics.
\newblock \doi{10.18653/v1/2024.findings-acl.517}.
\newblock URL \url{https://aclanthology.org/2024.findings-acl.517/}.

\bibitem[Meng et~al.(2026)Meng, Yue, Xu, Wang, Ren, Liu, Wang, Zhang, Tong, and
  Duan]{meng2026videozero}
Jiahao Meng, Tan Yue, Qi~Xu, Haochen Wang, Zhongwei Ren, Weisong Liu, Yuhao
  Wang, Renrui Zhang, Yunhai Tong, and Haodong Duan.
\newblock {VideoZeroBench}: {Probing} the limits of video {MLLM}s with
  spatio-temporal evidence verification.
\newblock \emph{arXiv preprint arXiv:2604.01569}, 2026.
\newblock \doi{10.48550/arXiv.2604.01569}.
\newblock URL \url{https://arxiv.org/abs/2604.01569}.

\bibitem[Min et~al.(2023)Min, Krishna, Lyu, Lewis, Yih, Koh, Iyyer,
  Zettlemoyer, and Hajishirzi]{min2023factscore}
Sewon Min, Kalpesh Krishna, Xinxi Lyu, Mike Lewis, Wen-tau Yih, Pang~Wei Koh,
  Mohit Iyyer, Luke Zettlemoyer, and Hannaneh Hajishirzi.
\newblock {FActScore}: Fine-grained atomic evaluation of factual precision in
  long form text generation.
\newblock In \emph{Proceedings of the 2023 Conference on Empirical Methods in
  Natural Language Processing}, pp.\  12076--12100, Singapore, 2023.
  Association for Computational Linguistics.
\newblock \doi{10.18653/v1/2023.emnlp-main.741}.
\newblock URL \url{https://aclanthology.org/2023.emnlp-main.741/}.

\bibitem[Min et~al.(2026)Min, Zhou, Li, Wu, Peng, Du, Zhao, Yang, and
  Wen]{min2026perceval}
Yingqian Min, Kun Zhou, Yifan Li, Yuhuan Wu, Han Peng, Yifan Du, Wayne~Xin
  Zhao, Min Yang, and Ji-Rong Wen.
\newblock Improving vision-language models with perception-centric process
  reward models.
\newblock In \emph{Proceedings of the IEEE/CVF Conference on Computer Vision
  and Pattern Recognition}, pp.\  33099--33109, 2026.
\newblock URL
  \url{https://openaccess.thecvf.com/content/CVPR2026/html/Min_Improving_Vision-language_Models_with_Perception-centric_Process_Reward_Models_CVPR_2026_paper.html}.

\bibitem[Park et~al.(2026)Park, Choi, Kang, Lee, Shin, and
  Shim]{park2026perceptual}
Seojeong Park, Jiho Choi, Junyong Kang, Seonho Lee, Jaeyo Shin, and Hyunjung
  Shim.
\newblock Mitigating perceptual judgment bias in multimodal {LLM}-as-a-judge
  via perceptual perturbation and reward modeling.
\newblock In \emph{Proceedings of the 43rd International Conference on Machine
  Learning}, 2026.
\newblock URL \url{https://arxiv.org/abs/2606.02578}.

\bibitem[P{\u{a}}tr{\u{a}}ucean et~al.(2023)P{\u{a}}tr{\u{a}}ucean, Smaira,
  Gupta, Recasens~Continente, Markeeva, Banarse, Koppula, Heyward, Malinowski,
  Yang, Doersch, Matejovicova, Sulsky, Miech, Frechette, Klimczak, Koster,
  Zhang, Winkler, Aytar, Osindero, Damen, Zisserman, and
  Carreira]{patraucean2023perception}
Viorica P{\u{a}}tr{\u{a}}ucean, Lucas Smaira, Ankush Gupta, Adri{\`a}
  Recasens~Continente, Larisa Markeeva, Dylan Banarse, Skanda Koppula, Joseph
  Heyward, Mateusz Malinowski, Yi~Yang, Carl Doersch, Tatiana Matejovicova,
  Yury Sulsky, Antoine Miech, Alex Frechette, Hanna Klimczak, Raphael Koster,
  Junlin Zhang, Stephanie Winkler, Yusuf Aytar, Simon Osindero, Dima Damen,
  Andrew Zisserman, and Jo{\~a}o Carreira.
\newblock Perception test: {A} diagnostic benchmark for multimodal video
  models.
\newblock In \emph{Advances in Neural Information Processing Systems 36
  (Datasets and Benchmarks Track)}, 2023.
\newblock URL \url{https://arxiv.org/abs/2305.13786}.

\bibitem[Pi et~al.(2025)Pi, Bai, Chen, Wang, Shan, Liu, and Cao]{pi2025mrjudge}
Renjie Pi, Haoping Bai, Qibin Chen, Xiaoming~Simon Wang, Jiulong Shan,
  Xiaojiang Liu, and Meng Cao.
\newblock {MR}. judge: Multimodal reasoner as a judge.
\newblock In \emph{Proceedings of the 2025 Conference on Empirical Methods in
  Natural Language Processing}, pp.\  20181--20205, Suzhou, China, 2025.
  Association for Computational Linguistics.
\newblock \doi{10.18653/v1/2025.emnlp-main.1021}.
\newblock URL \url{https://aclanthology.org/2025.emnlp-main.1021/}.

\bibitem[Qi et~al.(2022)Qi, Gao, Hu, Wang, Liu, Bai, Belongie, Yuille, Torr,
  and Bai]{qi2022ovis}
Jiyang Qi, Yan Gao, Yao Hu, Xinggang Wang, Xiaoyu Liu, Xiang Bai, Serge
  Belongie, Alan Yuille, Philip H.~S. Torr, and Song Bai.
\newblock Occluded video instance segmentation: A benchmark.
\newblock \emph{International Journal of Computer Vision}, 130\penalty0
  (8):\penalty0 2022--2039, 2022.
\newblock \doi{10.1007/s11263-022-01629-1}.
\newblock URL \url{https://arxiv.org/abs/2102.01558}.

\bibitem[Rahman(2026)]{rahman2026temporal}
Farrukh Rahman.
\newblock What does a temporal benchmark score measure? {Decomposing} channel
  use in video {VLM} evaluation.
\newblock \emph{arXiv preprint arXiv:2607.12304}, 2026.
\newblock URL \url{https://arxiv.org/abs/2607.12304}.

\bibitem[Saravanan et~al.(2025)Saravanan, Gupta, Singh, Khan, Gandhi, and
  Tapaswi]{saravanan2025velociti}
Darshana Saravanan, Varun Gupta, Darshan Singh, Zeeshan Khan, Vineet Gandhi,
  and Makarand Tapaswi.
\newblock {VELOCITI}: Benchmarking video-language compositional reasoning with
  strict entailment.
\newblock In \emph{Proceedings of the IEEE/CVF Conference on Computer Vision
  and Pattern Recognition}, pp.\  18914--18924, 2025.
\newblock URL
  \url{https://openaccess.thecvf.com/content/CVPR2025/html/Saravanan_VELOCITI_Benchmarking_Video-Language_Compositional_Reasoning_with_Strict_Entailment_CVPR_2025_paper.html}.

\bibitem[Sevilla-Lara et~al.(2021)Sevilla-Lara, Zha, Yan, Goswami, Feiszli, and
  Torresani]{sevillalara2021onlytime}
Laura Sevilla-Lara, Shengxin Zha, Zhicheng Yan, Vedanuj Goswami, Matt Feiszli,
  and Lorenzo Torresani.
\newblock Only time can tell: Discovering temporal data for temporal modeling.
\newblock In \emph{Proceedings of the IEEE/CVF Winter Conference on
  Applications of Computer Vision}, pp.\  535--544, 2021.

\bibitem[Shah et~al.(2026)Shah, Sharan, Goel, Choi, Munir, Pasula, Marculescu,
  and Chinchali]{shah2026neusqa}
Sahil Shah, S~P Sharan, Harsh Goel, Minkyu Choi, Mustafa Munir, Manvik Pasula,
  Radu Marculescu, and Sandeep Chinchali.
\newblock {NeuS-QA}: Grounding long-form video understanding in temporal logic
  and neuro-symbolic reasoning.
\newblock In \emph{Proceedings of the AAAI Conference on Artificial
  Intelligence}, volume~40, pp.\  8805--8813, 2026.
\newblock \doi{10.1609/aaai.v40i11.37834}.
\newblock URL \url{https://ojs.aaai.org/index.php/AAAI/article/view/37834}.

\bibitem[Sigurdsson et~al.(2016)Sigurdsson, Varol, Wang, Farhadi, Laptev, and
  Gupta]{sigurdsson2016charades}
Gunnar~A. Sigurdsson, G{\"u}l Varol, Xiaolong Wang, Ali Farhadi, Ivan Laptev,
  and Abhinav Gupta.
\newblock Hollywood in homes: Crowdsourcing data collection for activity
  understanding.
\newblock In \emph{Computer Vision -- {ECCV} 2016}, pp.\  510--526, 2016.
\newblock \doi{10.1007/978-3-319-46448-0_31}.
\newblock URL \url{https://arxiv.org/abs/1604.01753}.

\bibitem[Thrush et~al.(2022)Thrush, Jiang, Bartolo, Singh, Williams, Kiela, and
  Ross]{thrush2022winoground}
Tristan Thrush, Ryan Jiang, Max Bartolo, Amanpreet Singh, Adina Williams, Douwe
  Kiela, and Candace Ross.
\newblock {Winoground}: Probing vision and language models for visio-linguistic
  compositionality.
\newblock In \emph{Proceedings of the IEEE/CVF Conference on Computer Vision
  and Pattern Recognition}, pp.\  5238--5248, 2022.
\newblock URL \url{https://arxiv.org/abs/2204.03162}.

\bibitem[Wang et~al.(2026{\natexlab{a}})Wang, Guan, Qiu, Li, Gai, Yang, Zhou,
  Zhao, Jiang, and Jiang]{wang2026grounding}
Junxin Wang, Dai Guan, Weijie Qiu, Zhihang Li, Yongbo Gai, Zhengyi Yang, Mengyu
  Zhou, Erchao Zhao, Xiaoxi Jiang, and Guanjun Jiang.
\newblock Grounding the score: Explicit visual premise verification for
  reliable vision-language process reward models.
\newblock \emph{arXiv preprint arXiv:2603.16253}, 2026{\natexlab{a}}.
\newblock URL \url{https://arxiv.org/abs/2603.16253}.

\bibitem[Wang et~al.(2025)Wang, Gao, Gu, Pu, et~al.]{wang2025internvl35}
Weiyun Wang, Zhangwei Gao, Lixin Gu, Hengjun Pu, et~al.
\newblock {InternVL3.5}: Advancing open-source multimodal models in
  versatility, reasoning, and efficiency.
\newblock \emph{arXiv preprint arXiv:2508.18265}, 2025.
\newblock \doi{10.48550/arXiv.2508.18265}.
\newblock URL \url{https://arxiv.org/abs/2508.18265}.

\bibitem[Wang et~al.(2024)Wang, Ma, Hu, Weber-Genzel, R{\"o}ttger, Kreuter,
  Hovy, and Plank]{wang2024answerc}
Xinpeng Wang, Bolei Ma, Chengzhi Hu, Leon Weber-Genzel, Paul R{\"o}ttger,
  Frauke Kreuter, Dirk Hovy, and Barbara Plank.
\newblock {``My Answer is C''}: First-token probabilities do not match text
  answers in instruction-tuned language models.
\newblock In \emph{Findings of the Association for Computational Linguistics:
  ACL 2024}, pp.\  7407--7416, Bangkok, Thailand, 2024. Association for
  Computational Linguistics.
\newblock \doi{10.18653/v1/2024.findings-acl.441}.
\newblock URL \url{https://aclanthology.org/2024.findings-acl.441/}.

\bibitem[Wang et~al.(2026{\natexlab{b}})Wang, Chen, Qin, Wei, Qian, and
  Bai]{wang2026videohv}
Zheng Wang, Haoran Chen, Haoxuan Qin, Zhipeng Wei, Tianwen Qian, and Cong Bai.
\newblock Think, then verify: {A} hypothesis--verification multi-agent
  framework for long video understanding.
\newblock In \emph{Proceedings of the IEEE/CVF Conference on Computer Vision
  and Pattern Recognition}, pp.\  33784--33793, 2026{\natexlab{b}}.
\newblock URL
  \url{https://openaccess.thecvf.com/content/CVPR2026/html/Wang_Think_Then_Verify_A_Hypothesis-Verification_Multi-Agent_Framework_for_Long_Video_CVPR_2026_paper.html}.

\bibitem[Yu et~al.(2026)Yu, Ma, Huang, Lee, Yang, Choi, Brown, Michel, Zheng,
  Shin, and Xie]{yu2026vstat}
Sihyun Yu, Nanye Ma, Pinzhi Huang, Hyunseok Lee, Shusheng Yang, June~Suk Choi,
  Ellis Brown, Oscar Michel, Boyang Zheng, Jinwoo Shin, and Saining Xie.
\newblock Benchmarking visual state tracking in multimodal video understanding.
\newblock \emph{arXiv preprint arXiv:2606.03920}, 2026.
\newblock \doi{10.48550/arXiv.2606.03920}.
\newblock URL \url{https://arxiv.org/abs/2606.03920}.

\bibitem[Yuksekgonul et~al.(2023)Yuksekgonul, Bianchi, Kalluri, Jurafsky, and
  Zou]{yuksekgonul2023aro}
Mert Yuksekgonul, Federico Bianchi, Pratyusha Kalluri, Dan Jurafsky, and James
  Zou.
\newblock When and why vision-language models behave like bags-of-words, and
  what to do about it?
\newblock In \emph{International Conference on Learning Representations}, 2023.
\newblock URL \url{https://openreview.net/forum?id=KRLUvxh8uaX}.

\bibitem[Zhang et~al.(2024)Zhang, Cai, and Lee]{zhang2024vinoground}
Jianrui Zhang, Mu~Cai, and Yong~Jae Lee.
\newblock {Vinoground}: Scrutinizing {LMM}s over dense temporal reasoning with
  short videos.
\newblock \emph{arXiv preprint arXiv:2410.02763}, 2024.
\newblock \doi{10.48550/arXiv.2410.02763}.
\newblock URL \url{https://arxiv.org/abs/2410.02763}.

\bibitem[Zhao et~al.(2025)Zhao, Liu, Yu, Kung, Chen, Mi, and
  Yu]{zhao2025onetoken}
Yulai Zhao, Haolin Liu, Dian Yu, Sunyuan Kung, Meijia Chen, Haitao Mi, and Dong
  Yu.
\newblock One token to fool {LLM}-as-a-judge.
\newblock \emph{arXiv preprint arXiv:2507.08794}, 2025.
\newblock URL \url{https://arxiv.org/abs/2507.08794}.

\bibitem[Zheng et~al.(2023)Zheng, Chiang, Sheng, Zhuang, Wu, Zhuang, Lin, Li,
  Li, Xing, Zhang, Gonzalez, and Stoica]{zheng2023judging}
Lianmin Zheng, Wei-Lin Chiang, Ying Sheng, Siyuan Zhuang, Zhanghao Wu, Yonghao
  Zhuang, Zi~Lin, Zhuohan Li, Dacheng Li, Eric~P. Xing, Hao Zhang, Joseph~E.
  Gonzalez, and Ion Stoica.
\newblock Judging {LLM}-as-a-judge with {MT-Bench} and chatbot arena.
\newblock In \emph{Advances in Neural Information Processing Systems 36
  (Datasets and Benchmarks Track)}, 2023.
\newblock \doi{10.52202/075280-2020}.
\newblock URL
  \url{https://papers.neurips.cc/paper_files/paper/2023/hash/91f18a1287b398d378ef22505bf41832-Abstract-Datasets_and_Benchmarks.html}.

\bibitem[Zou et~al.(2026)Zou, Sridhar, Safarzadeh, and
  Roth]{zou2026withoutseeing}
Xiaohan Zou, Roshan Sridhar, Mohammadtaher Safarzadeh, and Dan Roth.
\newblock When vision-language models judge without seeing: Exposing
  informativeness bias.
\newblock In \emph{Proceedings of the 64th Annual Meeting of the Association
  for Computational Linguistics (Volume 1: Long Papers)}, pp.\  15417--15448,
  San Diego, California, United States, 2026. Association for Computational
  Linguistics.
\newblock \doi{10.18653/v1/2026.acl-long.703}.
\newblock URL \url{https://aclanthology.org/2026.acl-long.703/}.

\end{thebibliography}

\clearpage
\appendix

\setcounter{topnumber}{4}
\setcounter{bottomnumber}{4}
\setcounter{totalnumber}{8}
\renewcommand{\topfraction}{0.95}
\renewcommand{\bottomfraction}{0.95}
\renewcommand{\textfraction}{0.05}
\renewcommand{\floatpagefraction}{0.8}

\section{Cohorts, checkpoints, and reference annotation}
\label{app:cohorts}

\subsection{Experiments and shared units}
\label{app:cohorts-blocks}

\Cref{tab:studies} summarizes the four studies. The resampling unit is the program unit in the construction study, the program
pair in the truth-switch study (60 pairs), the video in the TOC-Bench and selection analyses (400 videos), and the pair in the second natural panel (400 Vinoground pairs, grouped by shared source).

No truth-switch pair was filtered after the pilot, and witness programs are
disjoint across pairs. Additional pairs and videos served technical pilot runs only and are
excluded from every estimate.

\subsection{Checkpoints and inference settings}
\label{app:checkpoints}

All models were loaded from the fixed Hugging Face revisions in
\cref{tab:checkpoints}; the model reports are \citet{bai2025qwen3vl},
\citet{hong2025glm45v}, \citet{clark2026molmo2}, \citet{wang2025internvl35},
\citet{moonshot2025kimivl}, and \citet{li2024aria}. Inference used bfloat16
weights, greedy decoding, flash-attention-2 for Qwen, GLM, Kimi-VL, and Aria,
SDPA attention for Molmo2 and InternVL, and Transformers 4.57.6 (4.50.3 for
Kimi). The Molmo--InternVL truth-switch study ran each model on A100 40~GB GPUs (one for
Molmo2, four for InternVL); other runs used H100 80~GB, A100 80~GB, or A100
40~GB GPUs.

\begin{table}[h]
  \caption{\textbf{Evaluated checkpoints.} Hugging Face repositories and
  revision identifiers of the six models.}
  \label{tab:checkpoints}
  \centering
  \small
  \begin{tabularx}{\linewidth}{@{}L{0.15\linewidth}L{0.30\linewidth}Y@{}}
  \toprule
  Model & Repository & Revision \\
  \midrule
  Qwen3-VL-32B & \revision{Qwen/Qwen3-VL-32B-Instruct} & \revision{0cfaf48183f594c314753d30a4c4974bc75f3ccb} \\
  GLM-4.5V & \revision{zai-org/GLM-4.5V} & \revision{ed47433b37111465ec527affaaddceff371bca04} \\
  Molmo2-8B & \revision{allenai/Molmo2-8B} & \revision{e28fa28597e5ec5e0cca2201dd8ab33d48bc4a1b} \\
  InternVL3.5-38B & \texttt{OpenGVLab/}\allowbreak\texttt{InternVL3\_5-38B-HF} & \revision{7c830fc25e874f9a861087e3c70587f2b9b555f3} \\
  Kimi-VL-A3B & \revision{moonshotai/Kimi-VL-A3B-Instruct} & \revision{398eede0903cd983a2bfa0cc634e9ac1d843f375} \\
  Aria & \revision{rhymes-ai/Aria} & \revision{53a09c77895d7025efa6239ea71cf1a8991064e8} \\
  \bottomrule
  \end{tabularx}
\end{table}

\begin{table}[h]
    \caption{\textbf{Token caps for CoT judgment in the per-model protocol of the submission.} All decoding is greedy.}
  \label{tab:caps}
  \centering
  \small
  \begin{tabular*}{\linewidth}{@{\extracolsep{\fill}}lrrrr@{}}
  \toprule
  Study & Qwen cap & GLM cap & Molmo cap & InternVL cap \\
  \midrule
  Truth-switch & 1,024 & 4,096 & 2,048 & 4,096 \\
  Natural-video & 8,192 & 4,096 & 2,048 & 2,048 \\
  \bottomrule
  \end{tabular*}
\end{table}

Under the common protocol of \cref{sec:outcomes}, every model receives the same instructions,
native thinking is off for GLM and InternVL, the two models that have it, and the requested cap
is 8,192 tokens; the context window lowers the effective cap on natural video for GLM,
InternVL, and Molmo (\cref{tab:generation-glm,tab:generation-internvl,tab:generation-molmo}). 

The per-model protocol is the protocol of our original submission. Its CoT settings differ by
model: Qwen and Molmo received an explicit reasoning instruction, GLM used native thinking, and
InternVL used the thinking-mode system prompt documented by its authors, under the token caps
of \cref{tab:caps}. CoT answers were parsed from the final answer line, with unparseable and
truncated outputs scored as incorrect, and direct judgment compared the A and B logits at the
first token after the prompt. This protocol applies to the material labeled ``per-model
protocol of the submission'' in \cref{app:truth-switch,app:natural}, to the rationale analysis
of \cref{app:trace-audit} and \cref{fig:bottle-occlusion}, and to verifier selection
(\cref{tab:selection} and \cref{app:selection}). The second panel keeps its CoT settings and
reads direct judgment at the validated position (\cref{app:revision-transfer}).
\cref{tab:protocol-labels} lists every protocol label used in the paper.

\subsection{Natural-video panel and reference annotation}
\label{app:provenance}

The 400-video panel was drawn from TOC-Bench before any evaluated model was
run: videos used in earlier development were excluded, the remainder were
ordered by question and video identifier and shuffled within stratum with
seed 20260817, and the first valid videos filled quotas of 120/100/100/80 for
event order, event-relative state, temporal location, and visibility
continuity. Videos are globally unique.

For each video, an LLM drafted a reference proposition from the video and the question, without
access to benchmark answers or model outputs. One of the authors then checked every reference
against the full video and accepted it only when the target was evaluable and unambiguous, all
required events were observed, and the categorical or numerical fields were valid; event-order
onsets within 0.05~s were excluded as ties. The natural-video study uses these references
unchanged. Two annotators who are not authors later validated every reference and both
negatives independently (\cref{app:revision-validation}).

\noindent\textbf{Negatives.} Both negatives are built by fixed rules from the reference and
event schema. No model is used to choose the alternatives. For event order, the denial template
is ``The event ‘E' does not occur anywhere in the video.'' It denies the second event in the
question schema, regardless of its chronological position. Event-relative state uses ``The
anchor event ‘E' does not occur anywhere in the video.'' Temporal location and visibility
continuity use ``The event ‘E' does not occur anywhere in the video,'' with the target event
for temporal location and the disappearance or reappearance event for visibility continuity.

The content-matched negative reverses the event order, swaps fully visible with partly hidden
or fully hidden with absent from the frame, or moves the event center to the next of eight
temporal segments. In the last segment it uses the preceding one. Visibility continuity swaps
the two schema values: full occlusion with leaving the frame, or the same physical instance
with a different instance after reappearance. \cref{sec:natural-design} gives one complete
example.

Shuffled evidence uses the identical decoded frame
multiset with a per-video seed derived from 20260817. TOC-Bench draws on Charades \citep{sigurdsson2016charades}, Perception Test
\citep{patraucean2023perception}, OVIS \citep{qi2022ovis}, and MOSE
\citep{ding2023mose} under their documented licenses.

\begin{revisedtext}
The second panel contains 400 Vinoground pairs (800 clips) selected by a seeded permutation before scoring, skipping reused clip intervals. The content-matched negative of a clip is the caption of the other clip in its
pair, and the event-denial negative denies the captioned event
(\cref{app:revision-transfer}).
\end{revisedtext}
\section{Controlled negatives and the construction study}
\label{app:controlled}

\subsection{Transition legality and rendering}
\label{app:rendering}

A controlled program fixes an initial state and an ordered list of events, and
execution yields the faithful trajectory. Shell tracks the ball's slot in
$\{0,1,2\}$, and an event swaps slots $(0,1)$, $(1,2)$, or $(0,2)$; toggle
tracks three binary switches, and an event flips one of them; counter tracks a
level in $\{0,\ldots,7\}$, and an event adds $-2$, $-1$, $+1$, or $+2$, with
programs that leave this range excluded. A trajectory is legal when some
allowed event sequence produces it; legality does not require that this be
the event sequence shown in the input. Because any two shell slots are one
swap apart, every well-formed shell state sequence is legal, so legality
constrains toggle and counter candidates more than shell candidates.
Endpoint-matched negatives also satisfy \cref{eq:endpoint-matched}.

Every program of length four or six is enumerated. A unit $(P,Q,R)$ consists
of three executable programs with the same initial and final states: $Q$
differs from the faithful program $P$ at exactly one intermediate state, and
$R$ differs at two or more intermediate states while using $P$'s multiset of
events. For shell and toggle, $R$ is the unused program with the most
differing intermediate states; for counter, it is the first program meeting
these conditions in a seeded order. Seeded selection with an equal quota in
every (initial state, final state, deviation position) stratum (shell and
toggle) or seeded greedy packing (counter), followed by a hash-ordered pick
balanced over the position of $Q$'s deviation, gives 54 units per task and
length, 324 in total, and no program appears in more than one unit. The four
negative types are built per unit. The endpoint-matched negatives are the
trajectories of $R$ (several internal deviations, the main type) and $Q$ (one
localized deviation, the single-deviation type). The visible-start negative
replays $P$'s events from a changed initial state at the same component
distance as $Q$'s deviation and is kept only when every step remains legal.
The isolated-start negative changes only the initial state and keeps $P$'s
later states; because every event is one-to-one, the first displayed event
cannot produce its first transition. The endpoint-matching drop compares the
visible-start negative with the several-deviation endpoint-matched negative;
the single-deviation and isolated-start negatives give context.

The renderer (\texttt{latent-event-board-v1}) produces
$320\times256$ RGB keyframes of a titled board: three panels centered at 0.24,
0.50, and 0.76 of the width for shell and toggle, and a tick-marked 0--7 number
line in the counter's initial frame. The initial frame reveals the state.
Every event frame carries an ``EVENT $k$ OF $n$'' marker and indicates the
event without drawing any state: a double-headed arrow between the swapped
slots and a ``SWAP $i$ <-> $j$'' label for shell, a pointer above the toggled
switch and a ``TOGGLE $k$'' label with all switch states masked for toggle,
and the signed step beside a masked level for counter. The final frame shows a
mask labelled ``STATE HIDDEN'' and draws no state. A unit-level seed shifts
the drawing by at most 3~px horizontally and 2~px vertically and scales the
palette brightness by at most $\pm3.5\%$, identically in every frame of $P$,
$Q$, and $R$. Compared conditions reuse the corresponding program and seed, and the
truth-switch witness videos are the renders of $P$ and $R$.

\subsection{Scores, prompts, and qualification criteria}
\label{app:scores}

For independently scored candidates, the win rate of \cref{sec:outcomes} is
\begin{equation}
W=\mathbb{E}\!\left[\mathbf 1\{s(c^+)>s(c^-)\}
+\tfrac12\mathbf 1\{s(c^+)=s(c^-)\}\right].
\label{eq:pairwise}
\end{equation}
The score $s$ of a candidate is the next-token log-odds of the answer letter
that means it matches the evidence, averaged over both assignments of A and B
to the two meanings. The pairwise template presents the two trajectories
together and counterbalances their order.

The pairwise template is, with braced fields filled from the unit:
\begin{quote}\small\ttfamily\raggedright
You are comparing two proposed state traces against evidence.\\
\{evidence\}\\[0.3em]
Task: \{task\}\\[0.3em]
Each candidate has exactly three fields: the initial state, every intermediate
post-event state in chronological order, and the final state. The final answer
is deliberately omitted.\\[0.3em]
Candidate A:\\
\{candidate\_a\}\\[0.3em]
Candidate B:\\
\{candidate\_b\}\\[0.3em]
Choose the one candidate whose complete state trace exactly matches the
evidence. Reply with exactly one token: A or B.
\end{quote}
With video, the evidence line reads ``You are given $N$ frames in
chronological order, labelled Frame 0 through Frame $N-1$.'' The shell task
text is ``A ball starts at one of three slots 0=left, 1=middle, 2=right. Each
shown event swaps exactly the two named slots. Track the ball slot after every
event.'' The toggle task text is ``Three switches are ordered left, middle,
right. Each shown event toggles exactly one named switch. A state is a
three-bit string in left-to-right order, where 1=ON and 0=OFF.'' Candidate
trajectories are written with the lines \texttt{Initial:}, \texttt{Intermediate:} (post-event
states joined by \texttt{->}), and \texttt{Final:}. The absolute template opens
``You are auditing one proposed state trace against evidence,'' shares the
evidence, task, and field paragraphs, presents one candidate trajectory, and
counterbalances A/B between ``every stated state exactly matches the
evidence'' and ``at least one stated state is wrong, missing, or extra.''

Three criteria were fixed before the endpoint-matching drop was inspected.
In the \emph{text-record test}, the model receives the exact state sequence as
text instead of video and must reach at least 0.90 accuracy, with a pointwise
95\% lower bound above 0.75, against every negative type. The
\emph{visible-start criterion} requires a win rate against visible-start
negatives of at least 0.75 with video (simultaneous lower bound above 0.50) and
a \emph{video effect}, the win rate with video minus the win rate without, of at
least 0.15 (simultaneous lower bound above zero). The \emph{drop criterion}
requires an endpoint-matching drop of at least 0.15 with a simultaneous lower
bound above zero, and a video effect on the drop (the drop with video minus the
drop without video) whose simultaneous lower bound is positive. A cell
\emph{qualifies} when it meets all three criteria.

The construction study uses the studentized 864-estimate family of
\cref{app:estimands}; renderer and text-record diagnostics use separately
declared pointwise procedures. \rev{GLM is scored at the validated answer position (\cref{app:revision-position}); the other models
keep their scores from the original answer position.}

\subsection{Complete results at the validated readout}
\label{app:controlled-results}

\Cref{tab:m2-absolute} reports all six models, three tasks, and four negative
types, including cells that do not qualify. On shell, GLM falls from 0.958 to 0.574 and InternVL from 0.926 to 0.583; GLM toggle falls from 0.769 to 0.611 and does not meet the drop criterion; on toggle, Molmo
falls from 0.901 to 0.556. Kimi fails the text-record test except on counter; Aria passes
the test, but none of its cells qualifies. Kimi-VL and Aria are scored at the original answer position. Their label mass at that position
was not measured (\cref{app:revision-position}), so Kimi-VL's failure of the text-record test
may reflect the readout as well as the model. No counter cell qualifies either:
Qwen's and InternVL's counter drops are significant, but no counter drop has a
significantly positive video effect, and the counter renderer is sensitive to
rendering details. The video effect on the drop is positive for Qwen, GLM,
Molmo, and InternVL, so a text-only preference between trajectories does not
by itself explain their drops; these diagnostics are included in the
artifact.

\begin{table}[ht]
\centering\scriptsize
\caption{\rev{\textbf{Construction win rates.} Simultaneous 95\% intervals. GLM uses the validated answer position.}}
\label{tab:m2-absolute}
\begin{tabular}{@{}llcccc@{}}
\toprule
Model & Task & Isolated-start & Visible-start & Several deviations & One deviation \\
\midrule
Qwen3-VL-32B & Shell & $0.97 [0.91,1.00]$ & $0.94 [0.87,1.00]$ & $0.50 [0.32,0.68]$ & $0.51 [0.32,0.70]$ \\
 & Toggle & $0.84 [0.74,0.94]$ & $0.85 [0.75,0.94]$ & $0.56 [0.39,0.74]$ & $0.56 [0.39,0.74]$ \\
 & Counter & $0.82 [0.70,0.94]$ & $0.83 [0.69,0.96]$ & $0.56 [0.38,0.75]$ & $0.46 [0.27,0.65]$ \\
GLM-4.5V & Shell & $0.99 [0.95,1.00]$ & $0.96 [0.89,1.00]$ & $0.57 [0.39,0.76]$ & $0.50 [0.31,0.69]$ \\
 & Toggle & $0.87 [0.79,0.95]$ & $0.77 [0.67,0.87]$ & $0.61 [0.43,0.79]$ & $0.51 [0.32,0.70]$ \\
 & Counter & $0.90 [0.80,0.99]$ & $0.79 [0.65,0.93]$ & $0.63 [0.45,0.80]$ & $0.53 [0.34,0.71]$ \\
InternVL3.5-38B & Shell & $0.99 [0.96,1.00]$ & $0.93 [0.85,1.00]$ & $0.58 [0.40,0.77]$ & $0.45 [0.26,0.64]$ \\
 & Toggle & $0.93 [0.87,0.99]$ & $0.90 [0.83,0.97]$ & $0.46 [0.27,0.65]$ & $0.55 [0.37,0.73]$ \\
 & Counter & $0.97 [0.92,1.00]$ & $0.95 [0.87,1.00]$ & $0.70 [0.53,0.87]$ & $0.64 [0.46,0.82]$ \\
Molmo2-8B & Shell & $0.82 [0.69,0.95]$ & $0.64 [0.49,0.80]$ & $0.57 [0.39,0.76]$ & $0.55 [0.36,0.73]$ \\
 & Toggle & $0.90 [0.82,0.97]$ & $0.90 [0.83,0.97]$ & $0.56 [0.37,0.74]$ & $0.53 [0.34,0.71]$ \\
 & Counter & $0.85 [0.74,0.95]$ & $0.75 [0.60,0.90]$ & $0.54 [0.35,0.72]$ & $0.45 [0.26,0.64]$ \\
Kimi-VL-A3B & Shell & $0.43 [0.28,0.57]$ & $0.45 [0.32,0.59]$ & $0.44 [0.26,0.62]$ & $0.40 [0.23,0.57]$ \\
 & Toggle & $0.53 [0.39,0.68]$ & $0.50 [0.36,0.64]$ & $0.40 [0.23,0.57]$ & $0.46 [0.29,0.64]$ \\
 & Counter & $0.47 [0.32,0.62]$ & $0.38 [0.21,0.54]$ & $0.41 [0.23,0.58]$ & $0.54 [0.36,0.72]$ \\
Aria & Shell & $0.56 [0.41,0.70]$ & $0.51 [0.36,0.67]$ & $0.55 [0.36,0.74]$ & $0.43 [0.24,0.61]$ \\
 & Toggle & $0.63 [0.50,0.76]$ & $0.56 [0.42,0.70]$ & $0.51 [0.33,0.70]$ & $0.45 [0.26,0.64]$ \\
 & Counter & $0.49 [0.34,0.64]$ & $0.46 [0.28,0.63]$ & $0.35 [0.17,0.54]$ & $0.46 [0.27,0.64]$ \\
\bottomrule
\end{tabular}
\end{table}

\FloatBarrier

\section{Fixed-candidate truth-switch protocol and diagnostics}
\label{app:truth-switch}

\begin{revisedtext}
\cref{app:truth-switch-parsing,app:truth-switch-uncertainty,app:truth-switch-margin} document
the truth-switch test under the per-model protocol of the submission (\cref{app:checkpoints}),
including the original readouts and the separate analyses of two models each.
\cref{tab:revision-truth} gives the same test under the common protocol,
\cref{app:revision-interface} gives its prompts and controls, and
\cref{app:revision-reconciliation} compares the two sets of results.
\end{revisedtext}

\subsection{Design and input controls}
\label{app:truth-switch-construction}

\begin{figure}[h]
  \centering
  \includegraphics[width=\linewidth]{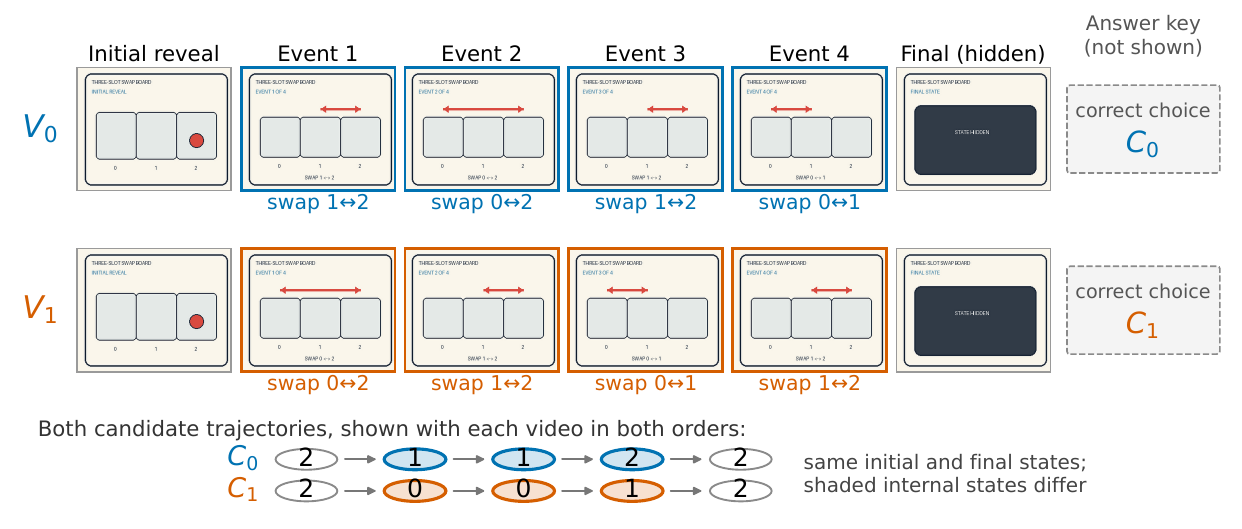}
  \caption{\rev{\textbf{A truth-switch pair.} The two witness videos share the initial reveal and the covered final frame and differ only in their events, so $V_0$ realizes $C_0$ and $V_1$ realizes $C_1$. Both candidate trajectories are shown with each video, in both orders.}}
  \label{fig:truth-switch}
\end{figure}

For each of the 60 candidate pairs, two witness programs realize the two
different trajectories while preserving the initial and final states. Initial
and final decoded pixels match within each pair. Candidate text is fixed across
the videos. Text prompts and input token IDs match within each candidate order
and interface, and each response is obtained in a fresh conversation.
\Cref{fig:truth-switch} shows the first four-event shell pair in the fixed pair
order, chosen without reading any model output. The Molmo--InternVL study reuses the same pairs, frames, prompts, and
request matrix.

\begin{table}[h]
  \caption{\textbf{Counterbalancing in the truth-switch studies.} The
  paired-success indicator requires both video answers to be correct within
  one candidate order. The two indicators are averaged within each program
  pair before estimation.}
  \label{tab:counterbalancing}
  \centering
  \small
  \begin{tabular}{@{}ccc@{}}
  \toprule
  Witness video & Candidate order & Correct label \\
  \midrule
  $V_0$ & $C_0,C_1$ & A \\
  $V_1$ & $C_0,C_1$ & B \\
  $V_0$ & $C_1,C_0$ & B \\
  $V_1$ & $C_1,C_0$ & A \\
  \bottomrule
  \end{tabular}
\end{table}

Each four-event program supplies six frames, and each six-event program
supplies eight: an initial frame, the event frames, and the covered final
frame. In every pair, the two witness programs contain the same events in different
orders (\cref{app:rendering}). Because each event frame also prints its
position, the frame multisets differ, and paired success cannot separate use of
presentation order from use of the printed event labels and counters. The design is an intervention on internal-event evidence with exact
truth conditions rather than a pure chronology intervention. \rev{Removing the counter yields the same frames in different orders (\cref{app:revision-render}).}

\subsection{Prompts, parsing, and outcome definitions}
\label{app:truth-switch-parsing}

Direct judgment uses the pairwise template of \cref{app:scores} unchanged,
with thinking disabled, and selects the unique larger of the A and B
next-token logits; a tie is scored as incorrect. CoT judgment replaces the template's final
sentence, ``Reply with exactly one token: A or B.'', with ``Work through the
visible initial state and each event in order, compare both complete traces,
and explain the decisive mismatch concisely. End with exactly one final line
in the form \verb+`Answer: A`+ or \verb+`Answer: B`+.'' Qwen receives this
prompt as its reasoning instruction; GLM also runs in native thinking
mode. Decoding is greedy.

The parser reads the last nonempty line and accepts \verb+Answer: X+ for X in
\{A, B\}; in the Qwen--GLM study it also accepts GLM's native final-answer
wrapper \glmwrapper{X}, and the Molmo--InternVL study accepts only the
requested line. Truncated CoT outputs, unaccepted answers, and tied direct
maxima are scored as incorrect. Answer switch and candidate-identity order
disagreement are computed
over parsed, nontruncated responses.

In the Molmo--InternVL study, Molmo receives the same reasoning prompt and
InternVL also receives the thinking-mode system prompt used in the
natural-video study; direct judgment is unchanged.

\begin{table}[h]
    \caption{\rev{\textbf{Response-level diagnostics for the truth-switch studies (per-model protocol of the submission).} 240 responses per model and interface; rates are percentages. Upper block: Qwen--GLM study. Lower block: Molmo--InternVL study.}}
  \label{tab:truth-switch-diagnostics}
  \centering
  \scriptsize
  \begin{tabular*}{\linewidth}{@{\extracolsep{\fill}}llcccccc@{}}
\toprule
Model & Interface & Correct/240 & Accuracy & Paired & Neither & Switch & Order disagr. \\
\midrule
Qwen3-VL-32B & Direct & 124 & 51.67 & 5.83 & 2.50 & 6.67 & 82.50 \\
Qwen3-VL-32B & CoT & 208 & 86.67 & 75.00 & 1.67 & 75.00 & 0.00 \\
GLM-4.5V & Direct & 120 & 50.00 & 0.00 & 0.00 & 0.00 & 100.00 \\
GLM-4.5V & CoT & 225 & 93.75 & 88.33 & 0.83 & 88.33 & 0.00 \\
\addlinespace[0.35em]
Molmo2-8B & Direct & 122 & 50.83 & 3.33 & 1.67 & 5.00 & 40.00 \\
Molmo2-8B & CoT & 104 & 43.33 & 14.17 & 27.50 & 24.17 & 42.50 \\
InternVL3.5-38B & Direct & 123 & 51.25 & 4.17 & 1.67 & 4.17 & 71.67 \\
InternVL3.5-38B & CoT & 225 & 93.75 & 88.33 & 0.83 & 88.33 & 0.00 \\
\bottomrule
\end{tabular*}

\vspace{0.6em}
\begin{tabular*}{\linewidth}{@{\extracolsep{\fill}}llcccccc@{}}
\toprule
Model & Interface & Invalid & Truncated & Ties & Format fail. & Format corr./240 & A / B / none \\
\midrule
Qwen3-VL-32B & Direct & 4 & 0 & 4 & 4 & 124 & 219 / 17 / 4 \\
Qwen3-VL-32B & CoT & 32 & 31 & 0 & 32 & 208 & 97 / 111 / 32 \\
GLM-4.5V & Direct & 0 & 0 & 0 & 0 & 120 & 240 / 0 / 0 \\
GLM-4.5V & CoT & 15 & 15 & 0 & 110 & 130 & 115 / 110 / 15 \\
\addlinespace[0.35em]
Molmo2-8B & Direct & 0 & 0 & 0 & 0 & 122 & 118 / 122 / 0 \\
Molmo2-8B & CoT & 35 & 0 & 0 & 35 & 104 & 126 / 79 / 35 \\
InternVL3.5-38B & Direct & 5 & 0 & 5 & 5 & 123 & 36 / 199 / 5 \\
InternVL3.5-38B & CoT & 15 & 14 & 0 & 15 & 225 & 114 / 111 / 15 \\
\bottomrule
\end{tabular*}

  \par\vspace{3pt}
  \parbox{\linewidth}{\footnotesize \rev{\textit{Note.} Accuracy is single-video
  accuracy. Neither-correct includes output failures. The invalid, truncated, tie,
  and format columns overlap. The last column counts A answers, B answers, and
  responses without an accepted answer.}}
\end{table}

Molmo's 35 CoT non-choices end with lines such as ``Both A and B'' (20),
``Neither'' (10), ``A or B'' (4), or ``Cannot determine'' (1), and its
accepted CoT answers are near chance (104 of 205 correct); every accepted CoT
answer of the other three models is correct.

\begin{table}[h]
  \caption{\rev{\textbf{Submitted truth-switch paired success under direct and CoT judgment (per-model protocol of the submission).} Simultaneous 95\% intervals within each two-model study. Score-level is the share of pair-orders whose A$-$B margin moves in the correct direction (chance 50\%).}}
  \label{tab:truth-switch}
  \centering
  \small
  \setlength{\tabcolsep}{3pt}
\begin{tabular*}{\linewidth}{@{\extracolsep{\fill}}lcccc@{}}
\toprule
Model & Direct (\%) & Score-level (\%) & CoT (\%) & CoT $-$ direct (pp) \\
\midrule
Qwen3-VL-32B & $5.83\;[0.00,\,15.00]$ & $73.33$ & $75.00\;[65.83,\,84.17]$ & $+69.17\;[60.00,\,78.33]$ \\
GLM-4.5V & $0.00\;[0.00,\,9.17]$ & $60.83$ & $88.33\;[79.17,\,97.50]$ & $+88.33\;[79.17,\,97.50]$ \\
\addlinespace[0.35em]
Molmo2-8B & $3.33\;[0.00,\,10.83]$ & $55.00$ & $14.17\;[6.67,\,21.67]$ & $+10.83\;[3.33,\,18.33]$ \\
InternVL3.5-38B & $4.17\;[0.00,\,11.67]$ & $59.58$ & $88.33\;[80.83,\,95.83]$ & $+84.17\;[76.67,\,91.67]$ \\
\bottomrule
\end{tabular*}

\end{table}

\subsection{Task-specific results and uncertainty}
\label{app:truth-switch-uncertainty}

Each study's primary family contains six estimates: CoT paired success,
direct paired success, and their difference for each of its two models
(\cref{tab:truth-switch-primary}). The bootstrap uses 10,000 paired draws of
whole program pairs within the four task-by-length strata; the two studies are
analyzed as separate families. The saved bounded-family companions remain
useful when empirical outcomes are constant. They are Hoeffding intervals
with a Bonferroni split over each study's six estimates, clipped to the
parameter range. For constant observed outcomes, percentile intervals can
have identical bounds. The bounded companions for all 72 task-and-length
estimates are in the supplementary table file. For example, GLM's direct
empirical paired-success rate is zero, but its simultaneous bootstrap interval
is $[0.00, 9.17]$\% and its bounded-family companion is $[0.00, 21.37]$\%. A
zero count still admits a nonzero rate, and the bounded interval reports the
range consistent with it. The Qwen, GLM, and InternVL improvements also
exclude zero under the bounded companions; Molmo's does not
($[-31.91, 53.58]$), so its small gain rests on the bootstrap family alone.

\begin{table}[h]
  \caption{\rev{\textbf{All twelve primary truth-switch estimates, six per study (per-model protocol of the submission).} Rates are percentages and differences are points. \Cref{app:truth-switch-uncertainty,app:statistics} define the three interval types.}}
  \label{tab:truth-switch-primary}
  \centering
  \scriptsize
  \begin{tabular}{@{}llcccc@{}}
\toprule
Model & Quantity & Estimate & Pointwise 95\% & Simultaneous 95\% & Bounded family 95\% \\
\midrule
Qwen3-VL-32B & Direct paired success (\%) & $5.83$ & $[2.50,\,10.00]$ & $[0.00,\,15.00]$ & $[0.00,\,27.20]$ \\
Qwen3-VL-32B & CoT paired success (\%) & $75.00$ & $[67.50,\,82.50]$ & $[65.83,\,84.17]$ & $[53.63,\,96.37]$ \\
Qwen3-VL-32B & CoT $-$ direct (pp) & $+69.17$ & $[60.00,\,78.33]$ & $[60.00,\,78.33]$ & $[26.42,\,100.00]$ \\
\addlinespace[0.2em]
GLM-4.5V & Direct paired success (\%) & $0.00$ & $[0.00,\,0.00]$ & $[0.00,\,9.17]$ & $[0.00,\,21.37]$ \\
GLM-4.5V & CoT paired success (\%) & $88.33$ & $[82.50,\,93.33]$ & $[79.17,\,97.50]$ & $[66.96,\,100.00]$ \\
GLM-4.5V & CoT $-$ direct (pp) & $+88.33$ & $[82.50,\,93.33]$ & $[79.17,\,97.50]$ & $[45.59,\,100.00]$ \\
\midrule
Molmo2-8B & Direct paired success (\%) & $3.33$ & $[0.83,\,6.67]$ & $[0.00,\,10.83]$ & $[0.00,\,24.70]$ \\
Molmo2-8B & CoT paired success (\%) & $14.17$ & $[9.17,\,20.00]$ & $[6.67,\,21.67]$ & $[0.00,\,35.54]$ \\
Molmo2-8B & CoT $-$ direct (pp) & $+10.83$ & $[5.00,\,16.67]$ & $[3.33,\,18.33]$ & $[-31.91,\,53.58]$ \\
\addlinespace[0.2em]
InternVL3.5-38B & Direct paired success (\%) & $4.17$ & $[0.83,\,7.50]$ & $[0.00,\,11.67]$ & $[0.00,\,25.54]$ \\
InternVL3.5-38B & CoT paired success (\%) & $88.33$ & $[81.67,\,94.17]$ & $[80.83,\,95.83]$ & $[66.96,\,100.00]$ \\
InternVL3.5-38B & CoT $-$ direct (pp) & $+84.17$ & $[76.67,\,90.83]$ & $[76.67,\,91.67]$ & $[41.42,\,100.00]$ \\
\bottomrule
\end{tabular}

\end{table}

\begin{table}[h]
  \caption{\rev{\textbf{Task and length summaries (per-model protocol of the submission).} Descriptive summaries with pointwise 95\% intervals. Each task has 30 pairs and each task-and-length cell 15.}}
  \label{tab:truth-switch-tasks}
  \centering
  \scriptsize
  \setlength{\tabcolsep}{4pt}
\begin{tabular*}{\linewidth}{@{\extracolsep{\fill}}lllcccc@{}}
\toprule
Task & Events & Model & Pairs & Direct (\%) & CoT (\%) & CoT $-$ direct (pp) \\
\midrule
Shell & 4 and 6 & Qwen3-VL-32B & 30 & $10.00\;[3.33,\,18.33]$ & $61.67\;[48.33,\,75.00]$ & $+51.67\;[35.00,\,68.33]$ \\
Shell & 4 and 6 & GLM-4.5V & 30 & $0.00\;[0.00,\,0.00]$ & $80.00\;[70.00,\,90.00]$ & $+80.00\;[70.00,\,90.00]$ \\
Shell & 4 and 6 & Molmo2-8B & 30 & $5.00\;[0.00,\,10.00]$ & $18.33\;[10.00,\,26.67]$ & $+13.33\;[3.33,\,23.33]$ \\
Shell & 4 and 6 & InternVL3.5-38B & 30 & $3.33\;[0.00,\,8.33]$ & $91.67\;[83.33,\,98.33]$ & $+88.33\;[78.33,\,96.67]$ \\
\addlinespace[0.2em]
Shell & 4 & Qwen3-VL-32B & 15 & $10.00\;[0.00,\,20.00]$ & $63.33\;[43.33,\,83.33]$ & $+53.33\;[26.67,\,76.67]$ \\
Shell & 4 & GLM-4.5V & 15 & $0.00\;[0.00,\,0.00]$ & $83.33\;[70.00,\,93.33]$ & $+83.33\;[70.00,\,93.33]$ \\
Shell & 4 & Molmo2-8B & 15 & $3.33\;[0.00,\,10.00]$ & $20.00\;[6.67,\,33.33]$ & $+16.67\;[0.00,\,30.00]$ \\
Shell & 4 & InternVL3.5-38B & 15 & $0.00\;[0.00,\,0.00]$ & $93.33\;[83.33,\,100.00]$ & $+93.33\;[83.33,\,100.00]$ \\
\addlinespace[0.2em]
Shell & 6 & Qwen3-VL-32B & 15 & $10.00\;[0.00,\,20.00]$ & $60.00\;[40.00,\,80.00]$ & $+50.00\;[30.00,\,70.00]$ \\
Shell & 6 & GLM-4.5V & 15 & $0.00\;[0.00,\,0.00]$ & $76.67\;[60.00,\,90.00]$ & $+76.67\;[60.00,\,90.00]$ \\
Shell & 6 & Molmo2-8B & 15 & $6.67\;[0.00,\,16.67]$ & $16.67\;[6.67,\,30.00]$ & $+10.00\;[-3.33,\,23.33]$ \\
Shell & 6 & InternVL3.5-38B & 15 & $6.67\;[0.00,\,16.67]$ & $90.00\;[73.33,\,100.00]$ & $+83.33\;[66.67,\,96.67]$ \\
\addlinespace[0.2em]
Toggle & 4 and 6 & Qwen3-VL-32B & 30 & $1.67\;[0.00,\,5.00]$ & $88.33\;[80.00,\,95.00]$ & $+86.67\;[78.33,\,93.33]$ \\
Toggle & 4 and 6 & GLM-4.5V & 30 & $0.00\;[0.00,\,0.00]$ & $96.67\;[91.67,\,100.00]$ & $+96.67\;[91.67,\,100.00]$ \\
Toggle & 4 and 6 & Molmo2-8B & 30 & $1.67\;[0.00,\,5.00]$ & $10.00\;[3.33,\,16.67]$ & $+8.33\;[3.33,\,15.00]$ \\
Toggle & 4 and 6 & InternVL3.5-38B & 30 & $5.00\;[0.00,\,10.00]$ & $85.00\;[73.33,\,95.00]$ & $+80.00\;[70.00,\,90.00]$ \\
\addlinespace[0.2em]
Toggle & 4 & Qwen3-VL-32B & 15 & $0.00\;[0.00,\,0.00]$ & $86.67\;[73.33,\,96.67]$ & $+86.67\;[73.33,\,96.67]$ \\
Toggle & 4 & GLM-4.5V & 15 & $0.00\;[0.00,\,0.00]$ & $100.00\;[100.00,\,100.00]$ & $+100.00\;[100.00,\,100.00]$ \\
Toggle & 4 & Molmo2-8B & 15 & $0.00\;[0.00,\,0.00]$ & $3.33\;[0.00,\,10.00]$ & $+3.33\;[0.00,\,10.00]$ \\
Toggle & 4 & InternVL3.5-38B & 15 & $0.00\;[0.00,\,0.00]$ & $86.67\;[73.33,\,96.67]$ & $+86.67\;[73.33,\,96.67]$ \\
\addlinespace[0.2em]
Toggle & 6 & Qwen3-VL-32B & 15 & $3.33\;[0.00,\,10.00]$ & $90.00\;[80.00,\,100.00]$ & $+86.67\;[76.67,\,96.67]$ \\
Toggle & 6 & GLM-4.5V & 15 & $0.00\;[0.00,\,0.00]$ & $93.33\;[83.33,\,100.00]$ & $+93.33\;[83.33,\,100.00]$ \\
Toggle & 6 & Molmo2-8B & 15 & $3.33\;[0.00,\,10.00]$ & $16.67\;[6.67,\,30.00]$ & $+13.33\;[3.33,\,26.67]$ \\
Toggle & 6 & InternVL3.5-38B & 15 & $10.00\;[0.00,\,20.00]$ & $83.33\;[63.33,\,100.00]$ & $+73.33\;[53.33,\,90.00]$ \\
\bottomrule
\end{tabular*}

\end{table}
\FloatBarrier

\subsection{Score-level test of direct judgment}
\label{app:truth-switch-margin}

For pair $p$ and candidate order $o$, let $\mu(V)$ be the direct A-minus-B
next-token logit margin for witness video $V$; one video has correct answer A
and the other B within the order. The score-level outcome is
$M_{p,o}=\mathbf 1\{\mu(V_{\mathrm A})>\mu(V_{\mathrm B})\}$, with ties counted one
half, averaged over the two orders and the 60 pairs. Any fixed position or
candidate preference gives 50\%, and both-orders-correct gives 25\%.
Averaging each video's margin over the two candidate orders removes the
position bias and yields a single-video score whose sign should favor the
video's own trajectory; its accuracy over the 120 witness videos and the
corresponding paired success are the operational counterparts of the
within-pair test. \Cref{tab:truth-switch-margin} reports the four models,
with pointwise 95\% intervals from 10,000 stratified pair-bootstrap draws for
the within-pair test; these estimates are descriptive companions to the
answer-level families and are not part of them.
Qwen's margins are large and mostly positive (91\% of the 240 margins), so its
argmax answers are A on 219 of 240 requests although the margins separate the
two videos in 73.33\% of pair-orders. GLM's margins are compressed (standard
deviation 0.65 logits, all positive), InternVL's are coarsely quantized (15.8\%
of pair-orders tie exactly), and Molmo's are centered near zero with no
directional signal.

At the validated answer position (\cref{tab:revision-score}), Qwen's margins move in the
correct direction in 72.50\% of pair-orders (pointwise 95\% interval [63.75, 80.83]) and GLM's
in 62.50\% [53.33, 71.67]. The intervals for InternVL (58.75\%) and Molmo (54.17\%) include
chance. After averaging each video's margin over the two orders, single-video accuracy is
61.25\% for Qwen and 51--55\% for the others.

\begin{table}[h]
      \caption{\rev{\textbf{Score-level test of direct judgment (per-model protocol of the submission).} Percentages. The within-pair test has pointwise 95\% intervals. The de-biased columns use order-averaged margins, and the last column is the share of positive margins (position bias).}}
  \label{tab:truth-switch-margin}
  \centering
  \scriptsize
  \setlength{\tabcolsep}{3pt}
\begin{tabular*}{\linewidth}{@{\extracolsep{\fill}}lcccccc@{}}
\toprule
Model & Within-pair & De-biased single-video & De-biased paired & Direct paired & CoT paired & Margins $>0$ \\
\midrule
Qwen3-VL-32B & $73.33\;[65.42,\,81.25]$ & 60.00 & 16.67 & 5.83 & 75.00 & 91.25 \\
GLM-4.5V & $60.83\;[54.58,\,67.08]$ & 49.58 & 5.00 & 0.00 & 88.33 & 100.00 \\
\addlinespace[0.35em]
Molmo2-8B & $55.00\;[46.67,\,63.33]$ & 51.25 & 3.33 & 3.33 & 14.17 & 49.17 \\
InternVL3.5-38B & $59.58\;[51.25,\,67.92]$ & 52.08 & 6.67 & 4.17 & 88.33 & 15.00 \\
\bottomrule
\end{tabular*}

\end{table}
\begin{table}[ht]
\centering\small\revcolor
\caption{\rev{\textbf{Score-level test at the validated position.} Pointwise 95\% intervals.}}
\label{tab:revision-score}
\begin{tabular}{@{}lcc@{}}
\toprule
Model & Within-pair margin (\%) & Debiased single-video accuracy (\%) \\
\midrule
Qwen3-VL-32B & 72.50 [63.75, 80.83] & 61.25 \\
GLM-4.5V & 62.50 [53.33, 71.67] & 55.00 \\
InternVL3.5-38B & 58.75 [50.00, 67.50] & 51.25 \\
Molmo2-8B & 54.17 [46.67, 61.67] & 52.50 \\
\bottomrule
\end{tabular}
\end{table}

\FloatBarrier

\section{Natural-video study details}
\label{app:natural}

\subsection{Design, prompts, and parsing}
\label{app:natural-design}

The natural-video study uses the 400 videos and reference propositions of
\cref{app:provenance} and the negatives of \cref{sec:candidates}. Each cell
(model, interface, evidence condition, negative type) contains 800 requests:
400 videos in two candidate orders. Direct and CoT conditions use the same
task, candidate text, media, and order. The three evidence conditions are
ordered frames, the fixed shuffled-frame control, and no video.

Nominal frame counts are 64 for Qwen, 48 for GLM, 80 for Molmo, and 48 for
InternVL, followed by each model's native processor. Ordered and shuffled
inputs use the same decoded frame multiset within a model. The prompt, with the question and candidate propositions filled per video,
is:
\begin{quote}\small\ttfamily\raggedright
You are given \{N\} frames from a video in the displayed sequence.\\[0.3em]
Question: \{question\}\\[0.3em]
Choose the statement that more accurately describes the target video. Use the
supplied visual evidence when available. Evaluate the statements themselves,
including whether their events occur.\\[0.3em]
A. \{first proposition\}\\
B. \{second proposition\}\\[0.3em]
\{instruction\}
\end{quote}
Without video, the first line reads ``No video frames are provided.''
Ordered and shuffled media receive identical wording, which neither reveals
the control assignment nor implies that shuffled frames are chronological.
The direct instruction is ``Give your choice immediately. Reply with exactly
one token: A or B.'' The CoT instruction is ``Reason through the available
evidence before deciding. End with a separate final line exactly in the form
Answer: A or Answer: B.''

\emph{Per-model protocol of the submission.} CoT answers are read from the final \verb+Answer: A+ or \verb+Answer: B+ line,
allowing minor formatting variants (markdown emphasis, a trailing period, an
echoed proposition, and GLM's native answer wrapper); unparseable or
truncated outputs and tied direct maxima are scored as incorrect.

\begin{revisedtext}
\noindent\textbf{Common protocol.} Direct judgment is read at the validated answer position
(\cref{app:revision-position}). CoT judgment uses prompt G1, which requests a brief
evidence-based rationale, and the generated-answer control uses prompt G0, which requests a
concise answer without explanation. Both prompts require a separate final answer line;
\cref{app:revision-interface} gives the verbatim instructions. Scoring uses the last expressed
choice, and \cref{app:revision-natural} also reports accuracy with every cap hit counted as
wrong.
\end{revisedtext}

\subsection{Statistical families and complete results}
\label{app:natural-results}

\begin{figure}[ht]
\centering
\includegraphics[width=\linewidth]{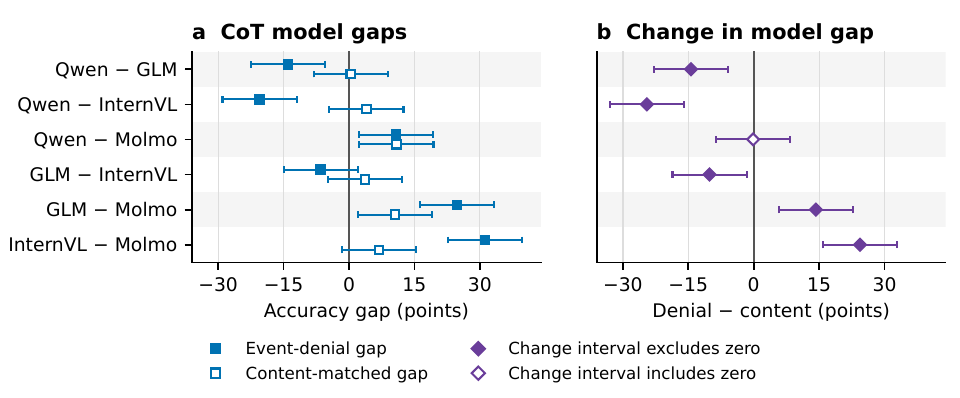}
\caption{\rev{\textbf{CoT model gaps and their changes with negative type.} (a) CoT accuracy gaps on ordered video against each negative type. (b) The event-denial gap minus the content-matched gap. Bars are simultaneous 95\% intervals (\cref{tab:natural-primary}).}}
\label{fig:natural-model-gaps}
\end{figure}

The common protocol uses a 12-estimate interface family
(\cref{tab:revision-natural}) and an 18-estimate model-gap family
(\cref{tab:natural-primary,fig:natural-model-gaps}). The interface family contains the
two CoT-minus-direct changes and their interaction for each model.
The model-gap family contains the two negative-specific CoT gaps and
their difference for each model pair. The submitted analysis used
10,000 shared paired draws within the four question strata, seed
20260911. Its interface and model-gap families are kept alongside
the submitted 16-estimate evidence and 18-estimate no-video-subtracted
families. Those CoT evidence contrasts retain the per-model protocol;
\cref{sec:natural-evidence} gives the validated direct controls.

\begin{table}[ht]
\centering\scriptsize\revcolor\setlength{\tabcolsep}{2.1pt}
\caption{\rev{\textbf{CoT minus direct judgment on natural video.} Accuracy under direct / CoT judgment (\%) on 400 ordered videos, and the change in points with simultaneous 95\% intervals. The last row is a separate sensitivity analysis with native thinking.}}
\label{tab:revision-natural}
\begin{tabular}{@{}lccccc@{}}
\toprule
 & \multicolumn{2}{c}{Event-denial} & \multicolumn{2}{c}{Content-matched} & Denial $-$ content \\
Model & Direct / CoT & Change & Direct / CoT & Change & Change \\
\midrule
Qwen3-VL-32B    & 66.62 / 43.12 & $-23.50$ [$-30.50$, $-16.50$] & 65.50 / 64.12 & $-1.38$ [$-8.38$, 5.63]  & $-22.12$ [$-29.12$, $-15.12$] \\
GLM-4.5V        & 63.25 / 57.12 & $-6.12$ [$-13.12$, 0.88]      & 59.00 / 63.75 & $+4.75$ [$-2.25$, 11.75] & $-10.88$ [$-17.88$, $-3.87$] \\
InternVL3.5-38B & 86.50 / 63.62 & $-22.88$ [$-29.88$, $-15.88$] & 54.50 / 60.12 & $+5.62$ [$-1.38$, 12.62] & $-28.50$ [$-35.50$, $-21.50$] \\
Molmo2-8B       & 37.50 / 32.38 & $-5.12$ [$-12.12$, 1.88]      & 59.38 / 53.25 & $-6.12$ [$-13.12$, 0.88] & $+1.00$ [$-6.00$, 8.00] \\
\midrule
GLM, native thinking & 63.25 / 44.12 & $-19.13$ [$-24.38$, $-13.88$] & 59.00 / 64.50 & $+5.50$ [0.25, 10.75] & $-24.63$ [$-29.88$, $-19.38$] \\
\bottomrule
\end{tabular}
\end{table}

\begin{table}[h]
  \caption{\rev{\textbf{Submitted CoT minus direct judgment under ordered video (per-model protocol of the submission).} Changes in points with simultaneous 95\% intervals. GLM's CoT cells contain 194 and 196 invalid or truncated responses out of 800, so its changes mix completion failure with judgment.}}
  \label{tab:natural-elicitation}
  \centering
  \footnotesize
  \begin{tabular*}{\linewidth}{@{\extracolsep{\fill}}lccc@{}}
\toprule
Model & Event-denial (pp) & Content-matched (pp) & Difference (pp) \\
\midrule
Qwen3-VL-32B & $-17.63\;[-24.63,\,-10.63]$ & $-0.50\;[-7.50,\,6.50]$ & $-17.13\;[-24.13,\,-10.13]$ \\
GLM-4.5V & $-32.13\;[-39.13,\,-25.13]$ & $-7.38\;[-14.38,\,-0.38]$ & $-24.75\;[-31.75,\,-17.75]$ \\
Molmo2-8B & $+6.38\;[-0.63,\,13.38]$ & $-5.38\;[-12.38,\,1.63]$ & $+11.75\;[4.75,\,18.75]$ \\
InternVL3.5-38B & $-8.00\;[-15.00,\,-1.00]$ & $+5.25\;[-1.75,\,12.25]$ & $-13.25\;[-20.25,\,-6.25]$ \\
\bottomrule
\end{tabular*}

\end{table}

\begin{table}[ht]
\centering\scriptsize\revcolor
\caption{\rev{\textbf{CoT model comparisons under the common protocol.} Differences in points with simultaneous 95\% intervals over 18 estimates.}}
\label{tab:natural-primary}
\begin{tabular}{@{}lccc@{}}
\toprule
Model pair & Event-denial & Content-matched & Denial $-$ content \\
\midrule
Qwen3-VL-32B $-$ GLM-4.5V & $-14.00\;[-22.50,-5.50]$ & $+0.38\;[-8.13,8.88]$ & $-14.38\;[-22.88,-5.88]$ \\
Qwen3-VL-32B $-$ InternVL3.5-38B & $-20.50\;[-29.00,-12.00]$ & $+4.00\;[-4.50,12.50]$ & $-24.50\;[-33.00,-16.00]$ \\
Qwen3-VL-32B $-$ Molmo2-8B & $+10.75\;[2.25,19.25]$ & $+10.88\;[2.38,19.38]$ & $-0.13\;[-8.63,8.38]$ \\
GLM-4.5V $-$ InternVL3.5-38B & $-6.50\;[-15.00,2.00]$ & $+3.63\;[-4.88,12.13]$ & $-10.13\;[-18.63,-1.63]$ \\
GLM-4.5V $-$ Molmo2-8B & $+24.75\;[16.25,33.25]$ & $+10.50\;[2.00,19.00]$ & $+14.25\;[5.75,22.75]$ \\
InternVL3.5-38B $-$ Molmo2-8B & $+31.25\;[22.75,39.75]$ & $+6.88\;[-1.63,15.38]$ & $+24.38\;[15.88,32.88]$ \\
\bottomrule
\end{tabular}
\end{table}

\begin{table}[ht]
\centering\scriptsize
\caption{\textbf{CoT model gaps (per-model protocol of the submission).}
Percentage-point gaps and simultaneous 95\% intervals.}
\label{tab:natural-primary-submitted}
\begin{tabular*}{\linewidth}{@{\extracolsep{\fill}}lccc@{}}
\toprule
Model pair & Event-denial gap & Content-matched gap & Change \\
\midrule
Qwen $-$ GLM & $+14.75\;[6.38,\,23.13]$ & $+15.63\;[7.25,\,24.00]$ & $-0.88\;[-9.25,\,7.50]$ \\
Qwen $-$ Molmo & $+6.75\;[-1.63,\,15.13]$ & $+11.75\;[3.38,\,20.13]$ & $-5.00\;[-13.38,\,3.38]$ \\
Qwen $-$ InternVL & $-30.13\;[-38.50,\,-21.75]$ & $+4.50\;[-3.88,\,12.88]$ & $-34.63\;[-43.00,\,-26.25]$ \\
GLM $-$ Molmo & $-8.00\;[-16.38,\,0.38]$ & $-3.88\;[-12.25,\,4.50]$ & $-4.13\;[-12.50,\,4.25]$ \\
GLM $-$ InternVL & $-44.88\;[-53.25,\,-36.50]$ & $-11.13\;[-19.50,\,-2.75]$ & $-33.75\;[-42.13,\,-25.38]$ \\
Molmo $-$ InternVL & $-36.88\;[-45.25,\,-28.50]$ & $-7.25\;[-15.63,\,1.13]$ & $-29.63\;[-38.00,\,-21.25]$ \\
\bottomrule
\end{tabular*}

\end{table}

The CoT evidence family shows significant ordered-minus-no-video effects for Qwen
on both negative types, Molmo on both, and InternVL on denial
(\cref{tab:natural-evidence}). The
no-video-subtracted comparison repeats the CoT between-model analysis after
subtracting no-video accuracy within model and negative type
(\cref{tab:natural-nonvisual}). It retains three significant negative-type
changes.

\begin{table}[h]
  \caption{\rev{\textbf{CoT evidence contrasts (per-model protocol of the submission).} Percentage-point differences
  between CoT ordered-video accuracy and CoT no-video or shuffled accuracy,
  with simultaneous 95\% intervals.}}
  \label{tab:natural-evidence}
  \centering
  \scriptsize
  \begin{tabular*}{\linewidth}{@{\extracolsep{\fill}}lcccc@{}}
\toprule
& \multicolumn{2}{c}{Event-denial negatives} & \multicolumn{2}{c}{Content-matched negatives} \\
\cmidrule(lr){2-3}\cmidrule(lr){4-5}
Model & Ordered $-$ no video & Ordered $-$ shuffled & Ordered $-$ no video & Ordered $-$ shuffled \\
\midrule
Qwen3-VL-32B & $+21.50\;[14.50,\,28.50]$ & $+3.88\;[-3.13,\,10.88]$ & $+11.38\;[4.38,\,18.38]$ & $+2.50\;[-4.50,\,9.50]$ \\
GLM-4.5V & $+5.50\;[-1.50,\,12.50]$ & $-2.38\;[-9.38,\,4.63]$ & $+0.50\;[-6.50,\,7.50]$ & $+1.13\;[-5.88,\,8.13]$ \\
Molmo2-8B & $+10.00\;[3.00,\,17.00]$ & $+6.50\;[-0.50,\,13.50]$ & $+7.75\;[0.75,\,14.75]$ & $+2.25\;[-4.75,\,9.25]$ \\
InternVL3.5-38B & $+42.13\;[35.13,\,49.13]$ & $+1.38\;[-5.63,\,8.38]$ & $+6.88\;[-0.13,\,13.88]$ & $+1.63\;[-5.38,\,8.63]$ \\
\bottomrule
\end{tabular*}

\end{table}

\begin{table}[h]
  \caption{\rev{\textbf{No-video-subtracted CoT comparisons (per-model protocol of the submission).} As in \cref{tab:natural-primary-submitted}, after subtracting each model's no-video accuracy within negative type. Simultaneous 95\% intervals.}}
  \label{tab:natural-nonvisual}
  \centering
  \small
  \begin{tabular*}{\linewidth}{@{\extracolsep{\fill}}lccc@{}}
\toprule
Model pair & Event-denial gap & Content-matched gap & Change \\
\midrule
Qwen $-$ GLM & $+16.00\;[5.24,\,26.76]$ & $+10.88\;[0.12,\,21.63]$ & $+5.13\;[-5.63,\,15.88]$ \\
Qwen $-$ Molmo & $+11.50\;[0.74,\,22.26]$ & $+3.63\;[-7.13,\,14.38]$ & $+7.88\;[-2.88,\,18.63]$ \\
Qwen $-$ InternVL & $-20.63\;[-31.38,\,-9.87]$ & $+4.50\;[-6.26,\,15.26]$ & $-25.13\;[-35.88,\,-14.37]$ \\
GLM $-$ Molmo & $-4.50\;[-15.26,\,6.26]$ & $-7.25\;[-18.01,\,3.51]$ & $+2.75\;[-8.01,\,13.51]$ \\
GLM $-$ InternVL & $-36.63\;[-47.38,\,-25.87]$ & $-6.38\;[-17.13,\,4.38]$ & $-30.25\;[-41.01,\,-19.49]$ \\
Molmo $-$ InternVL & $-32.13\;[-42.88,\,-21.37]$ & $+0.88\;[-9.88,\,11.63]$ & $-33.00\;[-43.76,\,-22.24]$ \\
\bottomrule
\end{tabular*}

\end{table}

\begin{table}[p]
  \caption{\textbf{All 48 natural-video accuracy cells (per-model protocol of the submission).} Accuracy averages
  two candidate orders within each of 400 videos; brackets are pointwise 95\%
  intervals.}
  \label{tab:natural-cells}
  \centering
  \scriptsize
  \begin{tabular}{@{}lllcc@{}}
\toprule
Model & Interface & Evidence & Event-denial accuracy (\%) & Content-matched accuracy (\%) \\
\midrule
Qwen3-VL-32B & Direct & Ordered & $61.38\;[57.38,\,65.38]$ & $64.38\;[60.38,\,68.13]$ \\
 &  & Shuffled & $56.38\;[52.25,\,60.38]$ & $63.50\;[59.75,\,67.13]$ \\
 &  & No video & $2.88\;[1.88,\,4.00]$ & $58.63\;[55.00,\,62.25]$ \\
 & CoT & Ordered & $43.75\;[39.63,\,47.88]$ & $63.88\;[59.75,\,67.75]$ \\
 &  & Shuffled & $39.88\;[35.63,\,44.00]$ & $61.38\;[57.25,\,65.50]$ \\
 &  & No video & $22.25\;[19.38,\,25.13]$ & $52.50\;[48.63,\,56.50]$ \\
\addlinespace[0.2em]
GLM-4.5V & Direct & Ordered & $61.13\;[58.00,\,64.38]$ & $55.63\;[52.63,\,58.63]$ \\
 &  & Shuffled & $61.13\;[58.00,\,64.38]$ & $54.38\;[51.63,\,57.13]$ \\
 &  & No video & $43.13\;[40.75,\,45.63]$ & $50.13\;[48.13,\,52.13]$ \\
 & CoT & Ordered & $29.00\;[25.50,\,32.50]$ & $48.25\;[44.25,\,52.25]$ \\
 &  & Shuffled & $31.38\;[28.00,\,34.75]$ & $47.13\;[43.38,\,51.00]$ \\
 &  & No video & $23.50\;[21.25,\,25.75]$ & $47.75\;[44.13,\,51.38]$ \\
\addlinespace[0.2em]
Molmo2-8B & Direct & Ordered & $30.63\;[27.38,\,33.88]$ & $57.50\;[53.63,\,61.38]$ \\
 &  & Shuffled & $28.88\;[25.75,\,32.00]$ & $56.75\;[52.88,\,60.63]$ \\
 &  & No video & $40.63\;[38.50,\,42.88]$ & $53.75\;[49.75,\,57.75]$ \\
 & CoT & Ordered & $37.00\;[33.25,\,40.75]$ & $52.13\;[48.25,\,56.00]$ \\
 &  & Shuffled & $30.50\;[26.88,\,34.25]$ & $49.88\;[45.88,\,54.00]$ \\
 &  & No video & $27.00\;[24.13,\,29.75]$ & $44.38\;[40.63,\,48.13]$ \\
\addlinespace[0.2em]
InternVL3.5-38B & Direct & Ordered & $81.88\;[78.50,\,85.13]$ & $54.13\;[49.75,\,58.50]$ \\
 &  & Shuffled & $84.63\;[81.38,\,87.63]$ & $55.50\;[51.63,\,59.50]$ \\
 &  & No video & $66.88\;[63.75,\,70.00]$ & $51.75\;[48.75,\,54.75]$ \\
 & CoT & Ordered & $73.88\;[70.13,\,77.50]$ & $59.38\;[55.25,\,63.38]$ \\
 &  & Shuffled & $72.50\;[68.88,\,76.00]$ & $57.75\;[53.63,\,61.88]$ \\
 &  & No video & $31.75\;[28.88,\,34.63]$ & $52.50\;[48.63,\,56.38]$ \\
\bottomrule
\end{tabular}

\end{table}
\begin{table}[ht]
\centering\scriptsize\revcolor
\caption{\rev{\textbf{Accuracy under the common protocol.} Percentages with pointwise 95\% intervals.}}
\label{tab:revision-evidence}
\begin{tabular}{@{}llcccc@{}}
\toprule
Model & Evidence & Direct denial & Direct content & CoT denial & CoT content \\
\midrule
Qwen3-VL-32B & Ordered & $66.62\;[62.50,70.75]$ & $65.50\;[61.37,69.50]$ & $43.12\;[39.00,47.25]$ & $64.12\;[60.00,68.13]$ \\
Qwen3-VL-32B & Shuffled & $61.00\;[56.75,65.25]$ & $65.88\;[61.88,69.75]$ &  &  \\
Qwen3-VL-32B & No video & $8.25\;[6.50,10.13]$ & $55.88\;[52.25,59.50]$ &  &  \\
GLM-4.5V & Ordered & $63.25\;[59.50,67.00]$ & $59.00\;[55.13,62.75]$ & $57.12\;[53.13,61.13]$ & $63.75\;[59.50,67.88]$ \\
GLM-4.5V & Shuffled & $64.50\;[60.63,68.13]$ & $57.50\;[53.88,61.13]$ &  &  \\
GLM-4.5V & No video & $51.00\;[47.88,54.25]$ & $57.75\;[54.13,61.38]$ &  &  \\
InternVL3.5-38B & Ordered & $86.50\;[83.75,89.25]$ & $54.50\;[50.25,58.50]$ & $63.62\;[59.63,67.50]$ & $60.12\;[56.12,64.00]$ \\
InternVL3.5-38B & Shuffled & $88.38\;[85.75,90.75]$ & $55.88\;[51.88,59.63]$ &  &  \\
InternVL3.5-38B & No video & $74.62\;[72.00,77.25]$ & $53.63\;[50.13,57.00]$ &  &  \\
Molmo2-8B & Ordered & $37.50\;[34.50,40.50]$ & $59.38\;[56.00,62.75]$ & $32.38\;[28.75,36.00]$ & $53.25\;[49.13,57.25]$ \\
Molmo2-8B & Shuffled & $36.13\;[33.13,39.13]$ & $58.75\;[55.38,62.13]$ &  &  \\
Molmo2-8B & No video & $37.38\;[35.13,39.50]$ & $54.00\;[50.13,57.88]$ &  &  \\
\bottomrule
\end{tabular}
\end{table}

\FloatBarrier
\begingroup\scriptsize
\providecommand{\rev}[1]{#1}
\setlength{\tabcolsep}{3.5pt}
\begin{longtable}{@{}llllrrrrrrrr@{}}
\caption{\rev{\textbf{Output diagnostics for all 48 natural-video cells (per-model protocol of the submission).} 800 requests per cell: 400 videos in two candidate orders.}\label{tab:natural-diagnostics}} \\
\toprule
Model & Interface & Evidence & Negative & Correct & Invalid & Trunc. & Ties & Fmt fail & Fmt corr. & Ord. corr. & Ord. ans. \\
\midrule
\endfirsthead
\caption[]{\rev{\textbf{Output diagnostics for all 48 natural-video cells (per-model protocol of the submission).} Continued.}} \\
\toprule
Model & Interface & Evidence & Negative & Correct & Invalid & Trunc. & Ties & Fmt fail & Fmt corr. & Ord. corr. & Ord. ans. \\
\midrule
\endhead
Qwen & Direct & Ordered & Denial & 491 & 12 & 0 & 12 & 12 & 491 & 69 & 76 \\
Qwen & Direct & Ordered & Content & 515 & 18 & 0 & 18 & 18 & 515 & 107 & 114 \\
Qwen & Direct & Shuffled & Denial & 451 & 10 & 0 & 10 & 10 & 451 & 81 & 86 \\
Qwen & Direct & Shuffled & Content & 508 & 13 & 0 & 13 & 13 & 508 & 144 & 146 \\
Qwen & Direct & No video & Denial & 23 & 3 & 0 & 3 & 3 & 23 & 23 & 26 \\
Qwen & Direct & No video & Content & 469 & 7 & 0 & 7 & 7 & 469 & 143 & 147 \\
Qwen & CoT & Ordered & Denial & 350 & 9 & 1 & 0 & 9 & 350 & 104 & 106 \\
Qwen & CoT & Ordered & Content & 511 & 1 & 3 & 0 & 1 & 511 & 101 & 102 \\
Qwen & CoT & Shuffled & Denial & 319 & 6 & 0 & 0 & 6 & 319 & 79 & 81 \\
Qwen & CoT & Shuffled & Content & 491 & 2 & 10 & 0 & 2 & 491 & 99 & 106 \\
Qwen & CoT & No video & Denial & 178 & 0 & 0 & 0 & 0 & 178 & 104 & 104 \\
Qwen & CoT & No video & Content & 420 & 1 & 0 & 0 & 1 & 420 & 122 & 122 \\
GLM & Direct & Ordered & Denial & 489 & 5 & 0 & 5 & 5 & 489 & 157 & 160 \\
GLM & Direct & Ordered & Content & 445 & 6 & 0 & 6 & 6 & 445 & 247 & 250 \\
GLM & Direct & Shuffled & Denial & 489 & 4 & 0 & 4 & 4 & 489 & 165 & 167 \\
GLM & Direct & Shuffled & Content & 435 & 5 & 0 & 5 & 5 & 435 & 263 & 266 \\
GLM & Direct & No video & Denial & 345 & 11 & 0 & 11 & 11 & 345 & 231 & 237 \\
GLM & Direct & No video & Content & 401 & 7 & 0 & 7 & 7 & 401 & 329 & 335 \\
GLM & CoT & Ordered & Denial & 232 & 184 & 10 & 0 & 727 & 28 & 92 & 168 \\
GLM & CoT & Ordered & Content & 386 & 179 & 17 & 0 & 704 & 45 & 112 & 158 \\
GLM & CoT & Shuffled & Denial & 251 & 172 & 10 & 0 & 727 & 26 & 91 & 148 \\
GLM & CoT & Shuffled & Content & 377 & 189 & 8 & 0 & 708 & 54 & 135 & 183 \\
GLM & CoT & No video & Denial & 188 & 7 & 0 & 0 & 323 & 143 & 70 & 70 \\
GLM & CoT & No video & Content & 382 & 15 & 0 & 0 & 200 & 298 & 156 & 168 \\
Molmo & Direct & Ordered & Denial & 245 & 0 & 0 & 0 & 0 & 245 & 87 & 87 \\
Molmo & Direct & Ordered & Content & 460 & 0 & 0 & 0 & 0 & 460 & 142 & 142 \\
Molmo & Direct & Shuffled & Denial & 231 & 0 & 0 & 0 & 0 & 231 & 87 & 87 \\
Molmo & Direct & Shuffled & Content & 454 & 0 & 0 & 0 & 0 & 454 & 138 & 138 \\
Molmo & Direct & No video & Denial & 325 & 0 & 0 & 0 & 0 & 325 & 73 & 73 \\
Molmo & Direct & No video & Content & 430 & 1 & 0 & 1 & 1 & 430 & 124 & 124 \\
Molmo & CoT & Ordered & Denial & 296 & 16 & 5 & 0 & 113 & 277 & 120 & 131 \\
Molmo & CoT & Ordered & Content & 417 & 43 & 3 & 0 & 81 & 390 & 139 & 154 \\
Molmo & CoT & Shuffled & Denial & 244 & 19 & 1 & 0 & 140 & 227 & 96 & 108 \\
Molmo & CoT & Shuffled & Content & 399 & 53 & 4 & 0 & 102 & 376 & 121 & 147 \\
Molmo & CoT & No video & Denial & 216 & 18 & 0 & 0 & 58 & 202 & 110 & 126 \\
Molmo & CoT & No video & Content & 355 & 75 & 0 & 0 & 175 & 306 & 155 & 177 \\
InternVL & Direct & Ordered & Denial & 655 & 11 & 0 & 11 & 11 & 655 & 51 & 54 \\
InternVL & Direct & Ordered & Content & 433 & 34 & 0 & 34 & 34 & 433 & 81 & 99 \\
InternVL & Direct & Shuffled & Denial & 677 & 9 & 0 & 9 & 9 & 677 & 43 & 47 \\
InternVL & Direct & Shuffled & Content & 444 & 44 & 0 & 44 & 44 & 444 & 112 & 133 \\
InternVL & Direct & No video & Denial & 535 & 40 & 0 & 40 & 40 & 535 & 117 & 124 \\
InternVL & Direct & No video & Content & 414 & 40 & 0 & 40 & 40 & 414 & 232 & 244 \\
InternVL & CoT & Ordered & Denial & 591 & 1 & 0 & 0 & 1 & 591 & 85 & 85 \\
InternVL & CoT & Ordered & Content & 475 & 2 & 0 & 0 & 2 & 475 & 107 & 107 \\
InternVL & CoT & Shuffled & Denial & 580 & 1 & 0 & 0 & 1 & 580 & 104 & 105 \\
InternVL & CoT & Shuffled & Content & 462 & 0 & 5 & 0 & 0 & 462 & 106 & 108 \\
InternVL & CoT & No video & Denial & 254 & 0 & 101 & 0 & 0 & 254 & 138 & 173 \\
InternVL & CoT & No video & Content & 420 & 1 & 19 & 0 & 1 & 420 & 122 & 127 \\
\bottomrule
\addlinespace[3pt]
\multicolumn{12}{@{}l@{}}{\parbox{\linewidth}{\footnotesize \rev{\textit{Note.}
Fmt is the requested format. Invalid counts exclude truncations and include direct ties.
Format columns use the requested \texttt{Answer:} line only. Order columns count videos
whose correctness or chosen label differs between the two candidate orders.}}} \\
\end{longtable}
\endgroup

Stratum-level accuracies for all 192 cells are provided in the supplementary
CSV.
\FloatBarrier

\subsection{Why CoT judgment loses on event-denial negatives: a rationale analysis}
\label{app:trace-audit}

\begin{revisedtext}
The rationale analysis uses Qwen under the per-model protocol of the
submission (61.38\% direct and 43.75\% CoT). The corresponding common
protocol values are 66.62\% and 43.12\%.
\end{revisedtext}

This analysis asks why Qwen's ordered-video accuracy on event-denial negatives
falls from 61.38\% under direct judgment to 43.75\% under CoT judgment while
content-matched accuracy changes by $-0.50$ points. It uses the 800 CoT and
800 direct responses of the ordered-video event-denial cells, the same model's responses under the other
evidence conditions and under the content-matched pair, and InternVL's denial
cells as a comparator. The analyses, the rationale codebook, and the coding sample were fixed before any rationale was
read.

\paragraph{Where the decline lives.} At video level, after averaging the two
candidate orders, 54 videos move from correct to wrong and 12 from wrong to
correct (\cref{tab:trace-transitions}); 53 of the 54 losses receive the denial
in both orders. Losses concentrate in the event-relative-state and
visibility-continuity strata and, across strata, in anchor events that
presuppose a prior absence (\cref{tab:trace-anchor}): on the 83 videos whose
anchor event reads ``comes back into view'', accuracy falls from 73.5\% to
33.1\%, whereas ``gets partly hidden'' and ``leaves the frame'' anchor events
change by less than five points. Video duration is unrelated to the losses. Without video, both
interfaces favor the denial (direct 96.8\% of responses, CoT 77.8\%); with
ordered video the direct denial choice falls to 37.1\% and the CoT choice only
to 55.0\%.

\begin{table}[h]
  \caption{\textbf{Video-level transitions from direct to CoT judgment,
  event-denial negatives, ordered video.} Each video's correctness is averaged
  over the two candidate orders; partial rows are videos with one correct
  order under either interface.}
  \label{tab:trace-transitions}
  \centering
  \small
  \setlength{\tabcolsep}{3pt}
\begin{tabular*}{\linewidth}{@{\extracolsep{\fill}}lrrrrrrcc@{}}
\toprule
Stratum & Videos & Both right & Loss & Gain & Both wrong & Partial & Direct (\%) & CoT (\%) \\
\midrule
Order & 120 & 24 & 12 & 3 & 27 & 54 & 57.92 & 41.67 \\
State & 100 & 28 & 25 & 1 & 13 & 33 & 77.50 & 44.50 \\
Location & 100 & 17 & 4 & 7 & 36 & 36 & 39.00 & 38.50 \\
Visibility & 80 & 28 & 13 & 1 & 10 & 28 & 74.38 & 52.50 \\
\midrule
All & 400 & 97 & 54 & 12 & 86 & 151 & 61.38 & 43.75 \\
\bottomrule
\end{tabular*}

\end{table}

\begin{table}[h]
  \caption{\textbf{Accuracy by the phrasing of the anchor event.}
  Ordered video, both interfaces; the change is CoT minus direct in percentage
  points; loss and gain counts are videos.}
  \label{tab:trace-anchor}
  \centering
  \small
  \begin{tabular*}{\linewidth}{@{\extracolsep{\fill}}lrcccrr@{}}
\toprule
Anchor-event phrasing & Videos & Direct (\%) & CoT (\%) & Change (pp) & Losses & Gains \\
\midrule
``comes back into view'' & 83 & 73.49 & 33.13 & $-40.36$ & 23 & 0 \\
``gets partly hidden'' & 80 & 58.75 & 55.63 & $-3.12$ & 5 & 5 \\
``becomes non-visible'' & 72 & 71.53 & 50.69 & $-20.84$ & 11 & 1 \\
``enters or first shows up'' & 64 & 71.88 & 54.69 & $-17.19$ & 6 & 2 \\
``gets fully hidden'' & 51 & 38.24 & 20.59 & $-17.65$ & 8 & 0 \\
``leaves the frame'' & 25 & 34.00 & 30.00 & $-4.00$ & 0 & 1 \\
``disappears for good'' & 23 & 45.65 & 54.35 & $+8.70$ & 1 & 3 \\
other & 2 & 75.00 & 50.00 & $-25.00$ & 0 & 0 \\
\bottomrule
\end{tabular*}

\end{table}

\paragraph{What the wrong rationales say.} All 440 wrong CoT answers were coded under the four categories of \cref{tab:trace-codes}. An
LLM assistant read a seeded sample of 60 wrong and 30 correct responses in full. The remaining wrong
responses were screened for conflict language and explicit absence assertions. The assistant read every flagged or unclassified rationale outside the sample in full (89 more).
It also checked 30 seeded excerpts from the explicit-assertion stratum and found 29 absence
conclusions and one timing misreading. Three excerpts overlap the initial sample. The check adds 27 wrong responses, for
176 wrong responses read individually. The remaining 264 explicit-assertion rationales were
assigned that code by rule. Of the 440 wrong CoT answers, 396 conclude that the anchor event
never happens; 23 acknowledge the event, dispute a detail of the faithful statement,
and choose the denial by elimination; 16 are long oscillating rationales
(median 17,070 characters against 666 for the first category) whose final
letter contradicts their own evaluation; and 5 misread the denial as a timing
claim.

Among the 132 absence conclusions read individually, 31 first locate the event at
a specific frame and then reason to its denial after disputing a detail; the
others report that the entity never left view or was already present (27),
that the event fails a strict threshold such as ``fully hidden'' (21), that it
never happens (42), that no reappearance was found (7), or that the entity was
not found (7). Mechanism disputes dominate the visibility-continuity losses:
the rationale affirms the disappearance but rejects the reference's mechanism
(occluded within the frame versus leaving it) and still picks the statement
that denies the event.

\begin{table}[h]
  \caption{\textbf{Codes for the 440 wrong CoT answers.} ``Read individually'' counts the rationales read in full or in the excerpt check; the other 264
answers of code C1 were assigned by rule from the explicit-assertion stratum described in the
text.}
  \label{tab:trace-codes}
  \centering
  \small
  \begin{tabular*}{\linewidth}{@{\extracolsep{\fill}}llrr@{}}
\toprule
Code & Description & All 440 & Read individually \\
\midrule
C1 & Concludes that the anchor event never happens & 396 & 132 \\
C2 & Acknowledges the event, disputes a detail, denies by elimination & 23 & 23 \\
C4 & Oscillating rationale whose final letter contradicts its evaluation & 16 & 16 \\
C3 & Reads the denial as a claim about timing & 5 & 5 \\
\midrule
 & All wrong CoT answers & 440 & 176 \\
\bottomrule
\end{tabular*}

\end{table}

\paragraph{The same frames, a different negative.} Among the 54 loss
videos, CoT judgment against the content-matched negative is fully correct
on 36 and half-correct on 13. Eight loss videos were read under both pairs;
in all eight the content-matched-pair rationale locates the event that the
denial-pair rationale declared absent. \Cref{tab:trace-example} shows one such
pair chosen by a fixed rule. We chose the bottle clip from these eight cases
for \cref{fig:bottle-occlusion} because adjacent sampled frames show the
occlusion and reappearance. The figure quotes the faithful-second responses
under the per-model protocol of the submission.

\begin{table}[!ht]
  \caption{\rev{\textbf{Same video, same frames, two candidate pairs.} Qwen's CoT rationales against the event-denial negative (left) and the content-matched negative (right), verbatim with omissions marked.}}
  \label{tab:trace-example}
  \centering
  \small
  \begin{tabularx}{\linewidth}{@{}YY@{}}
\toprule
\multicolumn{2}{@{}p{\linewidth}@{}}{\footnotesize\textit{Question:} At the moment the wooden cabinet comes back into view, what is the state of the blue pen? \quad \textit{A:} faithful statement in both pairs; \textit{B:} The anchor event 'the wooden cabinet comes back into view' does not occur anywhere in the video. (left) or At the event 'the wooden cabinet comes back into view', the entity 'the blue pen' is partly hidden. (right).} \\
\midrule
\footnotesize\textbf{Event-denial pair} & \footnotesize\textbf{Content-matched pair} \\
\footnotesize The wooden cabinet is visible throughout the entire video sequence, from Frame 0 to Frame 63. Therefore, the event ``the wooden cabinet comes back into view'' does not occur as a distinct event—it is continuously present. The phrase ``comes back into view'' implies it was previously out of view, which is not the case here. \newline Thus, the anchor event does not occur. \newline Answer: B & \footnotesize The event ``the wooden cabinet comes back into view'' occurs at the beginning of the video, specifically in Frame 0, where the cabinet is clearly visible in the background. […] There is no object or hand covering it, and it remains in the same position throughout the initial frames. In Frame 0, the blue pen is clearly visible on the table, and the cabinet is in full view. The pen remains visible in subsequent frames (e.g., Frame 1, 2, etc.) as the hand interacts with it, but at the moment the cabinet comes into view (Frame 0), the pen is fully visible. […] Answer: A \\
\bottomrule
\end{tabularx}

\end{table}

\begin{revisedtext}
\paragraph{Reading.} Some submitted rationales reject a queried detail or a
presupposition, such as the prior absence implied by \emph{comes back}, and
select the statement denying the whole event. Content-matched candidates
both assert the event and do not offer that choice. The example codes
describe submitted answers. The common-protocol comparison below separates
the effect of requesting a rationale from the effect of its content.
\end{revisedtext}

\paragraph{Frame-level check.} A short absence can fall between the 64
uniformly sampled frames that Qwen sees. For the 83 videos whose anchor event reads ``comes back into view'', the 64 frames presented to
Qwen were rendered as labelled contact sheets and each video was classified, without reference
to any model output or loss status, as A (the named object is out of view in at least one
sampled frame and visible again later), B (visible in every sampled frame), C (leaves view
without returning), or D (unclear). An LLM assistant proposed the classification with the
absent and return frames noted, and one of the authors reviewed every item and accepted all 83.
The two independent annotators of \cref{app:revision-validation} later agreed with it: both
mark all 47 A-videos visible and all 31 B/C-videos not visible at 64 frames. The counts are 47
in A, 25 in B, 6 in C, and 5 in D. Of the 23 loss videos, 15 are A, 7 B, and 1 D.
Qwen's ordered-denial accuracy on the
47 A-videos is 85.1\% direct and 41.5\% CoT; on the 31 B/C-videos it is 62.9\%
and 25.8\%. The decline persists where the reappearance is visible in the
model's input. On the B/C-videos the sampled frames do not show the
reappearance the reference asserts, so neither interface can be scored against
the frames there; these videos hold 7 of the 23 losses. The complete
tables, per-response codes, quotations, and code are in the artifact's
\texttt{data/m11-trace-audit} folder, and the frame-check manifest, responses,
and analysis under \texttt{data/m11-frame-check}.

\begin{revisedtext}
Under the common protocol, \cref{app:revision-natural} separates the request for a rationale from its
content. For InternVL the decline follows the rationale: read under one fixed prompt, its
event-denial accuracy is 79.88\% before the rationale and 61.25\% after it. For Qwen most
of the shift is present as soon as a rationale is requested: read before any rationale
token is written, its accuracy is 48.00\%, against 61.75\% for a concise generated answer
and 43.12\% after the full rationale. The codes of \cref{tab:trace-codes} describe the route the
rationales take to the event-denial choice.
\end{revisedtext}

\FloatBarrier

\section{Held-out verifier-selection analysis}
\label{app:selection}

\subsection{Objective, policies, and partitions}
\label{app:selection-protocol}

The target is ordered-video faithful-versus-content-matched accuracy on the
same 400 videos. Eligible sets contain all eight configurations, the
four direct configurations, or the four CoT configurations. For an eligible
set $\mathcal K$ and selection videos $S$,
\begin{equation}
\hat k_{\mathrm{den}}(S)\in\operatorname*{arg\,max}_{k\in\mathcal K}\hat A_k^{\mathrm{den}}(S),\qquad
\hat k_{\mathrm{cm}}(S)\in\operatorname*{arg\,max}_{k\in\mathcal K}\hat A_k^{\mathrm{cm}}(S),
\label{eq:selection}
\end{equation}
where $\hat A_k^{\mathrm{den}}$ and $\hat A_k^{\mathrm{cm}}$ are stratum-weighted event-denial and
content-matched accuracies on $S$; tied maximizers are selected uniformly.
Both selected configurations are evaluated by content-matched accuracy on the
held-out videos. Twenty stratified five-fold partitions (seed 20260916) give
evaluation folds of 24/20/20/16 videos across the four strata; every response
of a video stays with that video.

\subsection{Uncertainty and selection frequencies}
\label{app:selection-uncertainty}

Uncertainty uses 10,000 whole-video stratified bootstrap draws (seed
20260917); each duplicated video keeps its fold assignment, and selection on the
selection folds is repeated under the bootstrap weights, so the intervals include
selection variability. The three selection costs form one
simultaneous family (\cref{tab:selection-detail}). Fold-to-fold variation is
not an independent-sample standard error \citep{bates2023crossvalidation}, so the range and standard deviation over the twenty partitions are descriptive
only. \Cref{tab:selection-frequency} lists the configurations each policy
selects.

\begin{table}[h]
  \caption{\rev{\textbf{Repeated cross-validation summaries.} Selection costs in points. Range and SD describe the twenty partitions. The last two columns give the two policies' accuracy on held-out event-denial comparisons.}}
  \label{tab:selection-detail}
  \centering
  \scriptsize
  \begin{tabular*}{\linewidth}{@{\extracolsep{\fill}}lccccccc@{}}
\toprule
& & & & \multicolumn{2}{c}{20 partitions} & \multicolumn{2}{c}{Denial accuracy (\%)} \\
\cmidrule(lr){5-6}\cmidrule(lr){7-8}
Eligible set & Cost & Pointwise 95\% & Simult.\ 95\% & Range & SD & Denial-sel. & Content-sel. \\
\midrule
All eight & $+8.88$ & $[4.73,\,14.13]$ & $[2.46,\,15.30]$ & $7.75$--$10.25$ & $0.72$ & $81.88$ & $55.82$ \\
Four direct & $+10.25$ & $[5.88,\,14.25]$ & $[3.83,\,16.67]$ & $10.25$--$10.25$ & $0.00$ & $81.88$ & $61.38$ \\
Four CoT & $+4.50$ & $[-2.17,\,8.88]$ & $[-1.92,\,10.92]$ & $4.50$--$4.50$ & $0.00$ & $73.88$ & $43.75$ \\
\bottomrule
\end{tabular*}

\end{table}

\begin{table}[h]
  \caption{\textbf{Configuration selection frequencies} over the 100
  selections (20 partitions $\times$ 5 held-out folds) for each eligible set
  and criterion.}
  \label{tab:selection-frequency}
  \centering
  \small
  \begin{tabular*}{\linewidth}{@{\extracolsep{\fill}}lll@{}}
\toprule
Eligible set & Selection criterion & Configurations chosen in 100 selections \\
\midrule
All eight & Event-denial & Direct InternVL (100\%) \\
 & Content-matched & Direct Qwen (70\%), CoT Qwen (30\%) \\
Four direct & Event-denial & Direct InternVL (100\%) \\
 & Content-matched & Direct Qwen (100\%) \\
Four CoT & Event-denial & CoT InternVL (100\%) \\
 & Content-matched & CoT Qwen (100\%) \\
\bottomrule
\end{tabular*}

\end{table}
\FloatBarrier

\subsection{Separately fixed split}
\label{app:selection-split}

As a sensitivity check, a separately fixed split (seed 20260918) selects on
200 videos and evaluates on the other 200, with uncertainty from 10,000
bootstrap draws of the evaluation videos (seed 20261017) conditional on the
selected configurations; the three costs form their own simultaneous
family (\cref{tab:selection-split}). The direct-only cost remains
significant; the all-eight and CoT-only costs are not significant.

\begin{table}[h]
  \caption{\textbf{Selection on a separately fixed 200/200 split.} Selected
  configurations (denial-selected / content-matched-selected; -dir direct, -cot
  CoT), their content-matched accuracy on the 200 evaluation videos, and the
  selection cost with approximate simultaneous, pointwise, and bounded 95\%
  intervals.}
  \label{tab:selection-split}
  \centering
  \scriptsize
  \begin{tabular*}{\linewidth}{@{\extracolsep{\fill}}llcccc@{}}
\toprule
Eligible set & Selected & Target (\%) & Cost (pp) & Pointwise 95\% & Bounded 95\% \\
\midrule
All eight & InternVL-dir / Qwen-cot & $54.75$ / $59.50$ & $+4.75\;[-2.75,\,12.25]$ & $[-1.75,\,11.50]$ & $[-17.13,\,26.63]$ \\
Four direct & InternVL-dir / Qwen-dir & $54.75$ / $64.00$ & $+9.25\;[1.75,\,16.75]$ & $[3.50,\,15.00]$ & $[-12.63,\,31.13]$ \\
Four CoT & InternVL-cot / Qwen-cot & $57.50$ / $59.50$ & $+2.00\;[-5.50,\,9.50]$ & $[-4.25,\,8.25]$ & $[-19.88,\,23.88]$ \\
\bottomrule
\end{tabular*}

\end{table}
\FloatBarrier

\begin{revisedtext}
\Cref{app:revision-selection} reports the 21 splits, interval widths, the 26-configuration set, and cross-panel selection.
\end{revisedtext}

\section{Statistical procedures, validation, and artifact}
\label{app:statistics}

\subsection{Estimands and inference families}
\label{app:estimands}

The construction analysis uses the 864-estimate studentized program
bootstrap (50,000 draws, seed 20260731). The common truth-switch
comparison uses one twelve-estimate family, and the common natural-video
analysis uses a 12-estimate interface family and an 18-estimate model-gap
family. The two six-estimate truth-switch families and the 16-estimate
evidence and 18-estimate no-video-subtracted families of the submission
remain separate. The native-thinking GLM sensitivity uses its own
three-estimate family.

For the common-protocol families we use 10,000 shared stratified bootstrap
draws (seed 20260921), resampling whole program pairs or videos. The
simultaneous radius is the 95th percentile of the maximum centered
absolute deviation across estimates. Pointwise bounds are the 2.5th and
97.5th percentiles. Validation subsets use separate 12-estimate families.
Cross-panel draws refit the source selectors and resample the destination
separately. Reranking uses paired whole-video draws
(\cref{app:revision-validation,app:revision-selection,app:revision-rerank}).
Submitted seeds are 20260915 and 20260919 for truth-switch, 20260911 for
natural video, and 20260916 and 20260917 for selection.
No multiplicity correction is applied across families. In the main text, ``significant''
without qualification refers to a simultaneous interval within its family. Pointwise intervals
are unadjusted percentile intervals and are labeled as such where they appear, for example in
\cref{tab:truth-switch-tasks,tab:truth-switch-margin,tab:revision-score,tab:natural-cells,tab:revision-evidence,tab:revision-dose,tab:revision-staged-ci,tab:revision-staged-differences,tab:transfer-interaction}
and in the request and saved-rationale contrasts of \cref{app:revision-natural}.

\subsection{Validation}
\label{app:validation}

All summary statistics were reproduced from the per-response outputs by
independent code, and every study checks the completeness of its
model/unit/condition/order keys, parser outcomes, and candidate-order
aggregation. Infrastructure failures were retried with identical inputs and are excluded
from all counts.

\subsection{Artifact}
\label{app:availability}

The artifact includes controlled-task generators and renderer, prompts
and parsers, validated-position readouts, common-protocol responses, both
panels' propositions, annotation labels, the rationale codes, and analysis
scripts. Full truth-switch, selection, and reranking packages will be
released with the paper.

\FloatBarrier

\section{Extended related work}
\label{app:related}

\paragraph{Measurement validity and negatives.} Shortcut learning
\citep{geirhos2020shortcut} and construct validity
\citep{cronbach1955construct,jacobs2021measurement} explain benchmark success
that fails under distributions closer to the intended construct. Hard
negatives expose order and relation insensitivity
\citep{thrush2022winoground,yuksekgonul2023aro}, generated negatives carry
exploitable textual artifacts \citep{hsieh2023sugarcrepe}, and leaderboard
rankings move with ordering, prompting, and scoring
\citep{alzahrani2024benchmarks}. Contrast sets probe local decision boundaries
with small label-changing perturbations \citep{gardner2020contrast}; our
endpoint-matched negatives apply the same logic to executable state
trajectories while preserving both endpoints, and our witness videos apply it to
the video rather than the candidate.

\paragraph{Contrastive video-language evaluation.} VELOCITI
\citep{saravanan2025velociti} separates absolute decisions from relative
ranking and tests answer-position bias with absent-video and single-frame
controls; its appendix reports reduced performance under chain-of-thought
prompting for its two tested models. Vinoground \citep{zhang2024vinoground}
pairs short natural videos with temporally counterfactual captions and
reports a text score that requires the correct caption for both videos of a
pair; our paired success follows this convention with executable references,
shared endpoint pixels, and fixed text inputs within a candidate order.

\paragraph{State tracking and temporal diagnostics.}
Temporal-perception
benchmarks, atemporal probes, and evidence-perturbation audits show how
appearance or unordered content can sustain video-language performance
\citep{liu2024tempcompass,patraucean2023perception,buch2022atp,lei2023singleframe,sevillalara2021onlytime,lee2026accuracy,rahman2026temporal,chen2026tocbench,meng2026videozero}.
VET-Bench tracks visually identical objects through shuffles from motion
alone, finds near-chance frozen models, and itself separates that setting from
cup-game videos whose overlaid arrows disclose each swap \citep{liu2026shell};
VSTAT finds that models solve textual transcriptions of state-tracking videos
but fail on the videos \citep{yu2026vstat}. Both benchmarks evaluate solvers.
Our controlled frames disclose each event, and we evaluate verifiers of supplied trajectories.

\paragraph{Judgment interfaces, judge reliability, and reward use.}
First-token probabilities can disagree with generated answers
\citep{wang2024answerc}; judge studies and benchmarks document position,
length, informativeness, and perceptual biases
\citep{zheng2023judging,chen2024judgement,liu2023geval,zhao2025onetoken,lee2026mmjudgebias,zou2026withoutseeing,park2026perceptual,chen2024mllmjudge,li2024vlrewardbench,he2024videoscore,pi2025mrjudge}.
Reranking, process supervision, and reward overoptimization
\citep{cobbe2021verifiers,lightman2023verify,gao2023scaling} motivate
measuring what a verifier rewards.

\paragraph{Structured and evidence-grounded verification.} Atomic and
structured systems decompose judgments into facts, scenes, claims, clues, or
visual premises \citep{min2023factscore,wang2026videohv,min2026perceval};
NeuS-QA grounds question-derived propositions and applies temporal-logic
model checking before answering \citep{shah2026neusqa}; RT-NeuS reduces
acquisition cost through adaptive sampling and cache reuse
\citep{liang2026rtneus}; EVPV checks candidate reasoning premises against
visual constraints extracted once from the image and question
\citep{wang2026grounding}; SG-PVR trains video reward reasoning from atomic
claims and scene graphs \citep{kim2026sgpvr}; DEFT defers trajectory options
until after a scene-derived decision \citep{huang2026deft}. These systems
change the verification interface; our study measures how the interface and
the negative type change what an evaluation establishes about a fixed
verifier.

\clearpage
\begin{table}[ht]
\centering\scriptsize\revcolor
\caption{\rev{\textbf{Closest prior work and what the design adds.}}}
\label{tab:revision-related}
\begin{tabularx}{\linewidth}{@{}p{.24\linewidth}YY@{}}
\toprule
Prior work & Control or outcome & This study \\
\midrule
Contrast sets \citep{gardner2020contrast} & Local label-changing edits & Executable trajectories with the same endpoints and fixed text. \\
Winoground, ARO, SugarCrepe \citep{thrush2022winoground,yuksekgonul2023aro,hsieh2023sugarcrepe} & Reordered captions or controlled hard negatives & Legal trajectories that differ internally with the same endpoints. \\
VideoCon and VELOCITI \citep{bansal2024videocon,saravanan2025velociti} & Perturbed or in-video negative captions and evidence controls & Crossing judgment interface with negative type and probing stages. \\
Vinoground \citep{zhang2024vinoground} & Paired temporal counterfactuals on natural clips & Executable truth conditions, matched endpoints, and fixed text; a second natural-video panel. \\
Judge reliability \citep{zheng2023judging,chen2024judgement,wang2024answerc,lee2026mmjudgebias,zou2026withoutseeing,park2026perceptual} & Position, length, informativeness, and perceptual biases; readout versus text & Negative-type selection and reranking costs. \\
State tracking \citep{liu2026shell,yu2026vstat} & Solvers track state through video & Verifiers check supplied trajectories with disclosed events. \\
\bottomrule
\end{tabularx}
\end{table}

\FloatBarrier
\section{\texorpdfstring{\rev{Revision controls and reconciliation}}{Revision controls and reconciliation}}
\label{app:revision}
\begingroup
\revcolor

This appendix reports the controls added in revision. \cref{app:revision-validation} documents
the independent validation of the natural panel (\cref{sec:natural-design}).
\cref{app:revision-position} documents the validated answer position (\cref{sec:outcomes}).
\cref{app:revision-interface,app:revision-controlled,app:revision-render} vary the interface,
the input stage, and the printed labels on the truth-switch test
(\cref{sec:truth-switch-controls}). \cref{app:revision-denial,app:revision-natural} give the
event-denial control on the controlled task and the natural-video controls
(\cref{sec:natural-elicitation}). \cref{app:revision-transfer} reports the second panel and the
two added models (\cref{sec:natural-elicitation}).
\cref{app:revision-selection,app:revision-rerank} report the robustness of the selection cost
and the reranking study (\cref{sec:selection}). \cref{app:revision-reconciliation} reconciles
the numbers of the original submission with the revised ones, and \cref{tab:protocol-labels}
there defines every protocol label.

\FloatBarrier

\subsection{Independent validation}
\label{app:revision-validation}

\noindent\textbf{Protocol.} Two annotators who are not authors labeled all 400 videos
independently, with no AI pre-labeling. For each video they saw the three propositions
(faithful, content-matched, and event-denial) in shuffled order under neutral identifiers, so
they did not know which one was the reference, and they saw no model outputs, no scores, and
not each other's labels. They first judged each proposition against the full source video as
true, false, or cannot determine, recorded the time span that contradicts every negative they
judged false, and flagged the propositions they found ambiguous. After their truth values were locked, they were
shown the exact decoded frame grids the models received (48 frames for GLM and InternVL, which
share one grid, 64 for Qwen, and 80 for Molmo) and judged for each proposition whether its
evidence is visible, not visible, or unclear in those frames.

\begin{table}[ht]
\centering\small\revcolor
\caption{\rev{\textbf{Truth labels on the 400-video panel. Counts are true, false, and cannot determine.}}}

\begin{tabular}{@{}lcccc@{}}
\toprule
Proposition & Annotator 1 & Annotator 2 & Agreement & $\kappa$ \\
\midrule
Faithful & 390 / 3 / 7 & 390 / 4 / 6 & 97.5\% & 0.49 \\
Content-matched & 2 / 392 / 6 & 3 / 392 / 5 & 99.0\% & 0.75 \\
Event-denial & 3 / 386 / 11 & 2 / 389 / 9 & 98.2\% & 0.71 \\
All 1,200 &  &  & 98.2\% & 0.96 \\
\bottomrule
\end{tabular}
\end{table}

Per-type $\kappa$ is lower because almost every item of a type receives the
same label. The faithful-true/content-false counts are 386 and 388 (383
both), and the faithful-true/denial-false counts are 384 and 387 (383 both).
Both annotators judged a denial true on one video, in which a black sugar
jar did not return to view. In another clip, one found no fully hidden
spoon and the other could not determine visibility around the event.

A span was recorded for every negative judged false (778 and 781).
Of 775 jointly false negatives, 734 have spans overlapping within one
second (94.7\%). Ambiguity is flagged on 4.0\% and 3.8\% of videos, with
98.2\% agreement and $\kappa=0.77$.
\begin{table}[ht]
\centering\footnotesize\revcolor
\caption{\rev{\textbf{Visibility on the sampled frame grids. Counts are out of 1,200 propositions.}}}

\begin{tabular}{@{}lcccc@{}}
\toprule
Frames & Visible A1 / A2 & Agreement & $\kappa$ & All three visible (videos) \\
\midrule
48 (GLM, InternVL) & 1,083 / 1,086 & 99.5\% & 0.97 & 361 \\
64 (Qwen) & 1,092 / 1,092 & 99.8\% & 0.98 & 364 \\
80 (Molmo) & 1,104 / 1,107 & 99.8\% & 0.98 & 368 \\
\bottomrule
\end{tabular}
\end{table}

\begin{table}[ht]
\centering\footnotesize\revcolor
\caption{\rev{\textbf{Visibility by question type.} Numbers of visible propositions, Annotator 1 / 2.}}
\label{tab:revision-visibility-by-type}
\begin{tabular}{@{}lrrrr@{}}
\toprule
Question type & Propositions & 48 frames & 64 frames & 80 frames \\
\midrule
Event order & 360 & 357 / 357 & 360 / 360 & 360 / 360 \\
Event-relative state & 300 & 231 / 234 & 237 / 237 & 243 / 246 \\
Temporal location & 300 & 261 / 261 & 261 / 261 & 267 / 267 \\
Visibility continuity & 240 & 234 / 234 & 234 / 234 & 234 / 234 \\
\bottomrule
\end{tabular}
\end{table}

Event-relative-state visibility is lower (77--82\%), including many
reappearance cases. On the 83 reappearance clips, both annotators mark
all 47 A-videos visible and all 31 B/C-videos not visible at 64 frames.
Of the five D-videos, one is marked not visible by one annotator and the
remaining judgments are unclear.

The subset analysis reruns the interface contrasts on three subsets of videos: (i) both
annotators confirm the relevant label pair, that is, the faithful proposition true and the
negative false; (ii) both confirm it and both see the evidence on that model's frame grid; and
(iii) at least one annotator confirms it. Interaction estimates use the videos that are in the
subset for both negative types. Each subset has its own 12-estimate family. On subset (ii),
event-denial accuracy changes by $-21.8$ points [$-29.4$, $-14.2$] for Qwen and $-20.3$
[$-27.9$, $-12.7$] for InternVL.
\begin{table}[ht]
\centering\scriptsize\revcolor\setlength{\tabcolsep}{4pt}
\caption{\rev{\textbf{Validation subset (i). CoT-minus-direct change in points (simultaneous 95\% intervals).}}}

\begin{tabular}{@{}lcccccc@{}}
\toprule
Model & Denial $N$ & Denial & Content $N$ & Content & Joint $N$ & Interaction \\
\midrule
Qwen3-VL-32B & 383 & $-23.4 [-30.6, -16.2]$ & 383 & $-1.7 [-8.9, +5.5]$ & 382 & $-21.9 [-29.1, -14.7]$ \\
GLM-4.5V & 383 & $-6.0 [-13.2, +1.2]$ & 383 & $+5.4 [-1.8, +12.6]$ & 382 & $-11.5 [-18.7, -4.3]$ \\
InternVL3.5-38B & 383 & $-22.8 [-30.0, -15.6]$ & 383 & $+5.2 [-2.0, +12.4]$ & 382 & $-28.3 [-35.5, -21.1]$ \\
Molmo2-8B & 383 & $-5.0 [-12.2, +2.2]$ & 383 & $-6.5 [-13.7, +0.7]$ & 382 & $+1.6 [-5.6, +8.8]$ \\
\bottomrule
\end{tabular}
\end{table}
\begin{table}[ht]
\centering\scriptsize\revcolor\setlength{\tabcolsep}{4pt}
\caption{\rev{\textbf{Validation subset (ii). CoT-minus-direct change in points (simultaneous 95\% intervals).}}}

\begin{tabular}{@{}lcccccc@{}}
\toprule
Model & Denial $N$ & Denial & Content $N$ & Content & Joint $N$ & Interaction \\
\midrule
Qwen3-VL-32B & 347 & $-21.8 [-29.4, -14.2]$ & 347 & $-1.7 [-9.3, +5.9]$ & 346 & $-20.2 [-27.8, -12.6]$ \\
GLM-4.5V & 344 & $-5.2 [-12.8, +2.4]$ & 344 & $+5.4 [-2.2, +13.0]$ & 343 & $-10.8 [-18.4, -3.2]$ \\
InternVL3.5-38B & 344 & $-20.3 [-27.9, -12.7]$ & 344 & $+6.1 [-1.5, +13.7]$ & 343 & $-26.7 [-34.3, -19.1]$ \\
Molmo2-8B & 351 & $-6.4 [-14.0, +1.2]$ & 351 & $-6.8 [-14.4, +0.8]$ & 350 & $+0.4 [-7.2, +8.0]$ \\
\bottomrule
\end{tabular}
\end{table}
\begin{table}[ht]
\centering\scriptsize\revcolor\setlength{\tabcolsep}{4pt}
\caption{\rev{\textbf{Validation subset (iii). CoT-minus-direct change in points (simultaneous 95\% intervals).}}}

\begin{tabular}{@{}lcccccc@{}}
\toprule
Model & Denial $N$ & Denial & Content $N$ & Content & Joint $N$ & Interaction \\
\midrule
Qwen3-VL-32B & 388 & $-23.5 [-30.5, -16.4]$ & 391 & $-1.8 [-8.9, +5.3]$ & 388 & $-21.8 [-28.9, -14.7]$ \\
GLM-4.5V & 388 & $-6.3 [-13.4, +0.8]$ & 391 & $+4.7 [-2.4, +11.8]$ & 388 & $-11.2 [-18.3, -4.1]$ \\
InternVL3.5-38B & 388 & $-22.9 [-30.0, -15.9]$ & 391 & $+5.2 [-1.8, +12.3]$ & 388 & $-28.2 [-35.3, -21.1]$ \\
Molmo2-8B & 388 & $-4.8 [-11.9, +2.3]$ & 391 & $-6.3 [-13.4, +0.8]$ & 388 & $+1.7 [-5.4, +8.8]$ \\
\bottomrule
\end{tabular}
\end{table}
\begin{table}[ht]
\centering\small\revcolor
\caption{\rev{\textbf{Selection cost on validated subsets.} Eight per-model configurations from \cref{tab:selection}. Simultaneous 95\% intervals. The
second subset keeps the videos whose evidence both annotators see on all three frame grids.}}
\label{tab:validation-selection}

\begin{tabular}{@{}lcc@{}}
\toprule
Subset & Videos & Held-out cost (pp) \\
\midrule
Both confirm & 382 & $9.21 [2.80,15.63]$ \\
Both confirm and see the evidence & 343 & $10.70 [3.82,17.58]$ \\
At least one confirms & 388 & $9.36 [2.89,15.83]$ \\
\bottomrule
\end{tabular}
\end{table}
\FloatBarrier
\subsection{Answer position and original settings}
\label{app:revision-position}
Direct judgment compares the probabilities of the labels A and B at one token position. The
submission used the first token after the prompt, which we call the original position. Not
every model writes its answer letter there. At the original position GLM's two labels carry
under 1\% of the next-token probability in every cell
(\cref{tab:position-truth,tab:position-ordered,tab:position-shuffled,tab:position-none}), and
InternVL's carry 5--33\% on the truth-switch inputs (\cref{tab:position-truth}), so a
comparison of A with B at that position need not reflect the answer the model would give.

\noindent\textbf{Procedure.} For each model we generated concise answers on a calibration
sample with the correct answers withheld and took the position at which the model writes its
answer letter. For GLM this position is inside its native answer box. The rule is the same for
every model and uses neither label mass nor accuracy. We then re-scored the saved inputs at
this validated position. Label mass is the probability of A and B, summed over their bare and
leading-space variants, in the full next-token distribution. The original readout compares the bare label
tokens. The validated readout sums the distinct bare and leading-space variants for each label. The validated readout appends
\texttt{Answer:} for Qwen and Molmo, and an empty thinking block followed by
\texttt{Answer:} for InternVL. The task wording, candidate order, and frames are the same
at both positions.

\noindent\textbf{Label mass at the validated position.} With video it is at least 87.72\% for
every model, and at least 88.96\% on the text inputs of the truth-switch test
(\cref{tab:position-truth,tab:position-ordered,tab:position-shuffled}). Without video it is
99.93--99.97\% for Qwen, 88.68--89.94\% for GLM and Molmo, and 71.38\% and 58.88\% for InternVL
against event-denial and content-matched negatives (\cref{tab:position-none}). InternVL's
no-video direct accuracy should be read with that in mind. At the original position, where
InternVL's labels carry 100.00\% of the probability with and without video, its event-denial
accuracy is 81.62\% with ordered video and 66.75\% without
(\cref{tab:position-ordered,tab:position-none}), the same pattern as at the validated position
(86.50\% and 74.62\%).

\noindent\textbf{Effect on accuracy.} For Qwen, InternVL, and Molmo the original position
already carried at least 99.99\% of the mass on ordered natural video. Reading them at the
validated position changes their ordered-video accuracy by 0.63 to 7.00 points, the largest
change being Molmo against event-denial negatives (30.50\% to 37.50\%;
\cref{tab:position-ordered}).

\noindent\textbf{Reading
\cref{tab:position-truth,tab:position-ordered,tab:position-shuffled,tab:position-none}.}
``Original'' and ``Validated'' give the two positions in the same runtime, so ``Original
direct'' can differ slightly from the submitted accuracies of
\cref{app:truth-switch,app:natural}, which came from an earlier runtime. ``CoT'' is accuracy
under the common protocol, and a dash means that no CoT cell exists for that evidence
condition. All 400 natural videos and both candidate orders are retained in every cell, and the
controlled comparisons retain both orders and both witness videos.

\noindent\textbf{Construction study.} The 17,874 GLM forward passes cover all 324 programs.
Mean label mass at GLM's validated position is above 99.996\% in each of its 11 evidence cells,
and the full 864-estimate family is recomputed with the re-read GLM scores
(\cref{tab:m2-absolute,tab:construction-reread}). In the construction study label mass was
measured for Qwen, GLM, InternVL, and Molmo, and not for Kimi-VL and Aria.

\RevisionPositionTables

\cref{tab:original-qwen,tab:original-glm,tab:original-internvl,tab:original-molmo} return to
the 48 cells of the submission (model, interface, evidence condition, and negative type) and
measure the label mass at the original position on the original prompts. For a CoT cell that
position is the first generated token, before any rationale is written, so near-zero mass there
is expected and says nothing about the accuracy of the generated answer; the direct rows are
the informative ones. Margin variance is the population variance of the A-minus-B
log-probability margin across the 800 responses in a cell, using the summed bare and
leading-space variants for each label. ``A/B top 1'' is the share of responses whose
most probable token in the full vocabulary is one of the declared label tokens. Every row
uses 400 videos and 800 responses.

\RevisionSubmittedPositionTables
\FloatBarrier

\begin{table}[ht]
\centering\scriptsize\revcolor
\caption{\rev{\textbf{GLM construction re-read. Original / validated win rate at pooled lengths.}}}
\label{tab:construction-reread}

\begin{tabular}{@{}lccccc@{}}
\toprule
Task & Evidence & Isolated & Visible & Several & One \\
\midrule
Shell & video & 0.984 / 0.986 & 0.894 / 0.958 & 0.542 / 0.574 & 0.495 / 0.500 \\
Shell & no video & 0.537 / 0.551 & 0.537 / 0.519 & 0.532 / 0.528 & 0.454 / 0.454 \\
Shell & oracle text & 0.995 / 0.995 & 1.000 / 1.000 & 0.991 / 0.981 & 0.972 / 0.972 \\
Shell & events text & 0.894 / 0.792 & 0.792 / 0.657 & 0.532 / 0.528 & 0.523 / 0.491 \\
Toggle & video & 0.856 / 0.867 & 0.778 / 0.769 & 0.532 / 0.611 & 0.505 / 0.509 \\
Toggle & no video & 0.563 / 0.485 & 0.525 / 0.426 & 0.551 / 0.426 & 0.495 / 0.426 \\
Toggle & oracle text & 1.000 / 1.000 & 1.000 / 1.000 & 1.000 / 1.000 & 1.000 / 1.000 \\
Toggle & events text & 0.991 / 0.988 & 0.975 / 0.988 & 0.583 / 0.657 & 0.514 / 0.546 \\
Counter & video & 0.850 / 0.898 & 0.752 / 0.792 & 0.565 / 0.630 & 0.523 / 0.528 \\
Counter & no video & 0.644 / 0.648 & 0.451 / 0.463 & 0.440 / 0.435 & 0.463 / 0.370 \\
Counter & oracle text & 1.000 / 1.000 & 1.000 / 1.000 & 1.000 / 1.000 & 1.000 / 1.000 \\
\bottomrule
\end{tabular}
\end{table}

With these scores GLM qualifies on shell. Its toggle drop is 15.74 points with a simultaneous
lower bound of $-2.88$, and its counter drop is 16.20 points with a lower bound of $-3.24$, so
neither meets the drop criterion of \cref{app:scores}.
\FloatBarrier

\subsection{Interface controls}
\label{app:revision-interface}

\noindent\textbf{Answer extraction.} In the earlier extraction audit an LLM assistant checked
114 flagged outputs with the correct answers withheld, including 54 from controlled runs.
Of those 54, 17 were retained in the final generation analyses, and 11 of the 17 expressed
the correct choice.

Four prompts cross two factors. G0 and G1 use generic instructions and request a concise answer
(G0) or a brief rationale (G1). P0 and P1 add a verification procedure to G0 and G1. CoT
judgment is P1 on the controlled tasks (\cref{sec:truth-switch}) and G1 on natural video
(\cref{sec:natural}), where the generated-answer control is G0. The instructions are:

\begin{quote}\small
\textbf{G0 and P0.} Give a concise answer without an explanation. End with a
separate final line exactly in the form Answer: A or Answer: B.

\textbf{G1 and P1.} Give a brief rationale based on the supplied evidence
before deciding. End with a separate final line exactly in the form
Answer: A or Answer: B.

\textbf{Controlled procedure (P0, P1).} Verification procedure: identify
the initial state, identify each primitive event in chronological order,
apply each event to update the state, and compare the resulting complete
state sequence against both candidates.

\textbf{Natural procedure (P0, P1).} Verification procedure: identify the
entities and events referred to in the question and candidates, locate
evidence for the disputed event or relation, determine whether each
referred event actually occurs, and compare both statements against that
evidence. Check event occurrence separately from the state, visibility,
location, or ordering asserted about it.
\end{quote}

\begin{table}[ht]
\centering\scriptsize\revcolor
\caption{\rev{\textbf{Truth-switch readout and prompt controls.} Paired success (\%); mean generated tokens in parentheses. ``Generated answer, 8-token limit''
generates from the validated answer position with P0 under an eight-token limit. ``Submitted
one-token instruction, free completion'' lets the model complete freely after the direct prompt
of the submission. The last row is the native-thinking probe (P1, 16 pairs, GLM and InternVL).}}
\label{tab:truth-prompt-controls}

\begin{tabular}{@{}lcccc@{}}
\toprule
Readout & Qwen & GLM & InternVL & Molmo \\
\midrule
Validated direct & 9.17 & 9.17 & 5.00 & 1.67 \\
Generated answer, 8-token limit & 15.83 & 7.50 & 5.00 & 1.67 \\
Concise, generic (G0) & 17.50 (4) & 10.00 (6) & 97.50 (3,269) & 3.33 (107) \\
Concise, procedure (P0) & 15.83 (4) & 8.33 (6) & 95.00 (3,079) & 3.33 (169) \\
Rationale, generic (G1) & 100.00 (643) & 86.67 (408) & 62.50 (459) & 22.50 (279) \\
Rationale, procedure (P1) & 100.00 (620) & 90.00 (454) & 70.00 (418) & 25.00 (262) \\
Submitted one-token instruction, free completion & 5.83 (2) & 88.33 (453) & 94.17 (4,049) & 3.33 (2) \\
Native thinking, 16-pair probe &  & 100.00 & 100.00 &  \\
\bottomrule
\end{tabular}
\end{table}

The concise and rationale prompts share the same scoring convention.
InternVL writes long outputs even under the concise prompts (3,269 and 3,079 generated tokens
on average under G0 and P0) and reaches 97.50\% and 95.00\%, so its concise rows contain
written reasoning.
Adding the verification procedure to G1 changes paired success by at most
7.5 points in the original render. At the 8,192-token request, at most one
of 240 G1 or P1 original-render rationales per model hits its cap. This
cap result concerns the original render; harder arrow-only renders can
have more cap hits. With native thinking on, GLM and InternVL reach 100\%
in the 16-pair P1 probe. With Qwen on matched 16-unit subsets, raising
the requested cap from 8,192 to 16,384 leaves P0 and P1 estimates unchanged
for truth-switch (15.6/96.9\%), natural denial (40.6/37.5\%), and natural
content-matched (56.3/43.8\%), at the same frame budgets.

For the rationale-length control, we remove the final answer line from each saved rationale
without reference to the correct answer, keep a prefix of the rationale, and read the choice at
the validated answer position. Requested and realized prefix lengths differ when a saved rationale
is shorter. In the controlled P1 comparison, Qwen's paired success at
0/64/256/1,024 saved tokens is 18.3/19.2/77.5/99.2\%; GLM's is
9.2/10.0/84.2/89.2\%. These are within-prompt readouts of saved text.
At 0 tokens the answer is read under the rationale prompt, so it differs
from the direct column of \cref{tab:revision-truth}.
They complement the between-prompt comparison in \cref{tab:revision-truth}.

\begin{figure}[ht]
\centering
\includegraphics[width=0.92\linewidth]{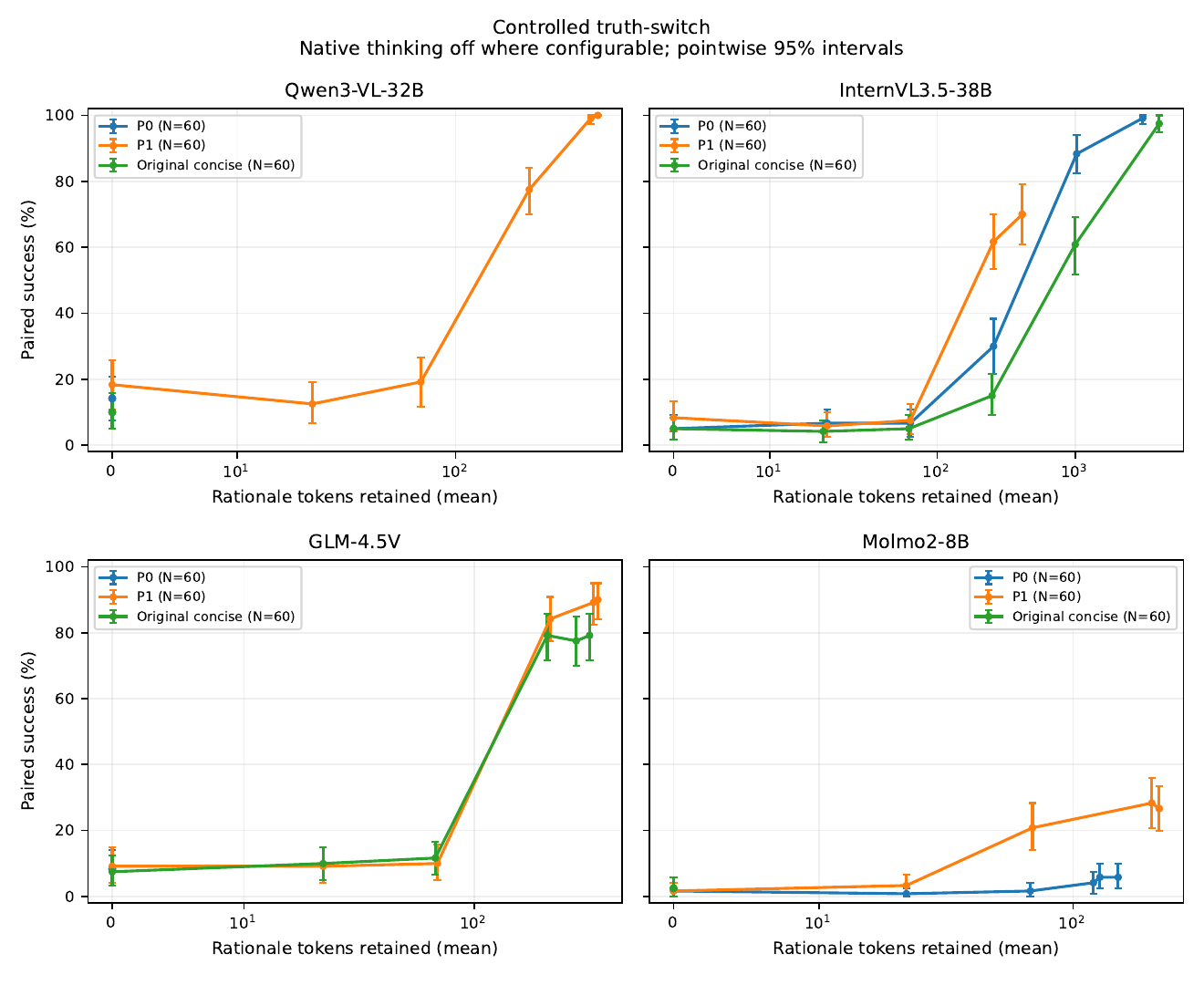}
\caption{\rev{\textbf{Controlled saved-token dose response.} Readout uses the same prompt and a varying prefix of its saved rationale.}}
\label{fig:truth-dose}
\end{figure}
\FloatBarrier

\begin{table}[ht]
\centering\scriptsize\revcolor
\caption{\rev{\textbf{Rationale prefixes on the truth-switch test.} Paired success (\%) with pointwise 95\% intervals.}}
\label{tab:revision-dose}
\begin{tabular}{@{}lcccc@{}}
\toprule
Tokens & Qwen & GLM & InternVL & Molmo \\
\midrule
0 & $18.33\;[10.83,25.83]$ & $9.17\;[4.17,15.00]$ & $8.33\;[4.17,13.33]$ & $1.67\;[0.00,4.17]$ \\
16 & $12.50\;[6.67,19.17]$ & $9.17\;[4.17,15.00]$ & $5.83\;[2.50,10.00]$ & $3.33\;[0.83,6.67]$ \\
64 & $19.17\;[11.67,26.67]$ & $10.00\;[5.00,15.83]$ & $7.50\;[3.33,12.50]$ & $20.83\;[14.17,28.33]$ \\
256 & $77.50\;[70.00,84.17]$ & $84.17\;[77.50,90.83]$ & $61.67\;[53.33,70.00]$ & $28.33\;[20.83,35.83]$ \\
1,024 & $99.17\;[97.50,100.00]$ & $89.17\;[82.50,95.00]$ & $70.00\;[60.83,79.17]$ & $26.67\;[20.00,33.33]$ \\
Full & $100.00\;[100.00,100.00]$ & $90.00\;[84.17,95.00]$ & $70.00\;[60.83,79.17]$ & $26.67\;[20.00,33.33]$ \\
\bottomrule
\end{tabular}
\end{table}
\begin{table}[ht]
\centering\scriptsize\revcolor
\caption{\rev{\textbf{Controlled generation diagnostics.} ``Capped'' and ``No choice'' count responses out of 240. ``Expressed choice'' is paired success
(\%) when an expressed choice counts even at the cap, and ``Cap fails'' is paired success (\%)
when every capped response counts as wrong. A capped response exhausts its effective allowance
without an end-of-sequence token.}}
\label{tab:revision-synthetic-caps}
\begin{tabular}{@{}llrrrrrr@{}}
\toprule
Model & Setting & Mean tokens & Capped & No choice & Requested cap & Expressed choice & Cap fails \\
\midrule
Qwen3-VL-32B & G0 & 4.0 & 0 & 0 & 8,192 & 17.50 & 17.50 \\
Qwen3-VL-32B & G1 & 643.2 & 0 & 0 & 8,192 & 100.00 & 100.00 \\
Qwen3-VL-32B & P0 & 4.0 & 0 & 0 & 8,192 & 15.83 & 15.83 \\
Qwen3-VL-32B & P1 & 620.1 & 0 & 0 & 8,192 & 100.00 & 100.00 \\
GLM-4.5V & G0 & 6.0 & 0 & 0 & 8,192 & 10.00 & 10.00 \\
GLM-4.5V & G1 & 408.0 & 0 & 0 & 8,192 & 86.67 & 86.67 \\
GLM-4.5V & P0 & 6.0 & 0 & 0 & 8,192 & 8.33 & 8.33 \\
GLM-4.5V & P1 & 454.2 & 0 & 0 & 8,192 & 90.00 & 90.00 \\
InternVL3.5-38B & G0 & 3269.5 & 6 & 4 & 8,192 & 97.50 & 95.83 \\
InternVL3.5-38B & G1 & 459.3 & 0 & 0 & 8,192 & 62.50 & 62.50 \\
InternVL3.5-38B & P0 & 3079.1 & 6 & 6 & 8,192 & 95.00 & 95.00 \\
InternVL3.5-38B & P1 & 417.9 & 0 & 0 & 8,192 & 70.00 & 70.00 \\
Molmo2-8B & G0 & 107.3 & 1 & 1 & 8,192 & 3.33 & 3.33 \\
Molmo2-8B & G1 & 278.7 & 1 & 30 & 8,192 & 22.50 & 22.50 \\
Molmo2-8B & P0 & 168.9 & 1 & 23 & 8,192 & 3.33 & 3.33 \\
Molmo2-8B & P1 & 262.3 & 0 & 14 & 8,192 & 25.00 & 25.00 \\
\bottomrule
\end{tabular}
\end{table}

\FloatBarrier

\subsection{Staged inputs}
\label{app:revision-controlled}
The three representations are rendered frames, a verified initial state plus
ordered primitive events, and the verified state sequence.
Supplying the events as text removes the need to read the frames and to order the events;
supplying the states also removes state updating. Under direct judgment paired success stays at
1.67--12.50\% for frames and events and is 100.00\% for states (\cref{tab:revision-staged-ci}).
All use the same
60 programs, candidates, and counterbalancing. With P1, moving from frames to
events raises paired success by 0.00, 6.67, 20.83, and 12.50 points for Qwen,
GLM, InternVL, and Molmo. Moving from events to states changes it by $-3.33$, +3.33, +9.17, and +52.50 points. Exact event-list transcription
from the original frames is 100.0\%, 85.0\%, 66.7\%, and 39.2\%.
The stages change the representation along with the task demand. Object identity and slot positions
are fixed by design, so object correspondence is not a demand of these tasks. Spatial reading
is measured by the arrows-only render, where the slots must be read from the position of the
arrow, and the use of frame order by the no-counter render (\cref{app:revision-render}). In the
shell task the state is the ball's slot, so the state-sequence stage also gives the object's
verified location after every event.

\Cref{tab:revision-trajectory-demands} summarizes the demands, and
\cref{tab:revision-staged-ci} reports the staged paired-success estimates.

\begin{table}[ht]
\centering\small\revcolor
\caption{\rev{\textbf{How each demand of an internal-trajectory judgment is handled.}
Results are paired success for Qwen, GLM, InternVL, and Molmo, in that order. Guessing gives 25\%.}}
\label{tab:revision-trajectory-demands}
\begin{tabularx}{\linewidth}{@{}L{.19\linewidth}YL{.30\linewidth}@{}}
\toprule
Demand & How the design handles it & Result \\
\midrule
Matching text to video & Removed by supplying events as text. &
Direct stays at 10.00--12.50\%. \\
Spatial localization & Probed with arrows only. The arrow's position identifies the slots. &
CoT: 85.00 / 48.33 / 40.00 / 24.17\%. \\
Event ordering & Probed without the counter. The witness videos differ only in frame order. &
CoT: 97.50 / 63.33 / 50.00 / 28.33\%. \\
Object correspondence & Fixed by design. Slots and switches do not move or change identity. &
Fixed across conditions. \\
State updating & Removed by supplying the states as text. &
Direct: 100.00\% for all four models. \\
\bottomrule
\end{tabularx}
\end{table}

\begin{table}[ht]
\centering\small\revcolor
\caption{\rev{\textbf{Construction with events supplied as text.} Visible-start / several-deviation endpoint-matched win rate.}}

\begin{tabular}{@{}lcc@{}}
\toprule
Model & Shell & Toggle \\
\midrule
Qwen3-VL-32B & 0.93 / 0.53 & 1.00 / 0.53 \\
GLM-4.5V & 0.66 / 0.53 & 0.99 / 0.66 \\
InternVL3.5-38B & 0.94 / 0.58 & 1.00 / 0.66 \\
Molmo2-8B & 0.63 / 0.55 & 0.98 / 0.61 \\
\bottomrule
\end{tabular}
\end{table}

\begin{table}[ht]
\centering\scriptsize\revcolor
\caption{\rev{\textbf{Staged paired success.} Percentages and pointwise 95\% intervals; 60 pairs.}}
\label{tab:revision-staged-ci}
\begin{tabular}{@{}llcc@{}}
\toprule
Model & Evidence & Direct & CoT \\
\midrule
Qwen3-VL-32B & Frames & $9.17\;[5.00,14.17]$ & $100.00\;[100.00,100.00]$ \\
 & Events as text & $10.00\;[5.00,15.00]$ & $100.00\;[100.00,100.00]$ \\
 & States as text & $100.00\;[100.00,100.00]$ & $96.67\;[93.33,99.17]$ \\
GLM-4.5V & Frames & $9.17\;[4.17,14.17]$ & $90.00\;[84.17,95.00]$ \\
 & Events as text & $10.83\;[5.83,16.67]$ & $96.67\;[93.33,99.17]$ \\
 & States as text & $100.00\;[100.00,100.00]$ & $100.00\;[100.00,100.00]$ \\
InternVL3.5-38B & Frames & $5.00\;[1.67,9.17]$ & $70.00\;[60.83,79.17]$ \\
 & Events as text & $12.50\;[7.50,18.33]$ & $90.83\;[84.17,95.83]$ \\
 & States as text & $100.00\;[100.00,100.00]$ & $100.00\;[100.00,100.00]$ \\
Molmo2-8B & Frames & $1.67\;[0.00,4.17]$ & $25.00\;[18.33,32.50]$ \\
 & Events as text & $10.83\;[5.83,16.67]$ & $37.50\;[29.17,45.83]$ \\
 & States as text & $100.00\;[100.00,100.00]$ & $90.00\;[85.83,94.17]$ \\
\bottomrule
\end{tabular}
\end{table}
\begin{table}[ht]
\centering\scriptsize\revcolor
\caption{\rev{\textbf{Paired changes between stages.} Percentage points and pointwise 95\% intervals from the same program pairs.}}
\label{tab:revision-staged-differences}
\begin{tabular}{@{}lcccc@{}}
\toprule
Model & Direct: frames to events & Direct: events to states & CoT: frames to events & CoT: events to states \\
\midrule
Qwen3-VL-32B & $0.83\;[-5.83,7.50]$ & $90.00\;[85.00,95.00]$ & $0.00\;[0.00,0.00]$ & $-3.33\;[-6.67,-0.83]$ \\
GLM-4.5V & $1.67\;[-4.17,7.50]$ & $89.17\;[83.33,94.17]$ & $6.67\;[0.00,13.33]$ & $3.33\;[0.83,6.67]$ \\
InternVL3.5-38B & $7.50\;[0.00,15.00]$ & $87.50\;[81.67,92.50]$ & $20.83\;[11.67,30.00]$ & $9.17\;[4.17,15.83]$ \\
Molmo2-8B & $9.17\;[3.33,15.00]$ & $89.17\;[83.33,94.17]$ & $12.50\;[1.67,23.33]$ & $52.50\;[43.33,61.67]$ \\
\bottomrule
\end{tabular}
\end{table}

\FloatBarrier

\subsection{Label removal}
\label{app:revision-render}
\begin{table}[ht]
\centering\small\revcolor
\caption{\rev{\textbf{Printed label and counter removal.} Direct / CoT paired success (\%), 60 programs per cell.}}
\begin{tabular}{@{}lrrrr@{}}
\toprule
Render & Qwen & GLM & InternVL & Molmo \\
\midrule
Original & 9.17 / 100.00 & 9.17 / 90.00 & 5.00 / 70.00 & 1.67 / 25.00 \\
No event counter & 8.33 / 97.50 & 8.33 / 63.33 & 1.67 / 50.00 & 3.33 / 28.33 \\
No text label & 2.50 / 76.67 & 5.83 / 73.33 & 0.83 / 45.00 & 0.83 / 21.67 \\
Arrows only & 3.33 / 85.00 & 4.17 / 48.33 & 2.50 / 40.00 & 2.50 / 24.17 \\
\bottomrule
\end{tabular}
\end{table}

For the no-counter pair, the frame multiset is identical and ordering differs.
For arrows-only transcription, 600 reference events per model yield
\ifdefined\ARXIVVERSION
443/\allowbreak353/\allowbreak197/\allowbreak316 correct for
Qwen/\allowbreak GLM/\allowbreak InternVL/\allowbreak Molmo
(73.83/\allowbreak58.83/\allowbreak32.83/\allowbreak52.67\%).
\else
443/353/197/316 correct for Qwen/GLM/InternVL/Molmo (73.83/58.83/32.83/52.67\%).
\fi
Equal-length lists are scored positionally, other lists by exact matches in
a minimum-edit alignment. Exact whole-list accuracy is 26.67/9.17/0/0\%.
Partial event reading can suffice to distinguish the two supplied candidates.

Counting capped responses as wrong gives Qwen 82.50\% with arrows only
and 71.67\% without text labels (15 and 14 of 240 capped).

\FloatBarrier

\subsection{Controlled event-denial negatives}
\label{app:revision-denial}
This control asks whether an event-denial negative lowers accuracy under CoT judgment when the
events are given. On the same 60 programs, the faithful statement is paired with a negative
that either asserts the event and changes a detail or denies the event, the two forms used on
natural video (\cref{sec:candidates}). For example, a switch video starts at 011 and shows toggles of switches 0, 1, 1, 0, 2, and 1. The faithful statement gives the state sequence $011 \to 111 \to 101 \to 111 \to 011 \to 010 \to 000$. The negative that also asserts the event gives $011 \to 001 \to 000 \to 010 \to 000 \to 100 \to 000$. The event-denial negative states, ``The event `toggle switch 0' does not occur in the displayed sequence.''

When the negative denies the event, CoT judgment raises accuracy for all four models, by
7.08 to 20.42 points. When both candidates assert the event, it raises accuracy by 9.58 to
45.42 points (\cref{tab:revision-controlled-denial}; \cref{fig:natural}a,b; the intervals
are pointwise). The decline on natural video is therefore not explained by the event-denial
negative alone. It appears where the event must also be found.

\begin{table}[ht]
\centering\scriptsize\revcolor
\caption{\rev{\textbf{Controlled event-denial control.} Response accuracy (\%), direct / CoT, with the change in points and its pointwise 95\% paired whole-program bootstrap interval. There are 60 programs per cell. The faithful statement is held fixed across negative types.}}
\label{tab:revision-controlled-denial}
\begin{tabular}{@{}lcccc@{}}
\toprule
Negative & Qwen & GLM & InternVL & Molmo \\
\midrule
Both assert the event & \shortstack{54.58 / 100.00\\$+45.42$ [41.67, 49.17]} & \shortstack{55.83 / 86.25\\$+30.42$ [24.99, 35.83]} & \shortstack{54.17 / 90.83\\$+36.67$ [32.08, 41.25]} & \shortstack{50.00 / 59.58\\$+9.58$ [2.92, 15.83]} \\
\addlinespace[2pt]
One denies the event & \shortstack{92.92 / 100.00\\$+7.08$ [2.92, 12.08]} & \shortstack{77.08 / 87.08\\$+10.00$ [2.08, 18.33]} & \shortstack{69.17 / 89.58\\$+20.42$ [12.50, 28.75]} & \shortstack{69.58 / 82.50\\$+12.92$ [3.75, 22.50]} \\
\bottomrule
\end{tabular}
\end{table}

\FloatBarrier

\subsection{Natural-video controls}
\label{app:revision-natural}
\cref{tab:interface-expressed} gives the common-protocol family: for each of the four models,
the CoT-minus-direct change against each negative type and the difference between the two
changes. All 12 estimates use the same 400 videos, with both candidate orders averaged within
video. This contrast changes the prompt and the interface together, so the contrasts below take
it apart.

\noindent\textbf{Request for a rationale (G0 to G1).} Both settings generate the answer, and
only the request for a rationale differs. With native thinking off, event-denial accuracy
changes by $-18.62$ points for Qwen (pointwise 95\% interval [$-23.12$, $-14.13$]), $-23.00$
for InternVL [$-26.88$, $-19.12$], +3.75 for GLM, and +5.00 for Molmo. For GLM and Molmo the
concise generated answer is less accurate than validated direct judgment (53.37\% against
63.25\%, and 27.38\% against 37.50\%; \cref{tab:generation-glm,tab:generation-molmo}), which is
why their change is positive here and negative in \cref{tab:interface-expressed}.

\noindent\textbf{Content of the rationale (saved-rationale readout).} Here the prompt is fixed
and the answer is read at the validated position after a prefix of the model's own saved
rationale (\cref{fig:natural-dose}). Under InternVL's P1 event-denial prompt, accuracy is
79.88\% before the rationale and 61.25\% after it, a change of $-18.62$ points (pointwise
interval [$-22.50$, $-14.75$]). Under Qwen's G1 event-denial prompt, accuracy is 48.00\% before
any rationale token and 43.12\% after the full rationale, against 61.75\% for the concise G0
answer, so for Qwen most of the shift is present as soon as a rationale is requested.

\noindent\textbf{Verification procedure (G1 to P1).} Adding the natural-video procedure to the
rationale prompt changes event-denial and content-matched accuracy by +0.75 and +0.88 points
for Qwen, $-1.75$ and $-1.25$ for GLM, and $-3.12$ and $-1.00$ for InternVL
(\cref{tab:generation-qwen,tab:generation-glm,tab:generation-internvl}). Molmo was not run with
P1 on the full panel.

\noindent\textbf{Native thinking.} GLM's G1 run with native thinking on and its validated
direct run cover the same 400 videos. Event-denial accuracy is 44.12\% against 63.25\%, a
change of $-19.13$ points, and content-matched accuracy is 64.50\% against 59.00\%, a change of
+5.50 points. Their simultaneous intervals within a separate three-estimate family are
[$-24.38$, $-13.88$] and [0.25, 10.75], and the interval of their difference is [$-29.88$,
$-19.38$] (10,000 paired bootstrap draws within the four question strata, seed 20260921).

\noindent\textbf{Review of unresolved choices.} In the later natural-video runs the extraction
rule flagged 40 outputs for LLM review, 29 from G1 with native thinking off and the rest from
diagnostic settings. The assistant read them with the correct answers withheld and found a
stated choice in 12, of which 7 were correct. Counting these 7 as correct instead of counting
all 40 as incorrect moves no cell by more than 0.50 points. The largest change is 4 of GLM's
800 G1 event-denial responses with native thinking off. An earlier extraction audit also
covered controlled and natural-video outputs (\cref{app:revision-interface}).

\noindent\textbf{Frame order.} Shuffling the frames changes direct accuracy at the validated
answer position by at most 5.62 points.

\cref{tab:interface-expressed,tab:interface-cap-fail} differ only in the scoring rule:
\cref{tab:interface-expressed} counts an expressed choice even when the output reached its cap,
and \cref{tab:interface-cap-fail} counts every capped output as wrong.
\cref{tab:generation-qwen,tab:generation-glm,tab:generation-internvl,tab:generation-molmo} list
the generation settings, with unselected and capped responses retained. ``Mode'' is native
thinking, with ``off'' also used for models without a configurable thinking mode. The effective
cap is the smaller of the requested 8,192 tokens and the space left in the context window
after the input and, for G0, G1, P0, and P1, a 128-token readout reserve. ``Tokens'' is the mean
length of the full saved generation. ``Cap'' counts responses that exhaust their allowance
without an end-of-sequence token. ``Unparsed'' counts responses with no extracted choice. The
two counts can overlap. ``Consist.'' is the share of videos for which both orders yield a
choice and select the same proposition. ``Original concise'' is the direct instruction of
the submission with free completion. Rows with 80 videos are probes on a subset.

\RevisionNaturalTables
\FloatBarrier

\begin{figure}[ht]
\centering
\includegraphics[width=0.92\linewidth]{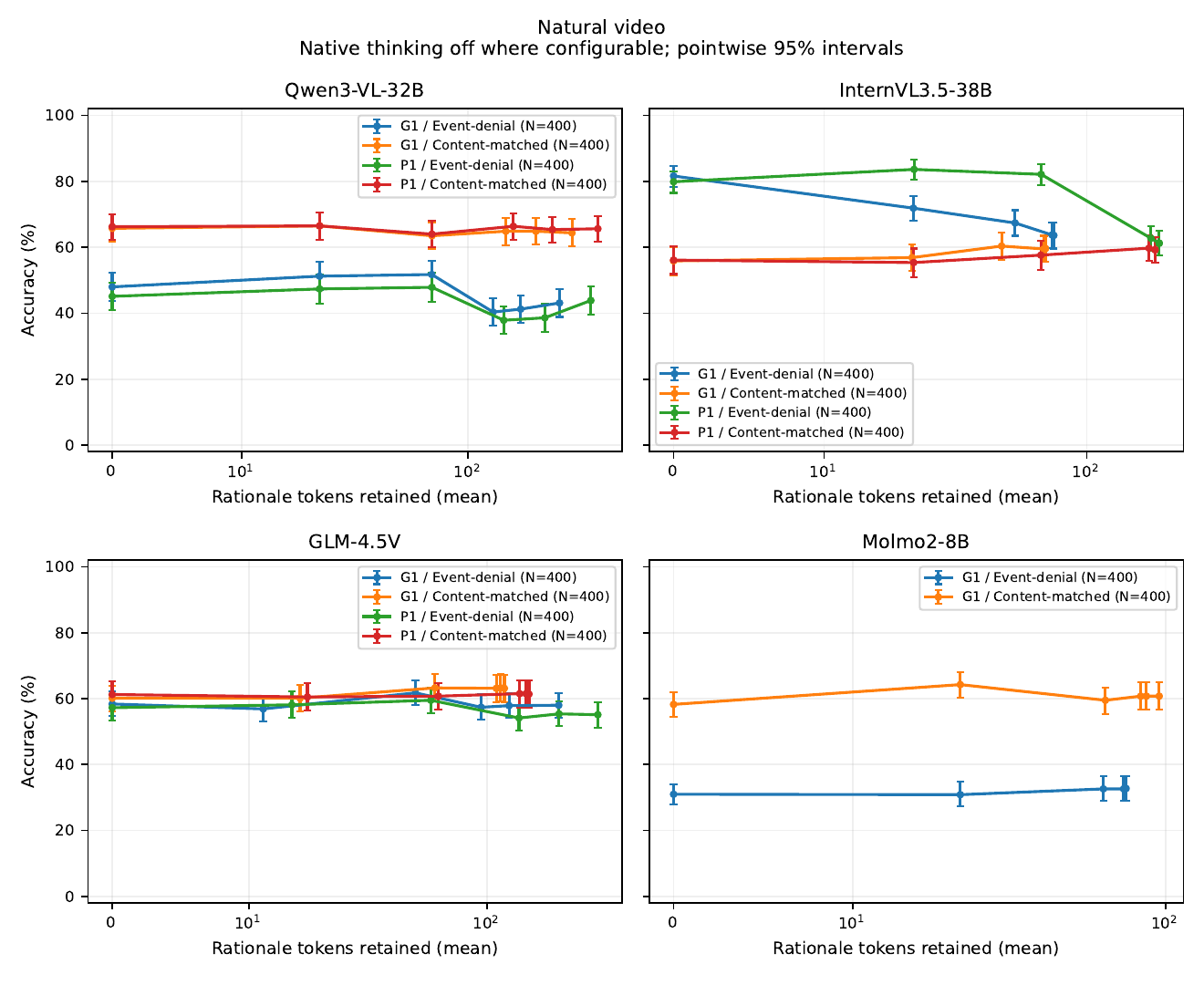}
\caption{\rev{\textbf{Natural-video saved-token dose response.} The curves retain the same complete video sets across cuts.}}
\label{fig:natural-dose}
\end{figure}

On the 47 clips with a visible return in Qwen's sampled frames, Qwen
affirms the event without candidates on 31.91\% of clips with a concise
answer and 55.32\% with a rationale. InternVL, using 48 frames from those
same clips, affirms it on 34.04\% and 59.57\%, respectively.
\FloatBarrier

\subsection{Second panel and added models}
\label{app:revision-transfer}

\noindent\textbf{Panel.} A fixed permutation (seed 20261108) selected complete Vinoground
caption pairs before any scoring, skipping reused clip intervals. The 400 pairs (800 clips)
contain 206 action, 128 object, and 66 viewpoint pairs. For each clip, the content-matched
negative is the caption of the other clip in its pair, and the event-denial negative denies the
captioned event. Two annotators who are not authors each checked the event-denial statements of 80 of the 800 clips. They judged 77 and 78 valid, with agreement on 79 ($\kappa$ = 0.79). Resampling groups clips that share a source.

\noindent\textbf{Settings.} The direct accuracies in \cref{tab:revision-transfer-accuracy}
use the validated answer position. For the four core models, the CoT runs use the
submission's model-specific prompts and native-thinking settings. We apply the same
answer-extraction rule to the saved TOC-Bench generations and the new Vinoground generations.
The extractor reads the final answer channel. An unambiguous choice counts even at the token
cap. A separate Qwen3-32B classifier checks flagged outputs for a stated choice, with the
correct answer withheld. Responses with no unique choice count as incorrect. This review
covered 298 outputs across both panels and found a stated choice in 217, of which 158 were
correct.
The TOC-Bench CoT column therefore differs from
\cref{tab:natural-cells} in its scoring rule and from \cref{tab:revision-natural} in its
generation settings. Kimi-VL-A3B and Aria use their own direct and CoT settings with the same
answer-extraction rule.

\noindent\textbf{Results.} On Vinoground, CoT judgment lowers event-denial accuracy for all
four core models, by 3.81 to 10.31 points, and changes content-matched accuracy by at most 2.31
points (\cref{tab:revision-transfer-accuracy}). \cref{tab:transfer-interaction} gives the
interaction, that is, the CoT-minus-direct change against event-denial negatives minus the
change against content-matched negatives, with pointwise intervals. The interaction is negative
with an interval that excludes zero for Qwen, GLM, InternVL, and Kimi-VL on TOC-Bench and for
Qwen, GLM, and InternVL on Vinoground. Under these settings Molmo's interaction is positive on
TOC-Bench (+6.75 [1.25, 12.38]) and not distinguishable from zero on Vinoground. Aria's
interaction interval includes zero on both panels.

The CoT-minus-direct changes for the two added models are given in points, with pointwise
95\% paired bootstrap intervals. In event-denial and content-matched order, on TOC-Bench
Aria changes by $+7.12\;[3.88, 10.38]$ and $+6.88\;[2.25, 11.50]$, and Kimi-VL by
$-4.50\;[-6.63, -2.25]$ and $-0.12\;[-2.50, 2.25]$. On Vinoground, Aria changes by
$-0.44\;[-3.03, 2.17]$ and $-2.31\;[-4.34, -0.32]$, and Kimi-VL by
$+1.62\;[0.30, 2.99]$ and $-0.12\;[-1.38, 1.14]$. The intervals
use 10,000 paired draws of whole videos within TOC-Bench question strata or source groups
on Vinoground, retaining both candidate orders.

\noindent\textbf{Selection sets.} \cref{tab:selection-cross-panel-sets} uses three eligible
sets. Both eight-configuration sets use the four core models and the same CoT scores under
the extraction rule above. The first uses bare-token direct scores at the original position
on each panel. The second replaces them with direct scores at the validated position.
The twelve-configuration set adds validated direct and CoT for Kimi-VL and Aria to the
second set. Intervals in \cref{tab:selection-cross-panel-sets,tab:selection-cross-panel} are
pointwise 95\% percentile bootstrap intervals in which the source selectors are refit in every
draw and the destination panel is resampled independently.

\noindent\textbf{Paired captions.} \cref{tab:transfer-paired} scores the natural counterpart of
the truth-switch test: for each caption order, both clips of a pair must be answered correctly.
The three readouts are bare-token direct at the original position, direct at the validated
position, and CoT with the same answer-extraction rule as the accuracy comparisons.

\begin{table}[ht]
\centering\scriptsize\revcolor
\caption{\rev{\textbf{Second-panel comparison and added models.} Accuracy (\%) under direct judgment at the validated position / CoT judgment with the settings
of \cref{app:revision-transfer}.}}
\label{tab:revision-transfer-accuracy}
\begin{tabular}{@{}llcc@{}}
\toprule
Panel & Model & Event-denial & Content-matched \\
\midrule
TOC-Bench & Qwen3-VL-32B & 66.62 / 44.75 & 65.50 / 64.12 \\
 & GLM-4.5V & 63.25 / 42.63 & 59.00 / 63.38 \\
 & InternVL3.5-38B & 86.50 / 73.88 & 54.50 / 59.63 \\
 & Molmo2-8B & 37.50 / 37.63 & 59.38 / 52.75 \\
 & Kimi-VL-A3B & 76.25 / 71.75 & 57.00 / 56.88 \\
 & Aria & 61.25 / 68.38 & 53.00 / 59.88 \\
Vinoground & Qwen3-VL-32B & 77.75 / 70.31 & 81.00 / 79.00 \\
 & GLM-4.5V & 82.06 / 71.75 & 78.81 / 81.12 \\
 & InternVL3.5-38B & 88.88 / 81.06 & 77.38 / 75.81 \\
 & Molmo2-8B & 73.44 / 69.63 & 65.00 / 63.88 \\
 & Kimi-VL-A3B & 75.25 / 76.88 & 67.13 / 67.00 \\
 & Aria & 59.63 / 59.19 & 65.88 / 63.56 \\
\bottomrule
\end{tabular}
\end{table}

\RevisionTransferTables
\FloatBarrier

\subsection{Selection robustness}
\label{app:revision-selection}
\noindent\textbf{Splits.} The cost is positive in all 20 cross-validation partitions and on the
separately fixed 200/200 split, for each of the three eligible sets of \cref{tab:selection}.
The partitions reuse the same 400 videos and are not independent replications. On the 200/200
split only the direct-only cost is significant (\cref{tab:selection-split}).

\noindent\textbf{Interval width.} With the selected configurations held fixed, the interval of
the all-eight cost is 33.88, 25.93, 18.50, and 13.00 points wide at 25, 50, 100, and 200
evaluation videos. At 200 videos the all-eight estimate of 4.75 points on the fixed split is
therefore compatible both with no cost and with the cross-validated 8.88 points.

\noindent\textbf{More configurations.} The 26 TOC-Bench configurations are the eight of
\cref{tab:selection}, the four validated direct readouts, the eight G0 and G1 settings with
native thinking off, GLM's G0 and G1 with native thinking on, and the direct and rationale
settings of Kimi-VL and Aria. With the same stored partitions and the same 400-video target the
cost is 8.39 points [1.98, 13.58]. On the separately fixed split it is 2.87 points (pointwise
95\% interval [$-4.75$, 13.75]).

\noindent\textbf{Common-protocol configurations.} On all 400 videos we also restricted the
eligible set to the four models' validated direct scores and their G1 scores with native
thinking off (\cref{tab:revision-natural}). With the same 20 five-fold partitions and 10,000
whole-video bootstrap draws refitting selection, the cost is 9.36 points [4.17, 14.50]
(pointwise; \cref{tab:revision-reconciliation}). \Cref{tab:validation-selection} separately
gives 9.36 points [2.89, 15.83] for the original eight configurations on the 388 videos
confirmed by at least one annotator.

\noindent\textbf{Validated videos.} On the three validated subsets of
\cref{app:revision-validation} the all-eight cost is 9.21, 10.70, and 9.36 points
(\cref{tab:validation-selection}).

\noindent\textbf{Second panel.} \cref{tab:selection-cross-panel} gives the cross-panel costs;
\cref{app:revision-transfer} defines the configurations.

\noindent\textbf{Mixture target.} Most of the generator's errors are content errors
(\cref{app:revision-rerank}). As a proxy for that target we weight each configuration's
event-denial and content-matched accuracy by the shares of the two error types among the
generator's errors as labeled by the grader over the full generator pool. The pooled shares
conditional on these two types are 4.11\% event-denial and 95.89\% content-matched. We
recompute the normalized weights separately on the selection and evaluation videos of each
fold, leaving errors of other kinds aside. Selecting on this weighted accuracy, and evaluating
the selected configuration on it held out, gives 7.45 points more than selecting on
event-denial accuracy among the eight configurations of \cref{tab:selection} (95\% interval
[2.47, 12.24]), and 9.40 points more with validated direct scores ([3.13, 13.30]). The proxy
covers only the two constructed negative types and does not estimate a full deployment mixture.

\begin{table}[ht]
\centering\small\revcolor
\caption{\rev{\textbf{Cross-panel selection costs. Points with pointwise 95\% intervals.}}}
\label{tab:selection-cross-panel}

\begin{tabular}{@{}lcc@{}}
\toprule
Configurations & Vinoground to TOC-Bench & TOC-Bench to Vinoground \\
\midrule
Eight with validated direct & $8.88 [4.75,14.75]$ & $3.63 [0.16,5.66]$ \\
Eight as submitted & $9.25 [4.75,14.25]$ & $3.25 [-0.41,5.39]$ \\
\bottomrule
\end{tabular}
\end{table}

\FloatBarrier

\subsection{Reranking and label audit}
\label{app:revision-rerank}

\noindent\textbf{Generation.} Qwen2.5-VL-7B, which is not one of the verifiers, wrote five
short answers to each question. The instruction requests one self-contained factual proposition
in declarative wording that names the relevant entities or events and preserves the temporal
relation, returned as a JSON field.

\noindent\textbf{Scoring.} Each verifier scores one answer at a time under direct judgment,
given the video and a clarified question; temporal-location questions use normalized time
labels where needed. This scoring prompt differs from the verification prompt of
\cref{sec:natural}. The verifier keeps its highest-scored candidate, and tied top scores share
the weight equally.

\noindent\textbf{Labels.} An LLM grader, given the reference answer, labels every candidate as
correct or incorrect and assigns an error type. Two annotators who are not authors audited 293
candidates across 138 videos without seeing the grader's labels and agreed with it on 97.6\%
($\kappa$ = 0.95). The candidates whose correctness remains unknown give the ranges of
\cref{tab:revision-rerank}.

\noindent\textbf{Results.} Direct Qwen leads direct InternVL by 4.96 points [1.29, 8.71] to
5.46 points [1.79, 9.21] over those ranges, by 5.65 points [2.06, 9.29] with audited labels
where available, and by 5.48 points [1.75, 9.16] on the 385 videos whose references both
annotators confirm. Excluding event-order questions gives a lower bound of 4.29 points [0.36,
8.57]. Over the first sample, Qwen gains 9--10 points and InternVL 4--5.

\noindent\textbf{Error types.} In a separate audit of 100 wrong answers from the original pool,
annotators who are not authors labeled 73 as content errors, 3 as event denials, and 24 as
errors of other kinds. Their labels and the grader's agreed on 95\% of the answers
($\kappa$ = 0.88).

\noindent\textbf{Second panel.} Only 9\% of the second panel's candidate pools contain both a
correct and an incorrect answer, so there are few pools where the verifier can change
correctness. We report reranking on TOC-Bench only.
\FloatBarrier

\subsection{Reconciliation of submitted and revised numbers}
\label{app:revision-reconciliation}
\begin{table}[ht]
\centering\scriptsize\revcolor
\caption{\rev{\textbf{Submitted and revised headline numbers.} Model order: Qwen, GLM, InternVL, Molmo. Accuracies are percentages.}}
\label{tab:revision-reconciliation}
\begin{tabularx}{\linewidth}{@{}p{.20\linewidth}YYp{.16\linewidth}@{}}
\toprule
Quantity & Submitted & Revised or added & Reason \\
\midrule
Qualifying construction cells & 7 & 6 & GLM answer position \\
Visible-start / endpoint win rates & 0.78--0.94 / 0.46--0.58 & 0.85--0.96 / 0.46--0.58 & GLM answer position \\
Truth-switch direct & 5.83 / 0.00 / 4.17 / 3.33 & 9.17 / 9.17 / 5.00 / 1.67 & Validated position \\
Truth-switch CoT & 75.00 / 88.33 / 88.33 / 14.17 & 100.00 / 90.00 / 70.00 / 25.00 & Common protocol \\
Natural direct, denial & 61.38 / 61.13 / 81.88 / 30.63 & 66.62 / 63.25 / 86.50 / 37.50 & Validated position \\
Natural direct, content & 64.38 / 55.63 / 54.13 / 57.50 & 65.50 / 59.00 / 54.50 / 59.38 & Validated position \\
Natural CoT, denial & 43.75 / 29.00 / 73.88 / 37.00 & 43.12 / 57.12 / 63.62 / 32.38 & Common protocol \\
Natural CoT, content & 63.88 / 48.25 / 59.38 / 52.13 & 64.12 / 63.75 / 60.12 / 53.25 & Common protocol \\
CoT minus direct, denial (pp) & $-17.63$ / $-32.13$ / $-8.00$ / $+6.38$ & $-23.50$ / $-6.12$ / $-22.88$ / $-5.12$ & Both changes \\
GLM native thinking, denial (pp) & $-32.13$; 194 invalid or capped & $-19.13 [-24.38,-13.88]$; 2 capped & Separate sensitivity \\
Significant denial declines & 3 of 4 & 2 of 4; third in the native-thinking comparison & Separate families \\
Significant negative-type shifts & 4 of 4 & 3 of 4 & Common protocol \\
Significant model-gap changes & 3 of 6 & 5 of 6 & Common protocol \\
Qwen $-$ InternVL gap change (pp) & $-34.63 [-43.00,-26.25]$ & $-24.50 [-33.00,-16.00]$ & Common protocol \\
Selection cost (pp) & $8.88 [2.46,15.30]$ & $8.39 [1.98,13.58]$ with 26; 8.88 with eight & Submitted eight retained \\
Selection cost, eight common-protocol configurations (pp) & -- & $9.36 [4.17,14.50]$ (pointwise) & Common protocol \\
\bottomrule
\end{tabularx}
\end{table}

\cref{tab:revision-reconciliation} lists every headline number that changed. The changes have
two sources, the answer position and the common protocol, and three consequences for the
claims.

\noindent\textbf{Answer position.} Direct judgment is now read at the validated answer position
(\cref{app:revision-position}). In the construction study this affects GLM only: its shell win
rate against visible-start negatives rises from 0.894 to 0.958, and its toggle drop of 15.74
points has a simultaneous lower bound of $-2.88$, so GLM on toggle no longer qualifies and six
cells qualify instead of seven (\cref{tab:construction-reread}). On the truth-switch test GLM
answered A on all 240 direct requests at the original position, for 0.00\% paired success, and
it scores 9.17\% at the validated position. On natural video the direct accuracies change by
0.37 to 6.87 points.

\noindent\textbf{Common protocol.} On the truth-switch test, Qwen rises from 75.00\% to
100.00\% because the 1,024-token cap of the submission truncated 31 of 240 responses. InternVL
falls from 88.33\% to 70.00\% because the submission used its thinking-mode system prompt and
the common protocol turns native thinking off; with native thinking on, a 16-pair probe gives
100\% (\cref{tab:truth-prompt-controls}). Molmo rises from 14.17\% to 25.00\%, with 14 instead
of 35 responses that make no choice. On natural video, GLM's event-denial accuracy under CoT
judgment changes from 29.00\% to 57.12\%: in the submission 194 of its 800 responses were
invalid or truncated under native thinking, and under the common protocol native thinking is
off and 8 of 800 responses express no choice (\cref{tab:generation-glm}). InternVL's changes
from 73.88\% to 63.62\% for the same reason as on the truth-switch test.

\noindent\textbf{Consequences for the claims.} Within one family, significant event-denial
declines go from three of four models (Qwen, GLM, InternVL) to two (Qwen, InternVL); GLM's
decline is significant with native thinking on ($-19.13$ points [$-24.38$, $-13.88$], with 2 of
800 responses capped). The change against event-denial negatives differs significantly from the
change against content-matched negatives for three models instead of four; Molmo's difference,
positive in the submission, is no longer significant. Significant changes in model gaps go from
three of six to five of six. The selection analysis keeps its inputs, so its cost is unchanged
at 8.88 points; with 26 configurations it is 8.39 points.

\begin{table}[ht]
\centering\small
\caption{\textbf{Protocol labels.} Meanings and other names used in tables and in the artifact.}
\label{tab:protocol-labels}
\begin{tabularx}{\linewidth}{@{}p{.22\linewidth}Yp{.21\linewidth}@{}}
\toprule
Label & Meaning & Also appears as \\
\midrule
Common protocol & One set of instructions for every model; native thinking off; direct judgment at the validated
answer position; CoT judgment requests a brief rationale and a final answer line under a
requested cap of 8,192 tokens; the last expressed choice is extracted. Used in
\cref{sec:truth-switch} to \cref{sec:natural-evidence} and
\cref{app:revision-interface,app:revision-controlled,app:revision-render,app:revision-denial,app:revision-natural}. & none \\
Per-model protocol & The protocol of the original submission (\cref{app:checkpoints}): CoT settings chosen per
model, CoT answers parsed from the final answer line with unparseable and truncated outputs
incorrect, direct judgment at the original answer position. Used in
\cref{app:truth-switch,app:natural,app:selection}, \cref{fig:bottle-occlusion}, and
\cref{tab:selection}. & ``submitted'', ``of the submission'' \\
Original answer position & The first token after the prompt. & ``original position'' \\
Validated answer position & The position at which the model's own concise answer places its answer letter
(\cref{app:revision-position}). & ``validated direct'' \\
G0, G1 & Generic instructions requesting a concise answer (G0) or a brief rationale (G1). & ``concise'', ``rationale'' \\
P0, P1 & G0 and G1 with a verification procedure added (\cref{app:revision-interface}). & none \\
CoT judgment & Under the common protocol, P1 on the controlled tasks and G1 on natural video. & ``rationale'' \\
Generated-answer control & G0 on natural video; an answer generated under an eight-token limit on the truth-switch test
using P0 from the validated answer position. & ``8 tokens'', ``concise'' \\
Native thinking & The built-in reasoning mode of GLM and of InternVL; off in the common protocol. & ``Mode'' \\
Eight per-model configurations & Four core models with the original direct readout and per-model CoT settings. The cross-panel comparison re-extracts the CoT choices (\cref{app:revision-transfer}). & ``all eight'', ``eight as submitted'' \\
Eight with validated direct & Four core models with validated direct scores and the cross-panel CoT scores described in \cref{app:revision-transfer}. & ``eight, validated direct'' \\
26 configurations & Listed in \cref{app:revision-selection}. & none \\
Interface effect & The change in a score from direct judgment to CoT judgment. & ``CoT minus direct'' \\
Events given, event must be found & On the controlled tasks, the frames disclose each event. On natural video, the event has to be located in the frames. & ``controlled'', ``natural'' \\
\bottomrule
\end{tabularx}
\end{table}
\FloatBarrier
\endgroup

\end{document}